\documentclass{article} % For LaTeX2e
\usepackage{iclr2024_conference,times}

\usepackage{amsmath,amsfonts,bm}

\def\eqref#1{equation~\ref{#1}}
\def\1{\bm{1}}

\DeclareMathAlphabet{\mathsfit}{\encodingdefault}{\sfdefault}{m}{sl}
\SetMathAlphabet{\mathsfit}{bold}{\encodingdefault}{\sfdefault}{bx}{n}

\usepackage{hyperref}
\usepackage{url}

\usepackage{booktabs}       % professional-quality tables
\usepackage{amsfonts}       % blackboard math symbols
\usepackage{nicefrac}       % compact symbols for 1/2, etc.
\usepackage{microtype}      % microtypography
\usepackage{xcolor}         % colors
\usepackage{multirow}
\usepackage{algorithm}
\usepackage{algpseudocode}
\usepackage{subfig}
\usepackage{wrapfig}
\usepackage{arydshln}
\usepackage{enumitem}
\usepackage{xfrac}
\setlist{leftmargin=5.5mm}
\usepackage{natbib}
\usepackage{multicol}
\usepackage{caption}
\usepackage{color, colortbl}
\definecolor{Gray}{rgb}{0.9, 0.9, 0.9}
\usepackage{titlesec}

\titlespacing*{\section}
{0pt}{0ex}{0ex}
\titlespacing*{\subsection}
{0pt}{0ex}{-1ex}
\titlespacing*{\subsubsection}
{0pt}{0ex}{-2ex}

\usepackage{amsmath}
\usepackage{amssymb}
\usepackage{mathtools}
\usepackage{amsthm}

\theoremstyle{plain}
\newtheorem{theorem}{Theorem}[section]

\newtheorem{lemma}[theorem]{Lemma}

\theoremstyle{definition}

\newtheorem{assumption}[theorem]{Assumption}
\theoremstyle{remark}
\newtheorem{remark}[theorem]{Remark}

\title{Internal Cross-layer Gradients for Extending Homogeneity to Heterogeneity in Federated Learning}

\author{Yun-Hin Chan, Rui Zhou, Running Zhao, Zhihan Jiang \& Edith C.H. Ngai\thanks{Corresponding author.}  \\
Department of Electrical and Electronic Engineering, The University of Hong Kong\\
\texttt{\{chanyunhin,zackery,rnzhao,zhjiang\}@connect.hku.hk,}\\
\texttt{chngai@eee.hku.hk}
}

\iclrfinalcopy % Uncomment for camera-ready version, but NOT for submission.

\begin{document}

\maketitle

\vskip -0.3in
\begin{abstract}
Federated learning (FL) inevitably confronts the challenge of system heterogeneity in practical scenarios. To enhance the capabilities of most model-homogeneous FL methods in handling system heterogeneity, we propose a training scheme that can extend their capabilities to cope with this challenge.
In this paper, we commence our study with a detailed exploration of homogeneous and heterogeneous FL settings and discover three key observations: (1) a positive correlation between client performance and layer similarities, %across distinct client models, 
(2) higher similarities in the shallow layers in contrast to the deep layers, and (3) the smoother gradient distributions indicate the higher layer similarities.
Building upon these observations, we propose InCo Aggregation that leverages internal cross-layer gradients, a mixture of gradients from shallow and deep layers within a server model, to augment the similarity in the deep layers without requiring additional communication between clients. 
Furthermore, our methods can be tailored to accommodate model-homogeneous FL methods such as FedAvg, FedProx, FedNova, Scaffold, and MOON, to expand their capabilities to handle the system heterogeneity.
Copious experimental results validate the effectiveness of InCo Aggregation, spotlighting internal cross-layer gradients as a promising avenue to enhance the performance in heterogeneous FL. 
\end{abstract}

\section{Introduction}

Federated learning (FL) is proposed to enable a federation of clients to effectively cooperate towards a global objective without exchanging raw data \citep{mcmahan2017communication}. While FL makes it possible to fuse knowledge in a federation with privacy guarantees \citep{huang2021evaluating, mcmahan2017communication, jeong2022factorized}, its inherent attribute of system heterogeneity \citep{li2020federated}, i.e., varying resource constraints of local clients, may hinder the training process and even lower the quality of the jointly-learned models \citep{kairouz2019advances,li2020federated, mohri2019agnostic, Survey_on_Heterogeneous}.

System heterogeneity refers to a set of varying physical resources $\{R_i\}_{i=1}^n$, where $R_i$ denotes the resource of client $i$, a high-level idea of resource that holistically governs the aspects of computation, communication, and storage.
Existing works cater to system heterogeneity through a methodology called model heterogeneity, which aligns the local models of varying architectures to make full use of local resources \citep{diao2021heterofl, baek2022joint, alam2022fedrolex, huang2022learn, fang2022robust, lin2020ensemble}. Specifically, model heterogeneity refers to a set of different local models $\{M_i\}_{i=1}^n$ with $M_i$ being the model of client $i$. Let $R(M)$ denote the resource requirement for the model $M$. Model heterogeneity is a methodology that manages to meet the constraints $\{R(M_i)\leq R_i\}_{i=1}^n$.
In the case of model heterogeneity, heterogeneous devices are allocated to a common model prototype tailored to their varying sizes, such as ResNets with different depths or widths of layers \citep{liu2022no, diao2021heterofl, horvath2021fjord, baek2022joint, caldas2018expanding, ilhan2023scalefl}, strides of layers \citep{tan2022fedproto}, or numbers of kernels \citep{alam2022fedrolex}, to account for their inherent resource constraints.
While several methods have been proposed to incorporate heterogeneous models into federated learning (FL), their performances often fall short compared to FL training using homogeneous models of the same size \citep{he2020group, diao2021heterofl}. Therefore, gaining a comprehensive understanding of the factors that limit the performance of heterogeneous models in FL is imperative. The primary objective of this paper is to investigate the underlying reasons behind this limitation and propose a potential solution that acts as a bridge between model homogeneity and heterogeneity to tackle this challenge.

In light of this, we first conduct a case study to reveal the obstacles affecting the performance of heterogeneous models in FL.
The observations from this case study are enlightening: 
(1) With increasing heterogeneity in data distributions and model architectures, we observe a decline in model accuracy and layer-wise similarity (layer similarity) as measured by Centered Kernel Alignment (CKA)\footnote{The detailed descriptions for CKA are introduced in Appendix \ref{sec:CKA}.} \citep{kornblith2019similarity}, a quantitative metric of bias \citep{luo2021no, raghu2021vision}; 
(2) The deeper layers share lower layer similarity across the clients, while the shallower layers exhibit greater alignment.
These insights further shed light on the notion that shallow layers possess the ability to capture shared features across diverse clients, even within the heterogeneous FL setting.
Moreover, these observations indicate that the inferior performances in heterogeneous FL are related to the lower similarity in the deeper layers.
Motivated by these findings, we come up with an idea: \textbf{Can we enhance the similarity of deeper layers, thereby attaining improved performance?}

To answer this question, we narrow our focus to the gradients, as the dissimilarity of deep layers across clients is a direct result of gradient updates \citep{ruder2016overview, chen2021communication}. Interestingly, we observe that (3) the gradient distributions originating from shallow layers are smoother and possess higher similarity than those from deep layers, establishing a connection between the gradients and the layer similarity.
Therefore, inspired by these insights, we propose a method called \textbf{InCo Aggregation}, deploying different model splitting methods and utilizing the \textbf{In}ternal \textbf{C}r\textbf{o}ss-layer gradients (\textbf{InCo}) in a server model to improve the similarity of its deeper layers without additional communications with the clients. 
More specifically, cross-layer gradients are mixtures of the gradients from the shallow and the deep layers.
We utilize cross-layer gradients as internal knowledge, effectively transferring knowledge from the shallow layers to the deep layers.
Nevertheless, mixing these gradients directly poses a significant challenge called gradient divergence \citep{wang2020tackling, zhao2018federated}. 
To tackle this issue, we normalize the cross-layer gradients and formulate a convex optimization problem that rectifies their directions. 
In this way, InCo Aggregation automatically assigns optimal weights to the cross-layer gradients, thus avoiding labor-intensive parameter tuning. Furthermore, \textbf{InCo Aggregation can extend to model-homogeneous FL methods that previously do not support model heterogeneity}, such as FedAvg\citep{mcmahan2017communication}, FedProx \citep{li2018federated}, FedNova \citep{wang2020tackling}, Scaffold \citep{karimireddy2020scaffold}, and MOON \citep{li2021model}, to develop their abilities in managing the model heterogeneity problem.

Our main contributions are summarized as follows:
\vspace{-0.1in}
\begin{itemize}
    \item We first conduct a case study on homogeneous and heterogeneous FL settings and find that (1) client performance is positively correlated to layer similarities across different client models, (2) similarities in the shallow layers are higher than the deep layers, and (3) smoother gradient distributions hint for higher layer similarities.
    \vspace{-0.03in}
    \item We propose InCo Aggregation, applying model splitting and the internal cross-layer gradients inside a server model. Moreover, our methods can be seamlessly applied to various model-homogeneous FL methods, equipping them with the ability to handle model heterogeneity. 
    \vspace{-0.03in}
    \item We establish the non-convex convergence of utilizing cross-layer gradients in FL and derive the convergence rate. 
    \vspace{-0.03in}
    \item Extensive experiments validate the effectiveness of InCo Aggregation, showcasing its efficacy in strengthening model-homogeneous FL methods for heterogeneous FL scenarios.
    \vspace{-0.03in}
\end{itemize}

\begin{figure}[!t]
\vskip -0.45in
\centering
\subfloat[The CKA similarity of ResNets.]
{\includegraphics[width=0.31\columnwidth]{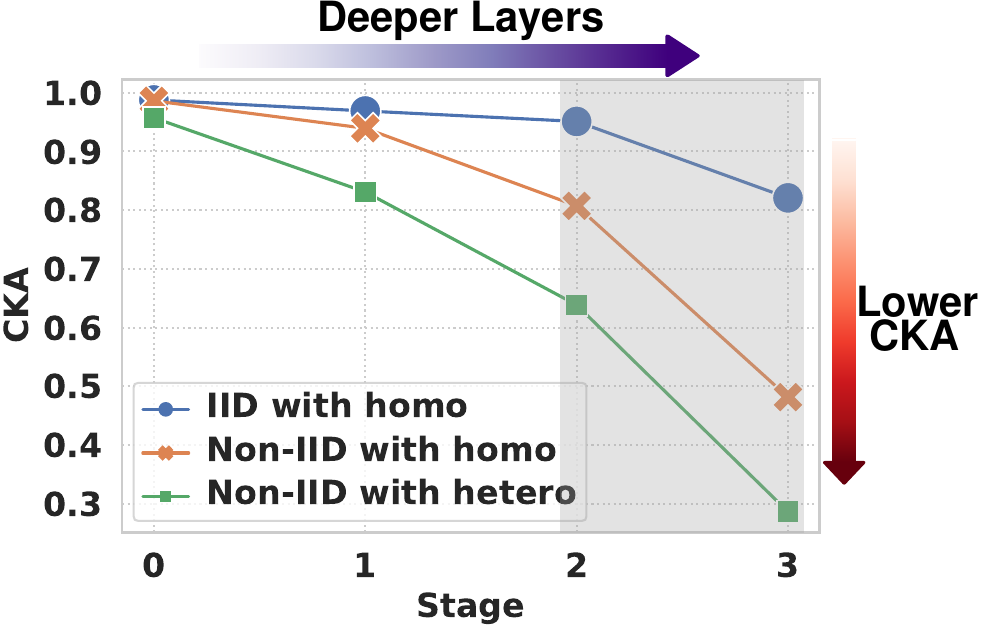}
\label{fig:Analysis_resnet_CKA}}
\hfil
\subfloat[The CKA similarity of ViTs.]
{\includegraphics[width=0.32\columnwidth]{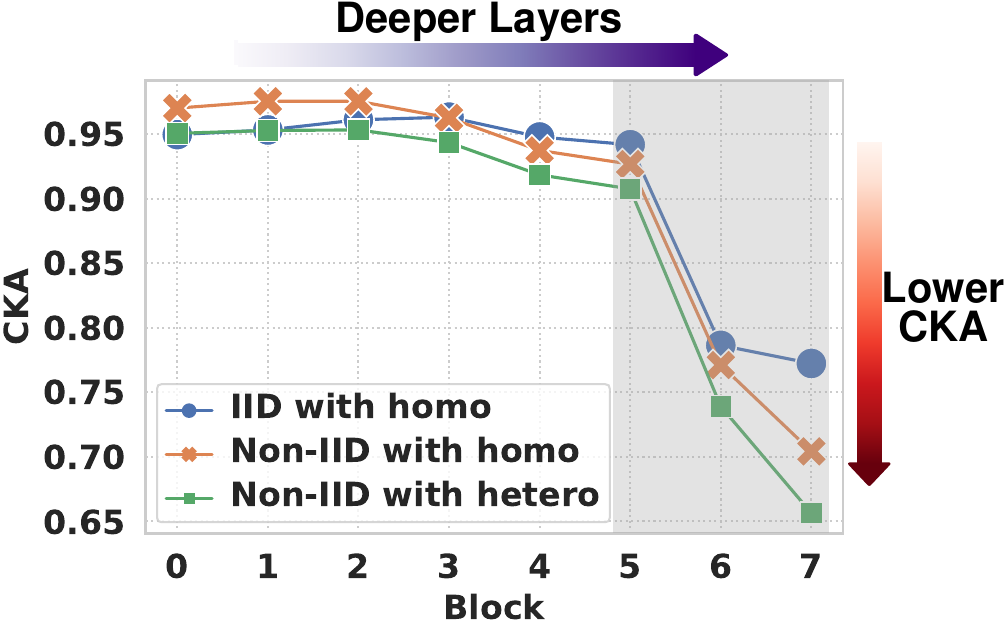}
\label{fig:Analysis_vit_CKA}}
\hfil
\subfloat[The relations between CKA and accuracy.]
{\includegraphics[width=0.287\columnwidth]{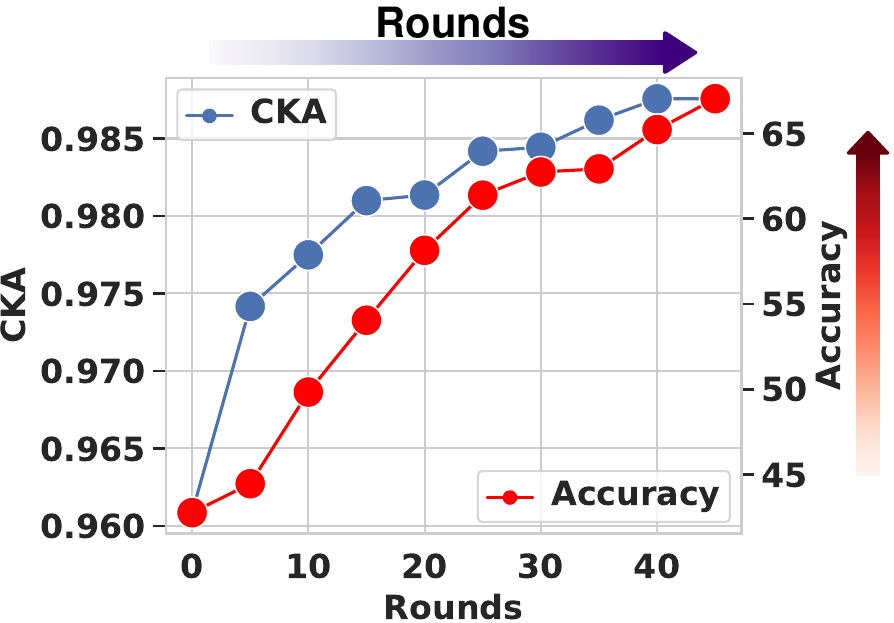}
\label{fig:Analysis_CKA_acc_relation}}
\vskip -0.07in
\caption{CKA similarity in different environments and the relation between accuracy and CKA similarity. (a) and (b): The CKA similarity of different federated settings. (c): The positive relation between CKA and accuracy during the training process.}
\label{fig:Analysis_CKA}
\vskip -0.2in
\end{figure}

\section{Preliminary}

To investigate the performance of clients in diverse federated learning settings, we present a case study encompassing both homogeneous and heterogeneous model architectures with CIFAR-10 and split data based on IID and Non-IID with ResNets \citep{he2016identity} and ViTs \citep{dosovitskiy2020image}.
We use CKA \citep{kornblith2019similarity} similarities among models to measure the level of bias exhibited by each model. 
More detailed results of the case study are provided in Appendix~\ref{apd:case_study}.

\subsection{A Case Study in Different Federated Learning Environments}

\begin{figure}[!t]
\vskip -0.45in
\centering
\subfloat[Similarity of gradients in Stage 2.]
{\includegraphics[width=0.23\columnwidth]{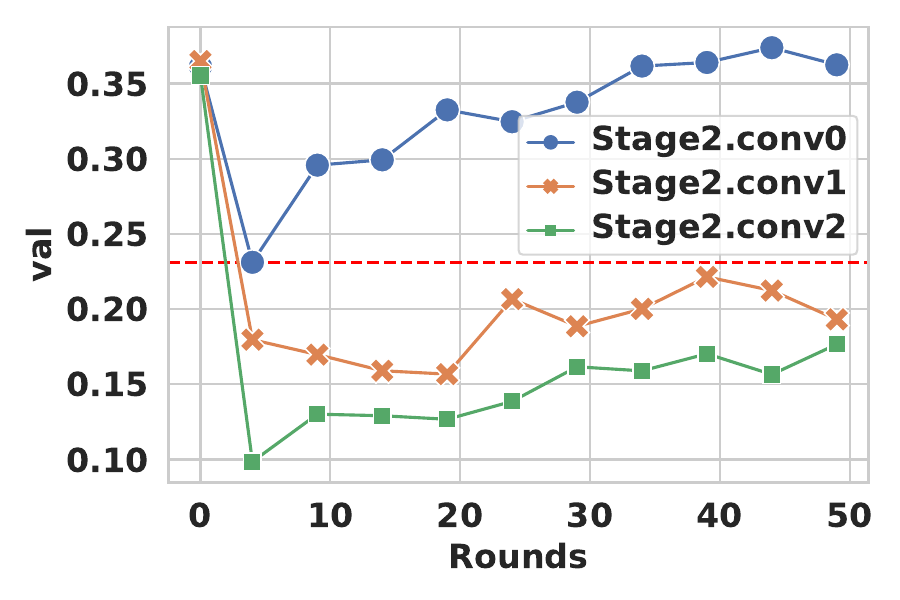}
\label{fig:Analysis_Cross_CKA_2}}
\hfil
\subfloat[Similarity of gradients in Stage 3.]
{\includegraphics[width=0.23\columnwidth]{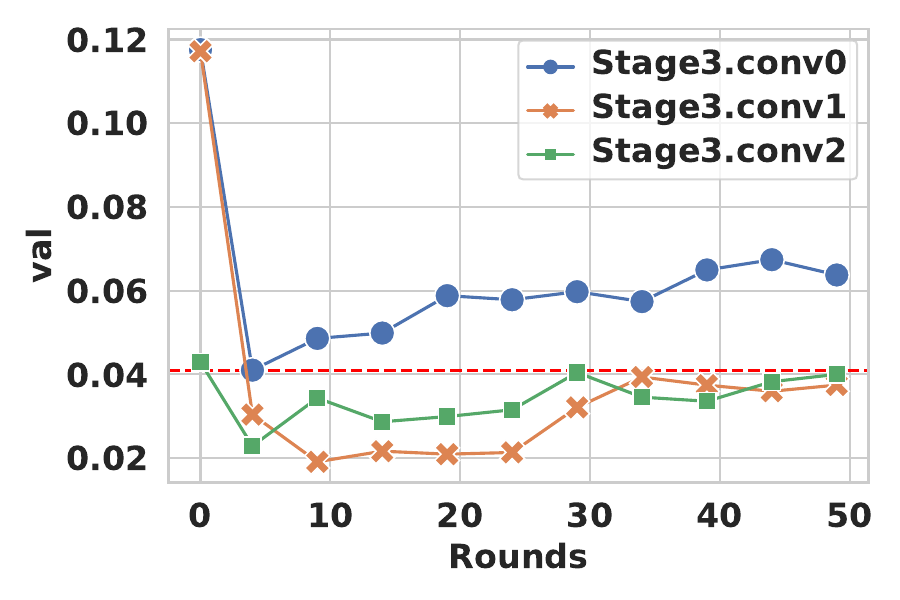}
\label{fig:Analysis_Cross_CKA_3}}
\hfil
\subfloat[Gradient distributions in Non-IID with hetero.]
{\includegraphics[width=0.23\columnwidth]{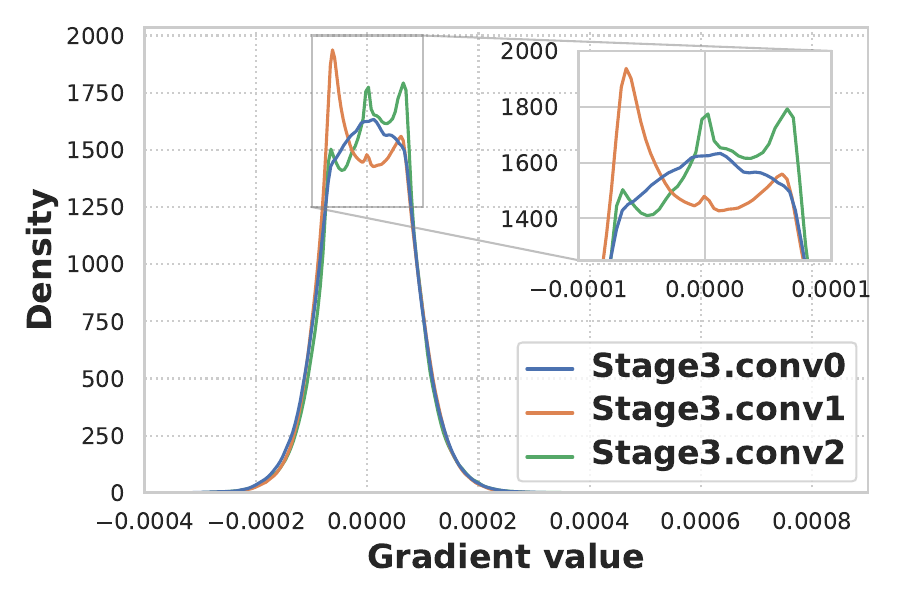}
\label{fig:Gradient_Dis_Hetero_FedAvg}}
\hfil
\subfloat[Gradient distributions in IID with homo.]
{\includegraphics[width=0.23\columnwidth]{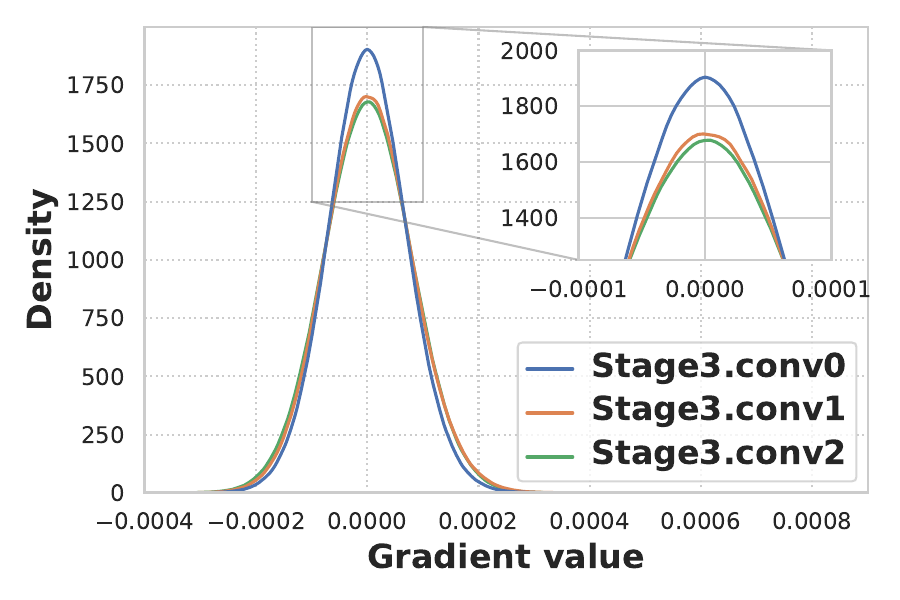}
\label{fig:Gradient_Dis_IID_FedAvg}}
\vskip -0.07in
\caption{Cross-environment similarity and gradients distributions. (a) and (b): Similarity from Stage 2 and Stage 3. (c) and (d): The gradient distributions of Non-IID with hetero and IID with homo.}
\label{fig:deep_analysis_gradients}
\vskip -0.15in
\end{figure}

\textit{Case Analysis.}
Generally, we find three intriguing observations from Table~\ref{tab:acc_train_loss_case} and Figure~\ref{fig:Analysis_CKA}:  
\begin{wraptable}{r}{0.35\columnwidth}
\vskip -0.1in
\def\arraystretch{1.2}
\scriptsize
    \caption{Accuracy of the case study.}
    \label{tab:acc_train_loss_case}
    \centering
    \begin{tabular}{ccc}
    \toprule
    % \hline
        \multicolumn{2}{c}{Settings}  & Test Accuracy  \\
      \midrule
      % \hline
      \multirow{3}{*}{\rotatebox{90}{ResNet}}  & IID with homo & 81.0   \\
       & Non-IID with homo & 62.3({\color{blue}{$\downarrow$18.7}})  \\
       & Non-IID with hetero & 52.3({\color{blue}{$\downarrow$28.7}})  \\
       \hline
       \multirow{3}{*}{\rotatebox{90}{ViT}}  & IID with homo & 81.0 \\
       & Non-IID with homo & 54.8({\color{blue}{$\downarrow$26.2}}) \\
       & Non-IID with hetero & 50.1({\color{blue}{$\downarrow$30.9}}) \\
    \bottomrule
    % \hline
    \end{tabular}
% \end{table}
\vskip -0.1in
\end{wraptable}
(i) The deeper layers or stages have lower CKA similarities than the shallow layers. 
(ii) The settings with higher accuracy also obtain higher CKA similarities in the deeper layers or stages.
(iii) The CKA similarity is positively related to the accuracy of clients, as shown in Figure~\ref{fig:Analysis_CKA_acc_relation}.
These observations indicate that increasing the similarity of deeper layers can serve as a viable approach to improving client performance. Considering that shallower layers exhibit higher similarity, a potential direction emerges: to improve the CKA similarity in deeper layers according to the knowledge from the shallower layers.

\subsection{Deep Insights of Gradients in the Shallower Layers}
\label{sec:deep_insights_of_grads}
\textit{Gradients as Knowledge.} 
In FL, there are two primary types of knowledge that can be utilized: features, which are outputs from middle layers, and gradients from respective layers. 
We choose to use gradients as our primary knowledge for two essential reasons. Firstly, our FL environment lacks a shared dataset, impeding the establishment of a connection between different clients using features derived from the same data. Secondly, utilizing features in FL would significantly increase communication overheads. Hence, taking these practical considerations into account, we select gradients as the knowledge.

\begin{figure}[htbp]
\vskip -0.2in
\centering
\subfloat[Stage3 conv0 in Non-IID with hetero.]
{\includegraphics[width=0.23\columnwidth]{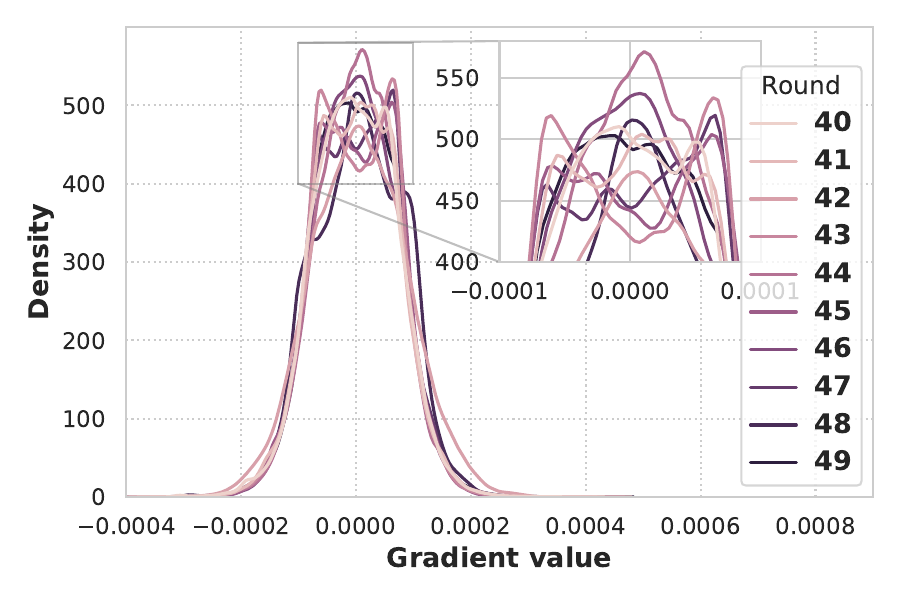}
\label{fig:Analysis_Gradient_Dis_Hetero_FedAvg_stage3_conv0_4050}}
\hfil
\subfloat[Stage3 conv1 in Non-IID with hetero.]
{\includegraphics[width=0.23\columnwidth]{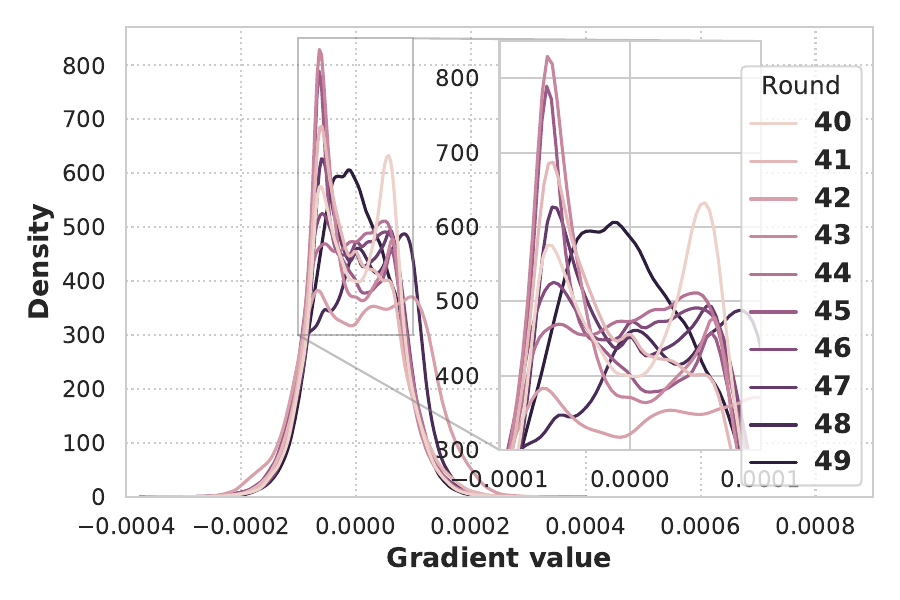}
\label{fig:Analysis_Gradient_Dis_Hetero_FedAvg_stage3_conv1_4050}}
\hfil
\subfloat[Stage3 conv0 in IID with homo.]
{\includegraphics[width=0.23\columnwidth]{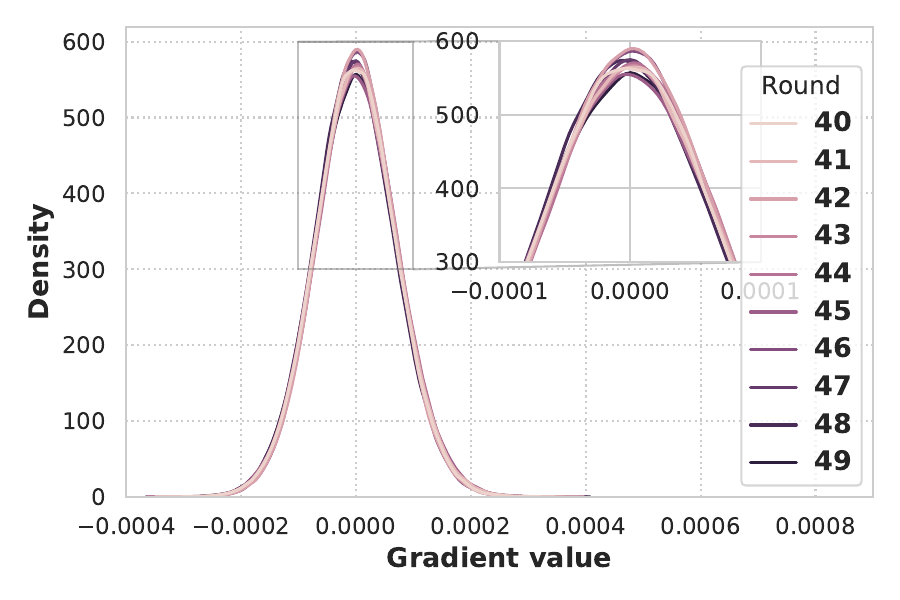}
\label{fig:Analysis_Gradient_Dis_IID_FedAvg_stage3_conv0_4050}}
\hfil
\subfloat[Stage3 conv1 in IID with homo.]
{\includegraphics[width=0.23\columnwidth]{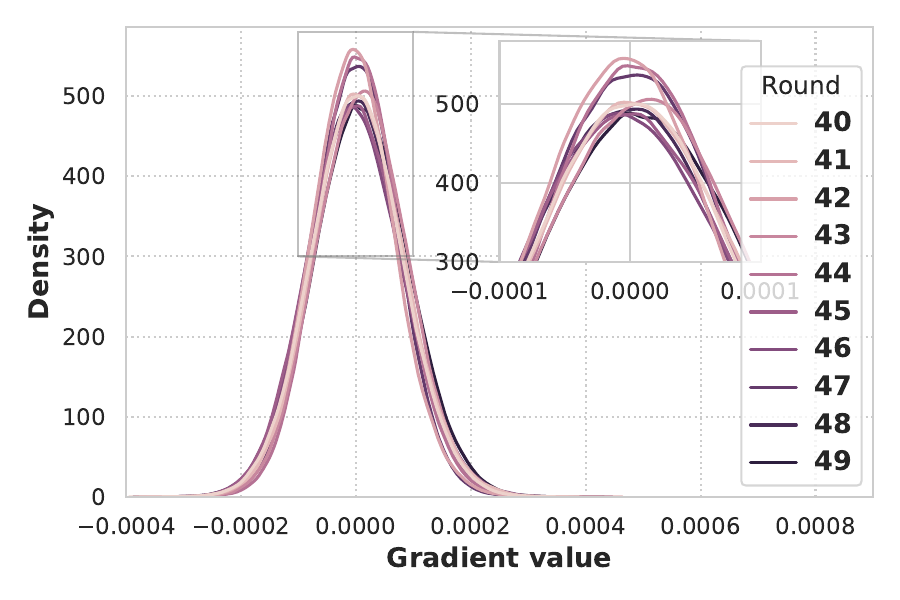}
\label{fig:Analysis_Gradient_Dis_IID_FedAvg_stage3_conv1_4050}}
\vskip -0.07in
\caption{The gradient distributions from round 40 to 50 in different environments.}
\label{fig:Analysis_Gradient_Dis_IID_Hetero_FedAvg_4050}
\vskip -0.1in
\end{figure}

\textit{Cross-environment Similarity.}
In this subsection, we deeply investigate the cross-environment similarity of gradients between two environments, i.e., \textit{IID with homo} and \textit{Non-IID with hetero}, to shed light on the disparities between shallow and deep layers in the same stage\footnote{We discuss a shallow layer (the first layer with the same shape in a stage) and deep layers (remaining layers) within a stage for ResNets and a block for ViTs. The gradient analyses for ViTs are introduced in Appendix~\ref{subsec:apd_gradients_analysis_ViTs}} and identify the gaps between the homogeneous and heterogeneous FL. As depicted in Figure~\ref{fig:Analysis_Cross_CKA_2} and \ref{fig:Analysis_Cross_CKA_3}, gradients from shallow layers (\textit{Stage2.conv0} and \textit{Stage3.conv0}) exhibit higher cross-environment CKA similarity than those from deep layers such as \textit{Stage2.conv1}, and \textit{Stage3.conv2}. Notably, even the lowest similarities (red lines) in \textit{Stage2.conv0} and \textit{Stage3.conv0} surpass the highest similarities in deep layers. 
These findings underscore the superior quality of gradients obtained from shallow layers relative to those obtained from deep layers, and also indicate that the layers within the same stage exhibit similar patterns to the layers throughout the entire model.

\textit{Gradient Distributions.}
To dig out the latent relations between gradients and layer similarity, we delve deeper into the analysis of gradient distributions across different FL environments.
More specifically, the comparison of Figure~\ref{fig:Gradient_Dis_Hetero_FedAvg} and Figure~\ref{fig:Gradient_Dis_IID_FedAvg} reveals that gradients from shallow layers (\textit{Stage3.conv0}) exhibit greater similarity in distribution between \textit{Non-IID with hetero} and \textit{IID with homo} environments, in contrast to deep layers (\textit{Stage3.conv1} and \textit{Stage3.conv2}). 
Additionally, as depicted in Figure~\ref{fig:Analysis_Gradient_Dis_IID_FedAvg_stage3_conv0_4050} and Figure~\ref{fig:Analysis_Gradient_Dis_IID_FedAvg_stage3_conv1_4050}, the distributions of gradients from a deep layer (Figure~\ref{fig:Analysis_Gradient_Dis_IID_FedAvg_stage3_conv1_4050}) progressively approach the distribution of gradients from a shallow layer (Figure~\ref{fig:Analysis_Gradient_Dis_IID_FedAvg_stage3_conv0_4050}) with each round, in contrast to Figure~\ref{fig:Analysis_Gradient_Dis_Hetero_FedAvg_stage3_conv0_4050} and Figure~\ref{fig:Analysis_Gradient_Dis_Hetero_FedAvg_stage3_conv1_4050}, where the distributions from deep layers (Figure~\ref{fig:Analysis_Gradient_Dis_Hetero_FedAvg_stage3_conv1_4050}) are less smooth than those from shallow layers (Figure~\ref{fig:Analysis_Gradient_Dis_Hetero_FedAvg_stage3_conv0_4050}) in \textit{Non-IID with hetero} during rounds 40 to 50. 
Consequently, drawing from the aforementioned gradient analysis, we can enhance the quality of gradients from deep layers in \textit{Non-IID with hetero} environments by leveraging gradients from shallow layers, i.e., cross-layer gradients as introduced in the subsequent section.

\section{InCo Aggregation}

\begin{figure}[!t]
\vskip -0.4in
\centering
\subfloat[Layer splitting.]
{\includegraphics[width=0.325\columnwidth]{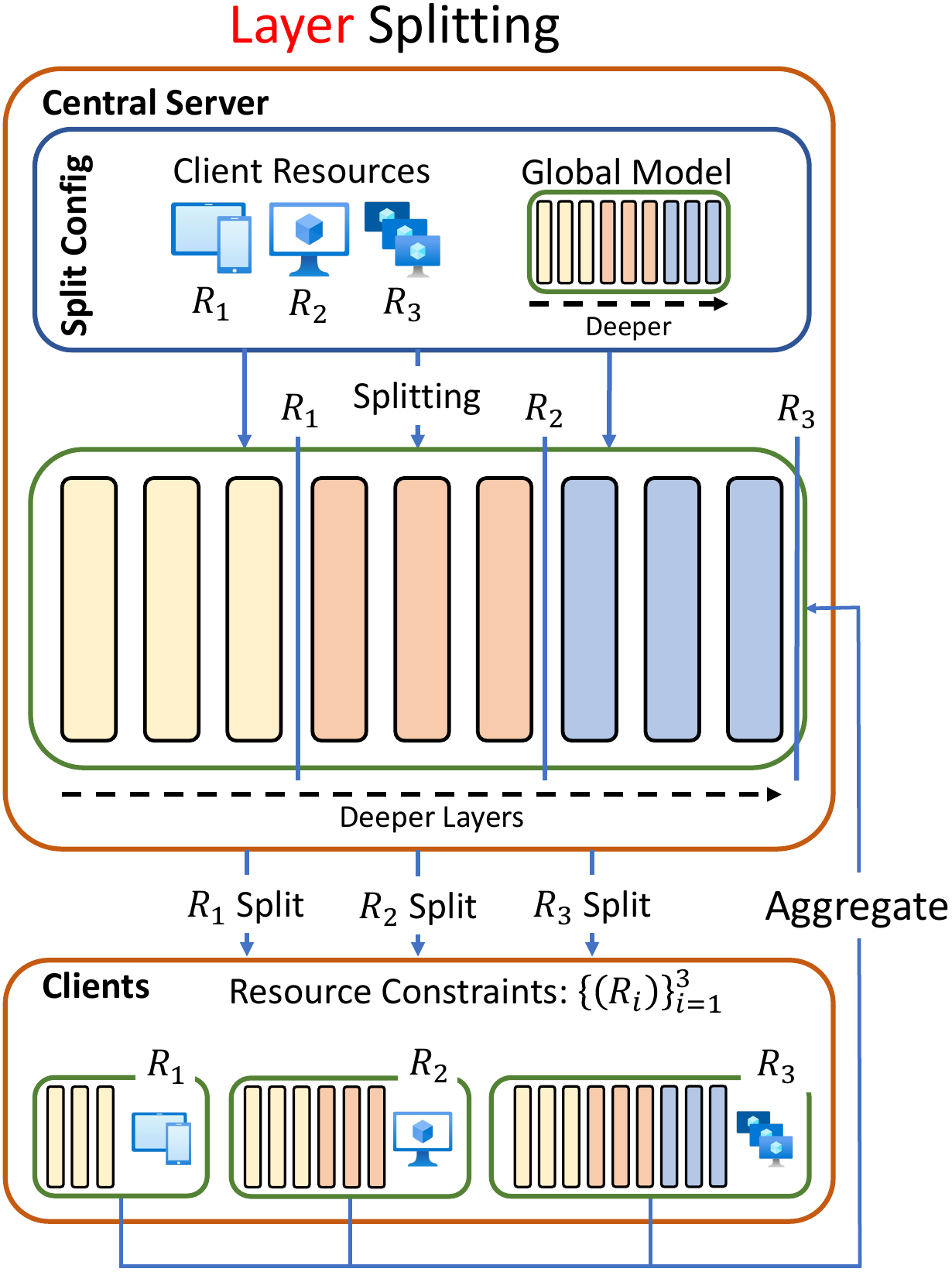}
\label{fig:layer_split}}
% \hfil
\subfloat[Stage splitting.]
{\includegraphics[width=0.325\columnwidth]{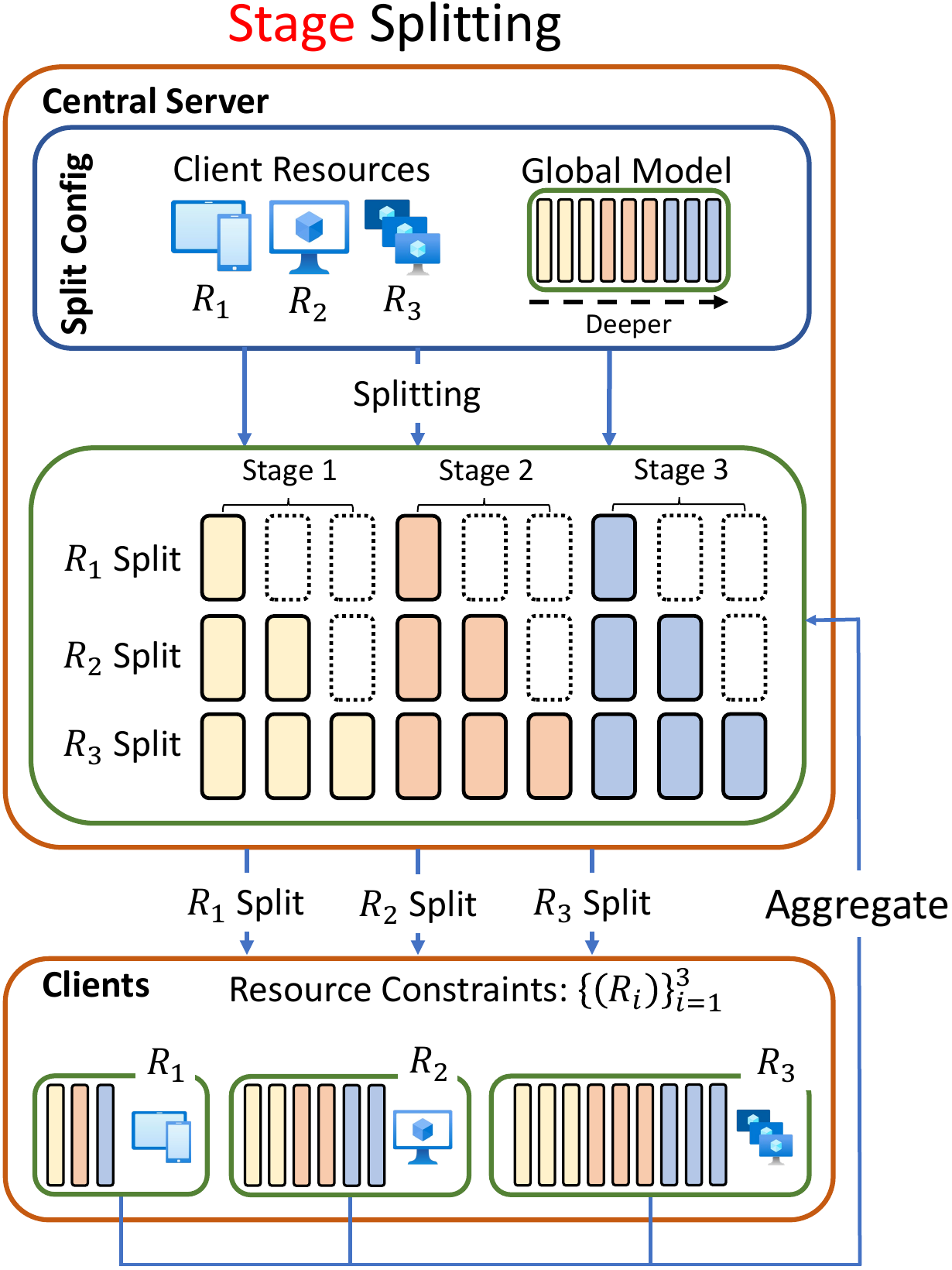}
\label{fig:stage_split}}
\hfil
\subfloat[Hetero splitting.]
{\includegraphics[width=0.325\columnwidth]{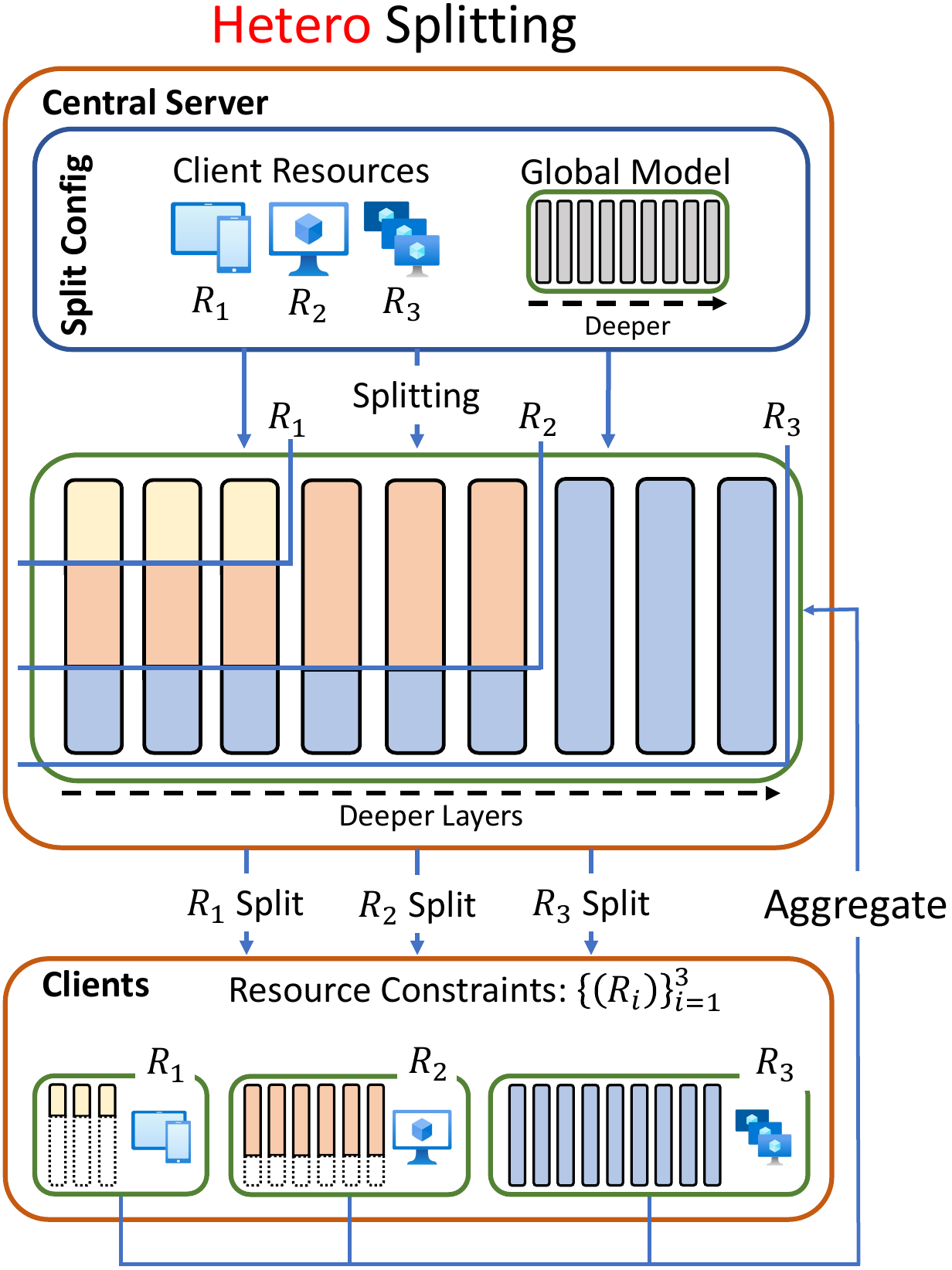}
\label{fig:hetero_split}}
\vskip -0.07in
\caption{The system architecture of three different model splitting methods: (a) layer splitting, (b) stage splitting, and (c) heterogeneous (hetero) splitting. (a): Layer splitting divides the entire model layer by layer. (b): Stage splitting separates each stage layer by layer. (c): Hetero splitting partitions the whole model in different widths and depths depending on the available resources $R_i$ of client $i$.}
\label{fig:split_types}
\vskip -0.15in
\end{figure}

We provide a concise overview of the three key components in InCo Aggregation at first. 
\begin{wrapfigure}{r}{0.4\columnwidth}
\vspace{-0.2in}
\begin{center}
\centerline{\includegraphics[width=0.3\columnwidth]{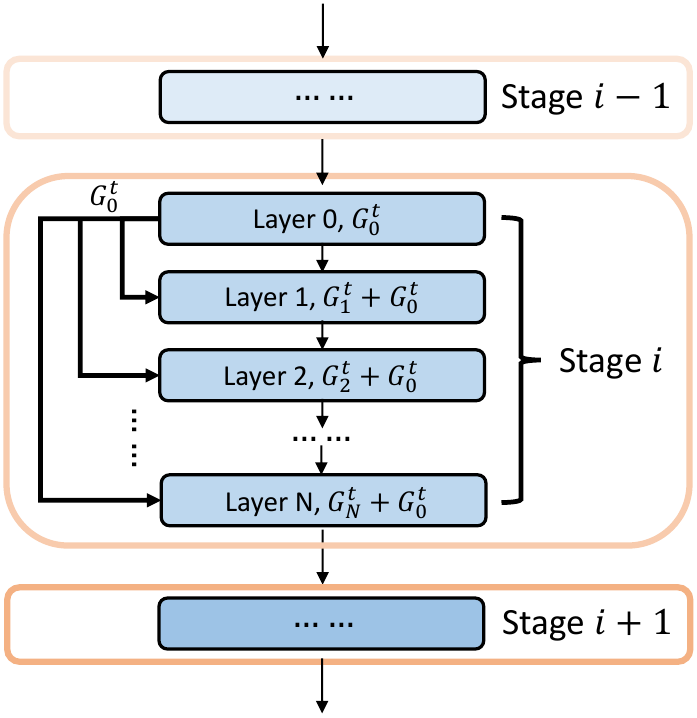}}
\vspace{-0.2cm}
\caption{Cross-layer gradients for the server model in InCo.}
\label{fig:cross_layer_grads}
\end{center}
\vskip -0.4in
\end{wrapfigure}
The first component is model splitting, including three types of model splitting methods, as shown in Figure~\ref{fig:split_types}. The second component involves the combination of gradients from a shallow layer and a deep layer, referred to as internal cross-layer gradients. 
To address gradient divergence, the third component employs gradient normalization and introduces a convex optimization formulation.
We elaborate on these three critical components of InCo Aggregation as follows.

\subsection{Model Splitting}

To facilitate model heterogeneity, we propose three model splitting methods: layer splitting, stage splitting, and hetero splitting, as illustrated in 
Figure~\ref{fig:split_types}. 
These methods distribute models with varying sizes to clients based on their available resources, denoted as $R_i$. In layer splitting, the central server initiates a global model and splits it layer by layer, considering the client resources $R_i$, as depicted in Figure~\ref{fig:layer_split}. In contrast, stage splitting separates each stage layer by layer in Figure~\ref{fig:stage_split}. 
For instance, Figure~\ref{fig:stage_split} illustrates how the smallest client with $R_1$ resources obtains the first layer from each stage in stage splitting, whereas it acquires the first three layers from the entire model in layer splitting. 
Furthermore, hetero splitting, depicted in Figure~\ref{fig:hetero_split}, involves the server splitting the global model into distinct widths and depths for different clients, similar to the approaches in HeteroFL \citep{diao2021heterofl} and FedRolex \citep{alam2022fedrolex}. Layer splitting and stage splitting offer flexibility for extending model-homogeneous methods to system heterogeneity, while hetero splitting enables the handling of client models with varied widths and depths. Finally, the server aggregates client weights based on their original positions in the server models.

\subsection{Internal Cross-layer Gradients}
Deploying model splitting methods directly in FL leads to a significant decrease in client accuracy, as demonstrated in Table~\ref{tab:acc_train_loss_case}. 
However, based on the findings of the case study, we observe that gradients from shallow layers contribute to increasing the similarity among layers from different clients, and CKA similarity exhibits a  positive correlation with client accuracy. 
Therefore, we enhance the quality of gradients from deep layers by incorporating the utilization of cross-layer gradients.
More specifically, when a \textbf{server model} updates the deep layers, we combine and refine the gradients from these layers with the gradients from the shallower layers to obtain appropriately updated gradients. Figure~\ref{fig:cross_layer_grads} provides a visual representation of how cross-layer gradients are employed.
We assume that this stage has $N$ layers. The first layer with the same shape in a stage (block) is referred to as \textbf{Layer 0}, and its corresponding gradients at time step $t$ are $G_0^t$. For Layer $k$, where $k \in \{1, 2, ..., N\}$ within the same stage, the cross-layer gradients are given by $G_k^t + G_0^t$. Despite a large number of works on short-cut paths in neural networks, our method differs fundamentally in terms of the goals and the operations. We provide a thorough discussion in Appendix~\ref{sec:apd_comparisons_rc}.

\subsection{Gradients Divergence Alleviation}
\begin{figure}[!t]
\vskip -0.4in
\centering
\subfloat[Gradient divergence.]
{\includegraphics[width=0.3\columnwidth]{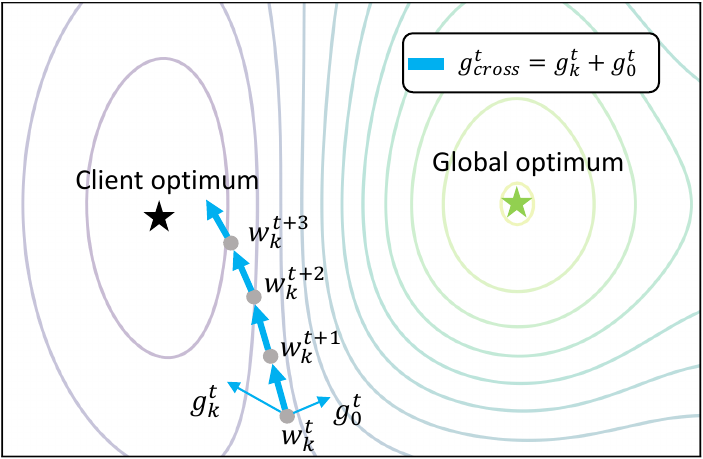}
\label{fig:cross_grads}}
\hfil
\subfloat[Normalized gradients.]
{\includegraphics[width=0.3\columnwidth]{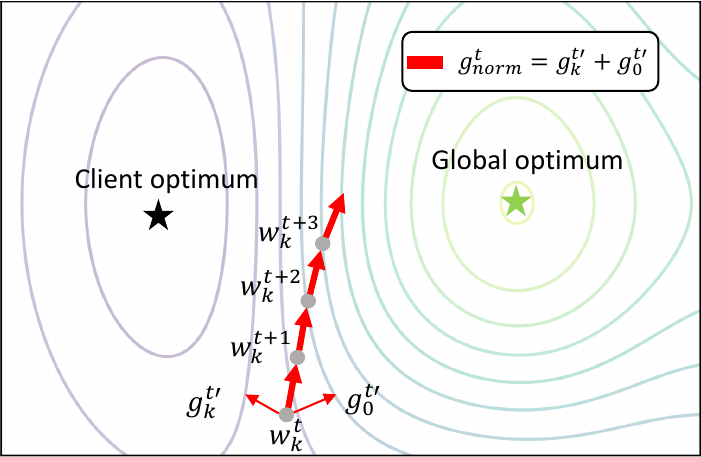}
\label{fig:cross_norm_grads}}
\hfil
\subfloat[Normalized and optimized.]
{\includegraphics[width=0.3\columnwidth]{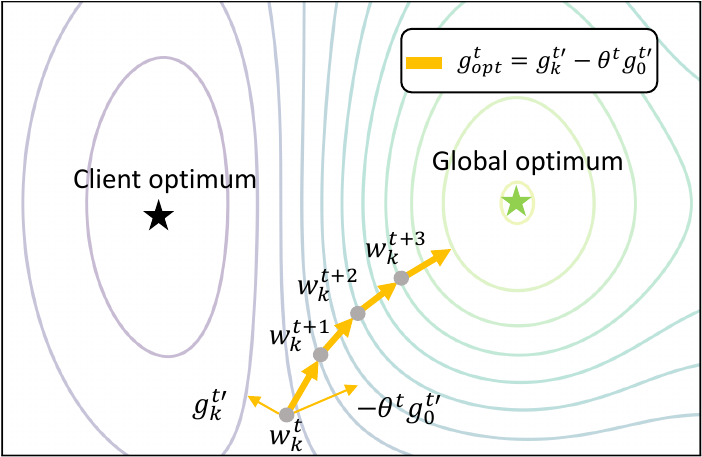}
\label{fig:cross_In_grads}}
\vskip -0.07in
\caption{A depiction of gradient divergence, as shown in Figure~\ref{fig:cross_grads}, along with its solutions. Despite the normalization portrayed in Figure~\ref{fig:cross_norm_grads}, the impact of gradient divergence persists. To mitigate this issue, we propose a convex optimization problem that is restricting gradient directions, as demonstrated in Figure~\ref{fig:cross_In_grads} and supported by Theorem~\ref{thm:vec_solution}.}
\label{fig:gradient_example}
\vskip -0.15in
\end{figure}

However, the direct utilization of cross-layer gradients leads to an acute issue known as gradient and weight divergence \citep{wang2020tackling, zhao2018federated}, as depicted in Figure~\ref{fig:cross_grads}. To counter this effect, we introduce gradient normalization (Figure~\ref{fig:cross_norm_grads}) and the proposed convex optimization problem to restrict gradient directions, as illustrated in Figure~\ref{fig:cross_In_grads}.

\textit{Cross-layer Gradients Normalization.}
Figure~\ref{fig:cross_norm_grads} depicts the benefits of utilizing normalized gradients. 
The normalized cross-layer gradient ${g_0^t}'+{g_k^t}'$ directs the model closer to the global optimum than the original cross-layer gradient ${g_0^t}+{g_k^t}$. 
In particular, our normalization approach emphasizes the norm of gradients, 
i.e., ${g_0^t}'=g_0^t/||g_0^t||$ and ${g_k^t}'=g_k^t/||g_k^t||$. The normalized cross-layer gradient is computed as $({g_0^t}'+{g_k^t}')\times({||g_0^t||+||g_k^t||})/2$ in practice.

\textit{Convex Optimization.} 
In addition to utilizing normalized gradients, incorporating novel projective gradients that leverage knowledge from both $g_0^t$ and $g_k^t$ serves to alleviate the detrimental impact of gradient divergence arising from the utilization of cross-layer gradients. Moreover, Our objective is to find the optimal projective gradients, denoted as $g_{opt}$, which strike a balance between being as close as possible to $g_k$ while maintaining alignment with $g_0$. This alignment ensures that $g_k$ is not hindered by the influence of $g_0$ while allowing $g_{opt}$ to acquire the beneficial knowledge for $g_k$ from $g_0$. In other words, we aim for $g_{opt}$ to capture the advantageous information contained within $g_0$ without impeding the progress of $g_k$. Pursuing this line of thought, we introduce a constraint aimed at ensuring the optimization directions of gradients, outlined as $\langle g_{0}^t,g_{k}^t \rangle \geq 0$,
where $\langle \cdot,\cdot \rangle$ is the dot product. To establish a convex optimization problem incorporating this constraint, we denote the projected gradient as $g_{opt}$ and formulate the following primal convex optimization problem,
\begin{equation}
\label{eq:primal_optim}
\begin{aligned}
 \min_{g_{opt}^t} && ||g_{k}^t-g_{opt}^t||_{2}^{2}, \ \ 
 s.t.\ \langle g_{opt}^t,g_{0}^t\rangle \geq 0,
\end{aligned}
\end{equation}
\vskip -0.15in
% \vspace{-0.1in}
where we preserve the optimization direction of $g_{0}^t$ in $g_{opt}^t$ while minimizing the distance between $g_{opt}^t$ and $g_{k}^t$. We prioritize the proximity of $g_{opt}^t$ to $g_{k}^t$ over $g_{0}^t$ since $g_{k}^t$ represents the true gradients of layer $k$. By solving this problem through Lagrange dual problem \citep{bot2009duality}, we derive the following outcomes,
\begin{theorem}
\label{thm:vec_solution}
(Divergence alleviation). If gradients are vectors, for the layers that require cross-layer gradients, their updated gradients can be expressed as,
\begin{equation}
\label{eq:analytic_sol}
    g_{opt}^t=
\begin{cases}
g_k^t,& \text{if } \beta\geq 0 \\
g_k^t-\theta^t g_0^t, & \text{if } \beta<0,
\end{cases}
\end{equation}
where $\theta^t=\frac{\beta}{\alpha}$, $\alpha=(g_0^t)^Tg_0^t$ and $\beta=(g_0^t)^Tg_k^t$.
\end{theorem}
\vskip -0.4in
\begin{remark}
This theorem can be extended to the matrix form.
\end{remark}
We provide proof for Theorem~\ref{thm:vec_solution} and demonstrate how matrix gradients are incorporated into the problem in Appendix~\ref{sec:apd_proof}. 
Our analytic solution in Equation~\ref{eq:analytic_sol} automatically determines the optimal settings for parameter $\theta^t$, eliminating the need for cumbersome manual adjustments.
In our practical implementation, we consistently update the server model using the expression $g_k^t-\theta^t g_0^t$, irrespective of whether $\beta \geq 0$ or $\beta < 0$. This procedure is illustrated in Algorithm~\ref{Algorithm:InCo_Aggregation} in Appendix~\ref{sec:apd_exp}.

% \vspace{-0.2in}
\textit{Communication Overheads.} According to the entire process, the primary process (internal cross-layer gradients) is conducted on the server. Therefore, our method does not impose any additional communication overhead between clients and the server.

% \vspace{-0.2in}
\section{Convergence Analysis}
In this section, we demonstrate the convergence of cross-layer gradients and propose the convergence rate in non-convex scenarios. To simplify the notations, we adopt $L_i$ to be the local objective. At first, we show the following assumptions frequently used in the convergence analysis for FL \citep{tan2022fedproto, li2018federated, karimireddy2020scaffold}.

\begin{assumption}
(Lipschitz Smooth). \textit{Each objective function $L_i$ is $L$-Lipschitz smooth and satisfies that $||\nabla L_i(x) - \nabla L_i(y)|| \leq L ||x-y||, \forall (x,y) \in D_i, i \in {1, ..., K}$}.
\label{ass:L-smooth}
\end{assumption}
\begin{assumption}
(Unbiased Gradient and Bounded Variance). \textit{At each client, the stochastic gradient is an unbiased estimation of the local gradient, with $\mathbb{E}[g_i(x)])=\nabla L_i(x)$, and its variance is bounded by $\sigma^2$, meaning that $\mathbb{E}[||g_i(x)-\nabla L_i(x)||^2]\leq \sigma^2, \forall i \in {1, ..., K}$, where $\sigma^2 \geq 0.$}
\label{ass:unbiased_gradients}
\end{assumption}
\begin{assumption}
(Bounded Expectation of Stochastic Gradients). \textit{The expectation of the norm of the stochastic gradient at each client is bounded by $\rho$, meaning that $\mathbb{E}[||g_i(x)||]\leq \rho, \forall i \in {1, ..., K}$.}
\label{ass:Bounded_gradients}
\end{assumption}
\begin{assumption}
(Bounded Covariance of Stochastic Gradients). \textit{The covariance of the stochastic gradients is bounded by $\Gamma$, meaning that $Cov(g_{i,l_k}, g_{i, l_j})\leq \Gamma, \forall i \in {1, ..., K}$, where $l_k, l_j$ are the layers belonging to a model at client $i$.}
\label{ass:Bounded_covar}
\end{assumption}
Following these assumptions, we present proof of non-convex convergence concerning the utilization of cross-layer gradients in Federated Learning (FL). We outline our principal theorems as follows.
\begin{theorem}
(Per round drift). Supposed Assumption~\ref{ass:L-smooth} to Assumption~\ref{ass:Bounded_covar} are satisfied, the loss function of an arbitrary client at round $t+1$ is bounded by,
\begin{equation}
% \nonumber
\small
\label{eq:per_round_drift}
\begin{aligned}
\mathbb{E}[L_{t+1,0}]\leq \mathbb{E}[L_{t,0}]-(\eta-\frac{L\eta^2}{2})\sum_{e=0}^{E-1}||\nabla L_{t,e}||^2 + 
\frac{LE\eta^2}{2}\sigma^2 + 2\eta(\Gamma+\rho^2)+L\eta^2(2\rho^2+\sigma^2+\Gamma).
\end{aligned}
\end{equation}
\vskip -0.15in
\label{thm:per_round_drift}
\end{theorem}
The Theorem~\ref{thm:per_round_drift} demonstrates the bound of the local objective function after every communication round. Non-convex convergence can be guaranteed by the appropriate $\eta$.
\begin{theorem}
(Non-convex convergence). The loss function L is monotonously decreased with the increasing communication round when,
\begin{equation}
% \nonumber
\small
\label{eq:non_convex_convergence}
\begin{aligned}
\eta < \frac{2\sum_{e=0}^{E-1}||\nabla L_{t,e}||^2-4(\Gamma+\rho^2)}{L(\sum_{e=0}^{E-1}||\nabla L_{t,e}||^2+E\rho^2+2(2\rho^2+\sigma^2+\Gamma)}.
\end{aligned}
\end{equation}
\vskip -0.15in
\label{thm:non_convex_convergence}
\end{theorem}
Moreover, after we prove the non-convex convergence for the cross-layer gradients, the non-convex convergence rate is described as follows.
\begin{theorem}
(Non-convex convergence rate). Supposed Assumption~\ref{ass:L-smooth} to Assumption~\ref{ass:Bounded_covar} are satisfied and $\kappa=L_0-L^*$, for an arbitrary client, given any $\epsilon > 0$, after
\begin{equation}
% \nonumber
\small
\label{eq:convergence_rate}
\begin{aligned}
T=\frac{2\kappa}{E\eta((2-L\eta)\epsilon-3L\eta\sigma^2-
2(2+L\eta)\Gamma-4(1+L\eta)\rho^2)}
\end{aligned}
\end{equation}
communication rounds, we have
\begin{equation}
% \nonumber
\small
\label{eq:convergence_rate_1}
\begin{aligned}
\frac{1}{TE}\sum_{t=0}^{T-1}\sum_{e=0}^{E-1}\mathbb{E}[||\nabla L_{t,e}||^2] \leq \epsilon,\ \textit{if}\ \ \ \eta < \frac{2\epsilon-4(\Gamma+\rho^2)}{L(\epsilon+E\rho^2+2(2\rho^2+\sigma^2+\Gamma))}.
\end{aligned}
\end{equation}
\vskip -0.2in
\label{thm:Non_convex_rate}
\end{theorem}
Following these theorems, the convergence of internal cross-layer gradients is guaranteed. The proof is presented in Appendix~\ref{subsec:apd_convergence_proof}.

\section{Experiments}
In this section, we conduct comprehensive experiments aimed at demonstrating three fundamental aspects: (1) the efficacy of InCo Aggregation and its extensions for various FL methods (Section~\ref{sec:Handling}), (2) the robustness analysis and ablation study of InCo Aggregation (Section~\ref{sec:ablation}), (3) in-depth analyses of the underlying principles behind InCo Aggregation (Section~\ref{sec:Reasons}). Our codes are released on GitHub \footnote{\href{https://github.com/ChanYunHin/InCo-Aggregation}{https://github.com/ChanYunHin/InCo-Aggregation}}. 
More experimental details and results can be found in Appendix~\ref{sec:apd_exp}.

\newcommand{\mycc}{\cellcolor{Gray}}

\begin{table*}[!t]
\vskip -0.2in
\def\arraystretch{1.2}
% \scalebox{}
\tiny
    \caption{Test accuracy of model-homogeneous methods with 100 clients and sample ratio 0.1. We shade in gray the methods that are combined with our proposed method, InCo Aggregation. We bold the best results and denote the improvements compared to the original methods in red. See Appendix \ref{error_bars} for the error bars of InCo methods.}
    % \vskip -0.1in
    \label{tab:acc_100sr01}
    \centering
    \begin{tabular}{cccccccccc}
    % \hline
    \toprule
         \multirow{2}*{Base} & \multirow{2}*{Methods} & \multicolumn{2}{c}{Fashion-MNIST} & \multicolumn{2}{c}{SVHN} & \multicolumn{2}{c}{CIFAR10} &  \multicolumn{2}{c}{CINIC10} \\
        \cline{3-10}
         & & $\alpha=0.5$ & $\alpha=1.0$ & $\alpha=0.5$ & $\alpha=1.0$ &  $\alpha=0.5$ & $\alpha=1.0$ & $\alpha=0.5$ & $\alpha=1.0$ \\
      % \hline
      \midrule
      \multirow{10}{*}{\rotatebox{90}{ResNet (Stage splitting)}}

        & HeteroAvg & 87.8$\pm$1.1 & 86.0$\pm$1.0 & 85.1$\pm$2.0 & 86.9$\pm$2.3 & 64.8$\pm$2.9 & 66.7$\pm$3.3 & 48.6$\pm$2.6 & 56.5$\pm$1.6 \\
        & HeteroProx & 86.8$\pm$1.5 & 83.9$\pm$1.8 & 87.8$\pm$2.1 & 89.9$\pm$1.7 & 72.5$\pm$2.1 & 73.1$\pm$1.9 & 56.4$\pm$2.0 & 60.9$\pm$1.8\\
        & HeteroScaffold & 85.2$\pm$0.8 & 86.4$\pm$0.7 & 80.6$\pm$2.3 & 86.3$\pm$2.7 & 65.5$\pm$3.0 & 69.7$\pm$2.8 & 50.8$\pm$2.9 & 57.8$\pm$3.4 \\
        & HeteroNova & 84.9$\pm$1.3 & 86.7$\pm$1.1 & 84.4$\pm$1.4 & 88.0$\pm$1.7 & 60.1$\pm$3.7 & 68.0$\pm$3.5 & 46.1$\pm$2.3 & 52.1$\pm$2.2 \\
        & HeteroMOON & 87.9$\pm$0.4 & 88.3$\pm$0.3 & 83.0$\pm$2.3 & 86.5$\pm$1.6 & 65.1$\pm$2.9 & 68.4$\pm$2.6 & 50.1$\pm$2.3 & 54.7$\pm$ 1.8\\
        \cline{2-10}
        
        & \mycc InCoAvg &  \mycc \textbf{90.2}({\color{red}{$\uparrow$2.4}}) & \mycc 88.4({\color{red}{$\uparrow$2.4}}) & \mycc 87.6({\color{red}{$\uparrow$2.5}}) & \mycc 89.0({\color{red}{$\uparrow$2.1}}) & \mycc 67.8({\color{red}{$\uparrow$3.0}}) & \mycc 70.7({\color{red}{$\uparrow$4.0}}) & \mycc 53.0({\color{red}{$\uparrow$4.4}}) & \mycc 57.5({\color{red}{$\uparrow$1.0}}) \\
        
        & \mycc InCoProx & \mycc 88.8({\color{red}{$\uparrow$2.0}}) & \mycc 86.4({\color{red}{$\uparrow$2.5}}) & \mycc \textbf{89.0}({\color{red}{$\uparrow$1.2}}) & \mycc \textbf{90.8}({\color{red}{$\uparrow$0.9}}) & \mycc \textbf{74.5}({\color{red}{$\uparrow$2.0}}) & \mycc \textbf{76.8}({\color{red}{$\uparrow$3.7}}) & \mycc \textbf{59.1}({\color{red}{$\uparrow$2.7}}) & \mycc \textbf{62.5}({\color{red}{$\uparrow$1.6}}) \\

        & \mycc InCoScaffold & \mycc 88.3({\color{red}{$\uparrow$3.1}}) & \mycc \textbf{90.1}({\color{red}{$\uparrow$3.7}}) & \mycc 85.4({\color{red}{$\uparrow$4.8}}) & \mycc 87.8({\color{red}{$\uparrow$1.5}}) & \mycc 67.3({\color{red}{$\uparrow$1.8}}) & \mycc 73.8({\color{red}{$\uparrow$4.1}}) & \mycc 53.5({\color{red}{$\uparrow$2.7}}) & \mycc 61.7({\color{red}{$\uparrow$3.9}}) \\

        & \mycc InCoNova & \mycc 86.6({\color{red}{$\uparrow$1.7}}) & \mycc 87.4({\color{red}{$\uparrow$0.7}}) & \mycc 86.4({\color{red}{$\uparrow$2.0}}) & \mycc 88.4({\color{red}{$\uparrow$0.4}}) & \mycc 62.8({\color{red}{$\uparrow$2.7}}) & \mycc 69.7({\color{red}{$\uparrow$2.7}}) & \mycc 48.0({\color{red}{$\uparrow$1.9}}) & \mycc 54.1({\color{red}{$\uparrow$2.0}})\\

       % \hdashline[1pt/5pt]

       & \mycc InCoMOON & \mycc 89.1({\color{red}{$\uparrow$1.2}}) & \mycc 89.5({\color{red}{$\uparrow$1.2}}) & \mycc 85.6({\color{red}{$\uparrow$2.6}}) & \mycc 89.3({\color{red}{$\uparrow$2.8}}) & \mycc 68.2({\color{red}{$\uparrow$3.1}}) & \mycc 71.8({\color{red}{$\uparrow$3.4}}) & \mycc 54.3({\color{red}{$\uparrow$4.2}}) & \mycc 57.6({\color{red}{$\uparrow$2.9}})\\
        \hline
        
        \multirow{10}{*}{\rotatebox{90}{ViT (Layer splitting)}}  
        & HeteroAvg & 92.2$\pm$0.6 & 92.0$\pm$0.6 & 92.9$\pm$1.0 & 93.8$\pm$0.9 & 93.6$\pm$1.0 & 94.1$\pm$0.9 & 84.2$\pm$1.6 & 85.3$\pm$1.3 \\
        & HeteroProx & 90.9$\pm$0.8 & 91.7$\pm$0.6 & 91.2$\pm$1.3 & 92.4$\pm$1.8 & 92.0$\pm$1.5 & 92.6$\pm$1.3 & 84.0$\pm$1.8 & 84.8$\pm$2.0 \\
        & HeteroScaffold & 91.9$\pm$0.6 & 92.1$\pm$0.4 & 92.5$\pm$0.9 & 93.7$\pm$0.6 & 93.8$\pm$0.8 & 94.3$\pm$0.4 & 83.8$\pm$1.9 & 85.3$\pm$1.6\\
        & HeteroNova & 92.1$\pm$0.9 & 92.4$\pm$0.4 & 92.3$\pm$1.0 & 94.1$\pm$1.2 & 93.6$\pm$0.5 & 94.5$\pm$0.6 & 85.3$\pm$1.7 & 86.7$\pm$1.5 \\
        & HeteroMOON & 92.0$\pm$0.4 & 92.3$\pm$0.3 & 92.7$\pm$1.1 & 94.0$\pm$0.9 & 93.5$\pm$0.8 & 94.6$\pm$0.5& 84.7$\pm$1.4 & 85.6$\pm$1.4 \\
        \cline{2-10}
       & \mycc InCoAvg & \mycc 93.0({\color{red}{$\uparrow$0.8}}) & \mycc 93.1({\color{red}{$\uparrow$1.1}})  & \mycc 94.2({\color{red}{$\uparrow$1.3}}) & \mycc 95.0({\color{red}{$\uparrow$1.2}}) & \mycc 94.6({\color{red}{$\uparrow$1.0}}) & \mycc 95.0({\color{red}{$\uparrow$0.9}}) & \mycc 85.9({\color{red}{$\uparrow$1.7}}) & \mycc 86.8({\color{red}{$\uparrow$1.5}}) \\
       
        & \mycc InCoProx & \mycc 92.6({\color{red}{$\uparrow$1.7}}) & \mycc 92.5({\color{red}{$\uparrow$0.8}}) & \mycc 93.9({\color{red}{$\uparrow$2.7}}) & \mycc 94.4({\color{red}{$\uparrow$2.0}}) & \mycc 94.0({\color{red}{$\uparrow$2.0}}) & \mycc 94.8({\color{red}{$\uparrow$2.2}}) & \mycc 85.1 ({\color{red}{$\uparrow$1.1}}) & \mycc 86.0({\color{red}{$\uparrow$1.2}})\\

        & \mycc InCoScaffold & \mycc 92.9({\color{red}{$\uparrow$1.0}}) & \mycc 93.0({\color{red}{$\uparrow$0.9}}) & \mycc 94.0({\color{red}{$\uparrow$1.5}}) & \mycc 94.8({\color{red}{$\uparrow$1.1}}) & \mycc 94.6({\color{red}{$\uparrow$0.8}}) & \mycc 95.0({\color{red}{$\uparrow$0.7}}) & \mycc 85.7({\color{red}{$\uparrow$1.9}}) & \mycc 86.5({\color{red}{$\uparrow$1.2}})\\

        & \mycc InCoNova & \mycc \textbf{93.1}({\color{red}{$\uparrow$1.0}}) & \mycc \textbf{93.6}({\color{red}{$\uparrow$1.2}}) & \mycc 94.7({\color{red}{$\uparrow$2.4}}) & \mycc \textbf{95.6}({\color{red}{$\uparrow$1.5}}) & \mycc \textbf{94.8}({\color{red}{$\uparrow$1.2}}) & \mycc \textbf{95.7}({\color{red}{$\uparrow$1.2}}) & \mycc \textbf{86.2}({\color{red}{$\uparrow$0.9}}) & \mycc \textbf{88.2}({\color{red}{$\uparrow$1.2}}) \\

       % \hdashline[1pt/5pt]

       & \mycc InCoMOON & \mycc 92.8({\color{red}{$\uparrow$0.8}}) & \mycc 93.0({\color{red}{$\uparrow$0.7}}) & \mycc \textbf{94.7}({\color{red}{$\uparrow$2.0}}) & \mycc 95.1({\color{red}{$\uparrow$1.1}}) & \mycc 94.2({\color{red}{$\uparrow$0.7}}) & \mycc 95.1 ({\color{red}{$\uparrow$0.5}}) & \mycc 86.0({\color{red}{$\uparrow$1.3}}) & \mycc 86.8({\color{red}{$\uparrow$1.2}})\\

    % \hline
    % \hline
    \bottomrule
    \end{tabular}
\vskip -0.2in
\end{table*}

\subsection{Experiment Setup}
\textit{Dataset and Data Distribution.} 
We conduct experiments on Fashion-MNIST \citep{xiao2017/online}, SVHN \citep{netzer2011reading}, CIFAR-10 \citep{krizhevsky2009learning} and CINIC-10 \citep{darlow2018cinic} under non-iid settings. 
We evaluate the algorithms under two Dirichlet distributions with $\alpha=0.5$ and $\alpha=1.0$ for all datasets.

\textit{Baselines.} 
To demonstrate the effectiveness of InCo Aggregation, we use five baselines in model-homogeneous FL: \textbf{FedAvg} \citep{mcmahan2017communication}, \textbf{FedProx} \citep{li2018federated}, \textbf{FedNova} \citep{wang2020tackling}, \textbf{Scaffold} \citep{karimireddy2020scaffold}, and \textbf{MOON} \citep{li2021model} for ResNets and ViTs. In the context of model heterogeneity, we extend the training procedures of these baselines by incorporating model splitting methods, denoting the modified versions with the prefix "\textbf{Hetero}". Furthermore, by incorporating these methods with InCo Aggregation, we prefix the names with "\textbf{InCo}".
Moreover, we also extend our methods to four state-of-the-art methods in model-heterogeneous FL: \textbf{HeteroFL}\citep{diao2021heterofl}, \textbf{InclusiveFL}\citep{liu2022no}, \textbf{FedRolex}\citep{alam2022fedrolex} and \textbf{ScaleFL}\citep{ilhan2023scalefl} for ResNets. We take the average accuracy of three different random seeds.

\textit{Federated Settings.} 
In heterogeneous FL, we consider two architectures, ResNets and ViTs. The largest models are ResNet26 and ViT-S/12 (ViT-S with 12 layers). We deploy stage splitting for ResNets and obtain five sub-models, which can be recognized as ResNet10, ResNet14, ResNet18, ResNet22, and ResNet26. For the pre-trained ViT models, we employ layer splitting and result in five sub-models, which are ViT-S/8, ViT-S/9, ViT-S/10, ViT-S/11, and ViT-S/12. Moreover, we consider five different model capacities $\beta=\{1, \sfrac{1}{2}, \sfrac{1}{4}, \sfrac{1}{8}, \sfrac{1}{16}\}$ in hetero splitting, where for instance, $\sfrac{1}{2}$ indicates the widths and depths are half of the largest model ResNet26. Our experimental setup involves 100 clients, categorized into five distinct groups, with a sample ratio of 0.1. The detailed model sizes are shown in Appendix~\ref{apd:model_sizes}.

\subsection{InCo Aggregation Improves All Baselines.}
\label{sec:Handling}
Table~\ref{tab:acc_100sr01} and Table~\ref{tab:acc_100sr01_hetero} present the test accuracy of 100 clients with a sample ratio of 0.1. 
Table~\ref{tab:acc_100sr01} provides compelling evidence for the efficacy of InCo Aggregation in enhancing the performance of all model-homogeneous baselines. Table~\ref{tab:acc_100sr01_hetero} demonstrates the improvements of deploying InCo Aggregation in the model-heterogeneous methods. 
Moreover, Table~\ref{tab:acc_100sr01_hetero} highlights that InCo Aggregation introduces no additional communication overhead and only incurs 0.4M FLOPs, which are conducted on the server side, indicating that InCo Aggregation does not impose any burden on client communication and computation resources.

\begin{table*}[!t]
\vskip -0.2in
\def\arraystretch{1.2}
% \scalebox{}
\tiny
    \caption{Test accuracy of model-heterogeneity methods with 100 clients and sample ratio 0.1. We shade in gray the methods that are combined with our proposed method, InCo Aggregation. We denote the improvements compared to the original methods in red. See Appendix \ref{error_bars} for the error bars of InCo methods.}
    \label{tab:acc_100sr01_hetero}
    \centering
    \begin{tabular}{ccccccccccc}
    % \hline
    \toprule
         \multirow{2}*{Base} & \multirow{2}*{Splitting} & \multirow{2}*{Methods} & \multicolumn{2}{c}{Fashion-MNIST} & \multicolumn{2}{c}{SVHN} & \multicolumn{2}{c}{CIFAR10} & Comm. & \multirow{2}*{FLOPs}\\
        \cline{4-9}
         & & & $\alpha=0.5$ & $\alpha=1.0$ & $\alpha=0.5$ & $\alpha=1.0$ &  $\alpha=0.5$ & $\alpha=1.0$ & overheads & \\
      % \hline
      \midrule
      \multirow{10}{*}{\rotatebox{90}{ResNet}}  

        & \multirow{2}*{Hetero} & HeteroFL & 88.9$\pm$1.0 & 89.7$\pm$0.7 & 90.5$\pm$1.6 & 92.2$\pm$1.3 & 65.2$\pm$3.2 & 68.4$\pm$3.6 & 4.6M & 33.4M \\

        & & \mycc +InCo &  \mycc 90.0({\color{red}{$\uparrow$1.1}}) & \mycc 90.4({\color{red}{$\uparrow$0.7}}) & \mycc 92.1({\color{red}{$\uparrow$1.6}}) & \mycc 93.5({\color{red}{$\uparrow$1.3}}) & \mycc 68.2({\color{red}{$\uparrow$3.0}}) & \mycc 71.2({\color{red}{$\uparrow$2.8}}) & \mycc 4.6M & \mycc 33.8M \\

        % \cline{2-9}
        
        & \multirow{2}*{Stage} & InclusiveFL & 89.1$\pm$1.1 & 89.8$\pm$1.0 & 88.6$\pm$2.0 & 90.0$\pm$2.2 & 65.7$\pm$3.5 & 68.4$\pm$3.3 & 12.3M & 75.2M  \\

        &  & \mycc +InCo & \mycc 90.1({\color{red}{$\uparrow$1.0}}) & \mycc 90.5({\color{red}{$\uparrow$0.7}}) & \mycc 90.6({\color{red}{$\uparrow$2.0}}) & \mycc 90.9({\color{red}{$\uparrow$0.9}}) & \mycc 69.1({\color{red}{$\uparrow$3.4}}) & \mycc 72.3({\color{red}{$\uparrow$3.9}}) & \mycc 12.3M & \mycc 75.6M \\

        % \cline{2-9}
        
        & \multirow{2}*{Hetero} & FedRolex & 88.2$\pm$1.0 & 90.2$\pm$0.8 & 90.9$\pm$1.3 & 91.6$\pm$1.7 & 64.7$\pm$4.1 & 72.3$\pm$3.0 & 4.6M & 33.4M \\

        & & \mycc +InCo & \mycc 90.4({\color{red}{$\uparrow$2.2}}) & \mycc 91.3({\color{red}{$\uparrow$1.1}}) & \mycc 92.8({\color{red}{$\uparrow$1.9}}) & \mycc 93.4({\color{red}{$\uparrow$1.8}}) & \mycc 67.9({\color{red}{$\uparrow$3.2}}) & \mycc 75.6({\color{red}{$\uparrow$3.3}}) & \mycc 4.6M & \mycc 33.8M\\

        % \cline{2-9}
        
        & \multirow{2}*{Hetero} & ScaleFL & 90.9$\pm$0.5 & 91.0$\pm$0.4 & 92.6$\pm$1.0 & 92.9$\pm$0.9 & 71.1$\pm$2.9 & 74.7$\pm$3.1 & 9.5M & 51.9M\\

        & & \mycc +InCo & \mycc 91.5({\color{red}{$\uparrow$0.6}}) & \mycc 91.7({\color{red}{$\uparrow$0.7}}) & \mycc 93.4({\color{red}{$\uparrow$0.8}}) & \mycc 93.6({\color{red}{$\uparrow$0.7}}) & \mycc 73.8({\color{red}{$\uparrow$2.7}}) & \mycc 76.1({\color{red}{$\uparrow$2.4}}) & \mycc 9.5M & \mycc 52.3M\\

        \cline{2-11}

        & N/A & AllSmall & 83.5$\pm$1.7 & 84.0$\pm$1.7 & 72.1$\pm$3.5 & 81.0$\pm$2.9 & 39.2$\pm$2.0 & 44.9$\pm$2.3 & 0.07M & 3.7M\\
        
        & N/A & AllLarge & 91.8$\pm$0.5 & 92.5$\pm$0.8 & 93.4$\pm$0.8 & 93.8$\pm$0.5 & 79.6$\pm$2.9 & 82.5$\pm$1.0 & 17.5M & 112.4M\\
       
    % \hline
    % \hline
    \bottomrule
    \end{tabular}
\vskip -0.1in
\end{table*}

\subsection{Robustness Analysis and Ablation Study.}
\label{sec:ablation}
We delve into the robustness analysis of InCo Aggregation, examining two aspects: the impact of varying batch sizes and noise perturbations on gradients during transmission. Additionally, we perform an ablation study for InCo Aggregation. 
We provide more experiments in Appendix~\ref{sec:apd_exp}.

\begin{figure}[!t]
\vskip -0.15in
\centering
\subfloat[Different batch sizes in CIFAR-10.]
{\includegraphics[width=0.23\columnwidth]{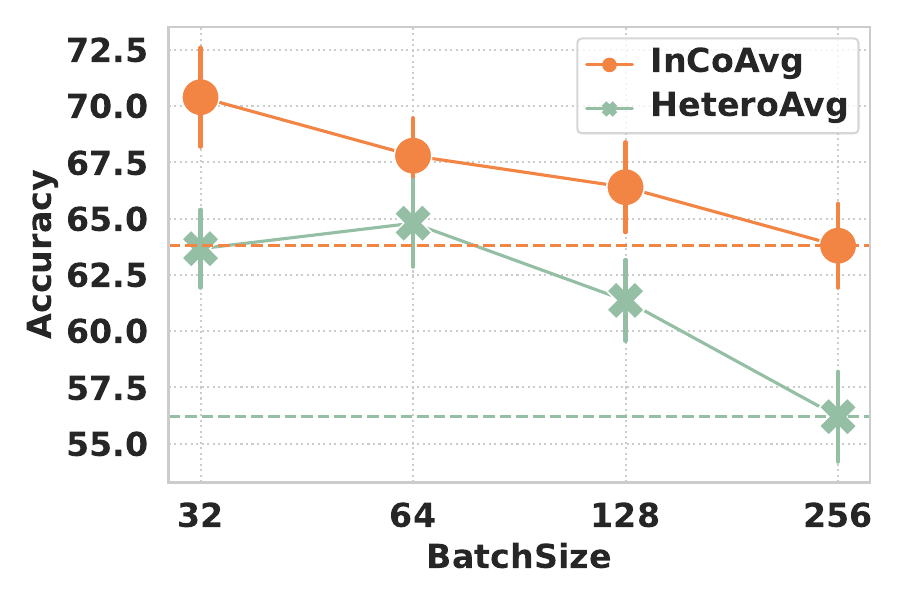}
\label{fig:batchsize_resnet_cifar}}
\hfil
\subfloat[Different batch sizes in CINIC-10.]
{\includegraphics[width=0.23\columnwidth]{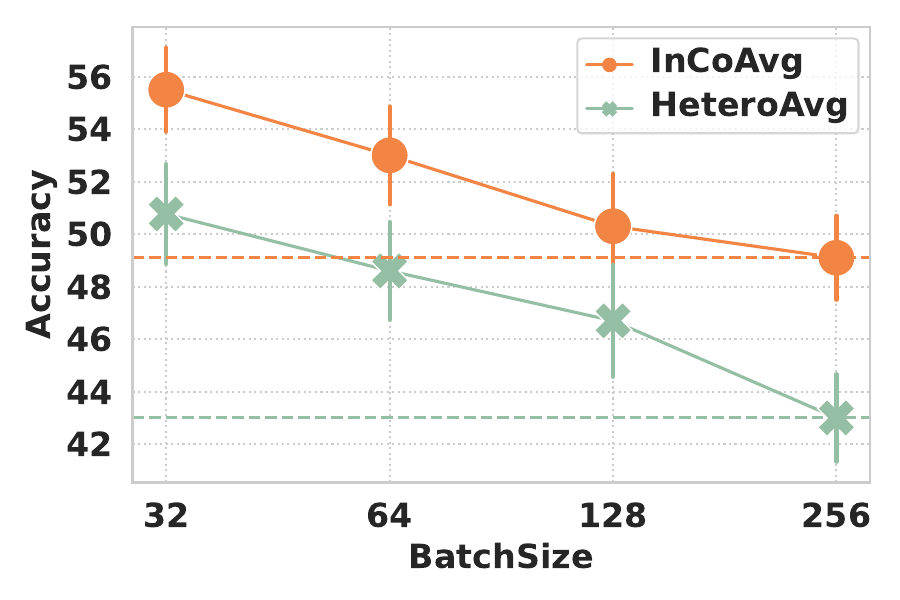}
\label{fig:batchsize_resnet_cinic}}
\hfil
\subfloat[Different noise perturbations in CIFAR-10.]
{\includegraphics[width=0.23\columnwidth]{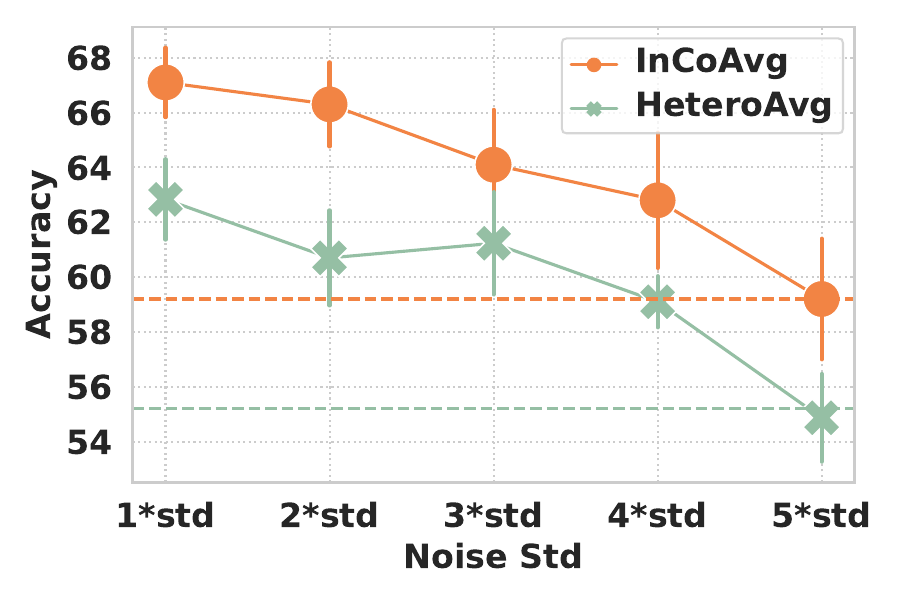}
\label{fig:noise_resnet_cifar}}
\hfil
\subfloat[Different noise perturbations in CINIC-10.]
{\includegraphics[width=0.23\columnwidth]{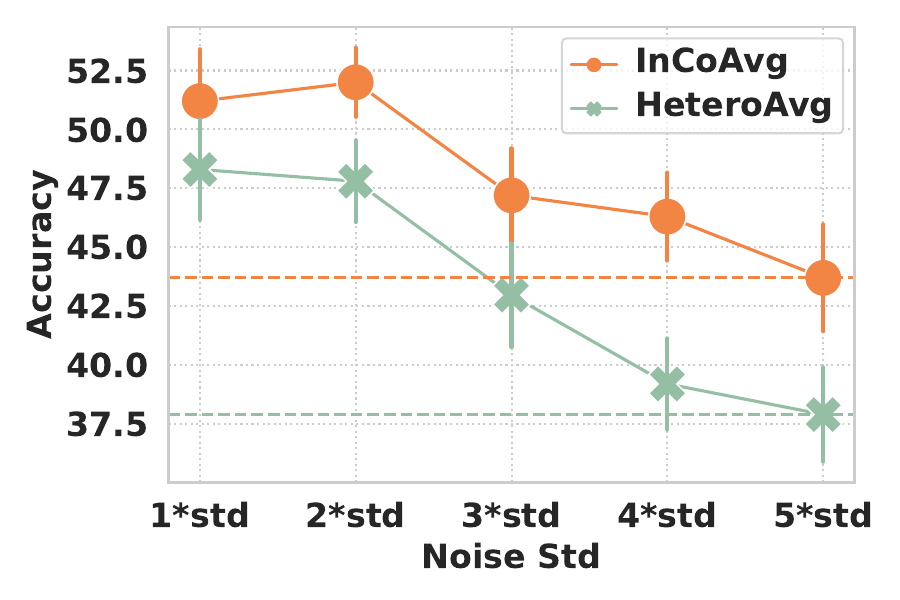}
\label{fig:noise_resnet_cinic}}
\vskip -0.07in
\caption{Robustness analysis for InCo Aggregation.}
\label{fig:robustness_resnet}
\vskip -0.1in
\end{figure}

\textit{Effect of Batch Size and Noise Perturbation.} 
Notably, when compared to FedAvg as depicted in Figure~\ref{fig:batchsize_resnet_cifar} and Figure~\ref{fig:batchsize_resnet_cinic}, our method exhibits significant improvements while maintaining comparable performance across all settings. 
Furthermore, as illustrated in Figure~\ref{fig:noise_resnet_cifar} and Figure~\ref{fig:noise_resnet_cinic}, we explore the impact of noise perturbations by simulating noise with standard deviations following the gradients.

\begin{figure}[!t]
\vskip -0.15in
\centering
\subfloat[Fashion-MNIST.]
{\includegraphics[width=0.23\columnwidth]{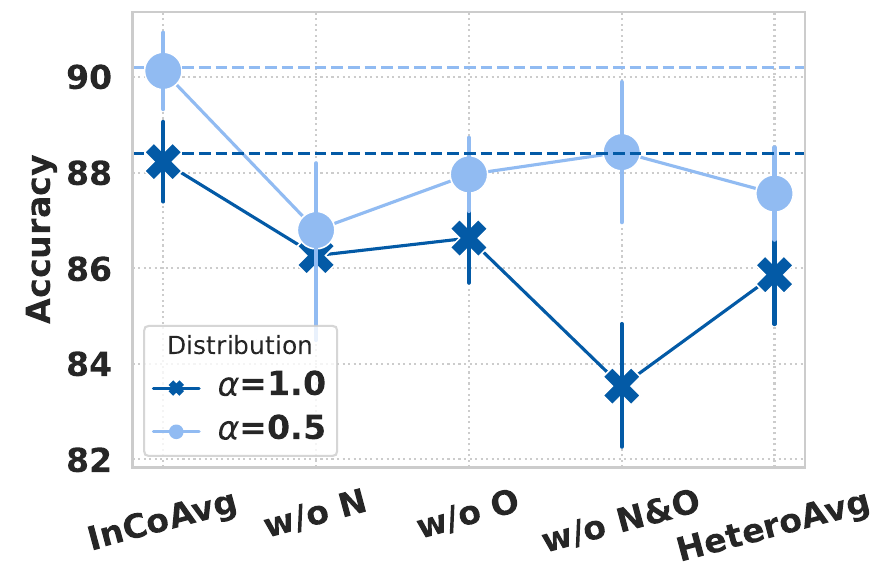}
\label{fig:ablation_resnet_fashion_mnist}}
\hfil
\subfloat[SVHN.]
{\includegraphics[width=0.23\columnwidth]{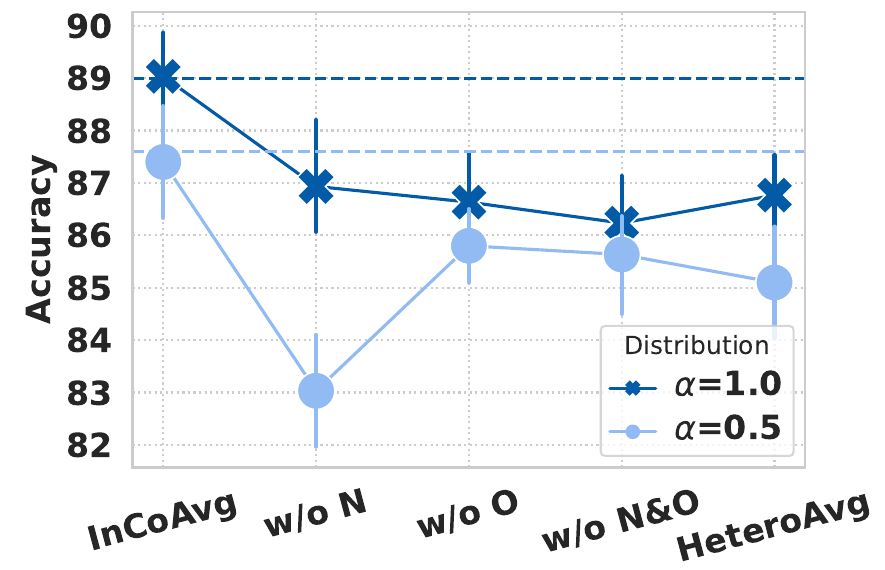}
\label{fig:ablation_resnet_svhn}}
\hfil
\subfloat[CIFAR-10.]
{\includegraphics[width=0.23\columnwidth]{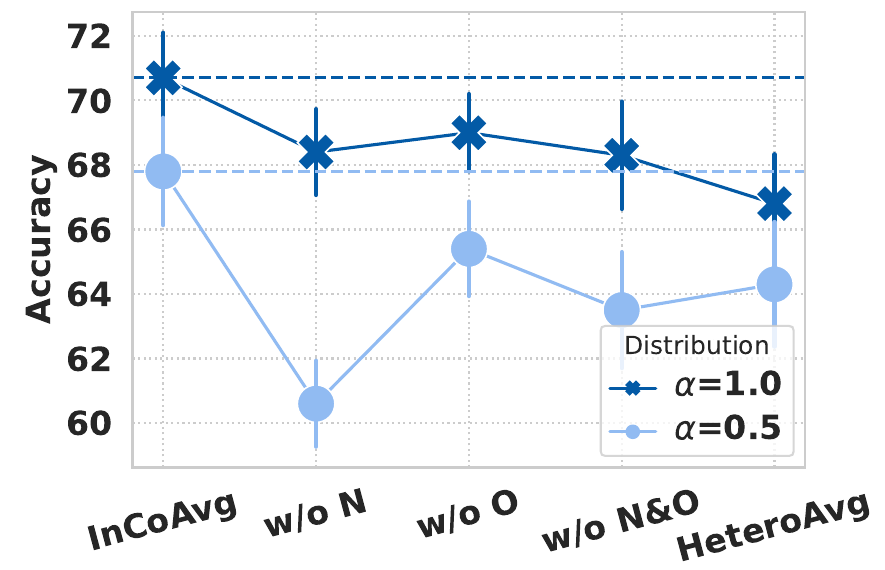}
\label{fig:ablation_resnet_cifar}}
\hfil
\subfloat[CINIC-10.]
{\includegraphics[width=0.23\columnwidth]{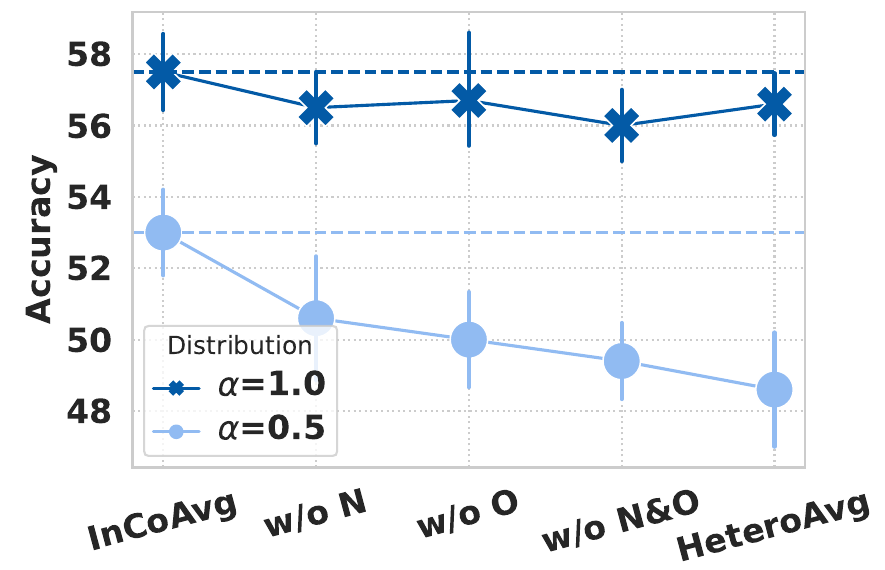}
\label{fig:ablation_resnet_cinic}}
\vskip -0.07in
\caption{Ablation studies for InCo Aggregation. The federated settings are the same as Table~\ref{tab:acc_100sr01}.}
\label{fig:ablation_resnet}
\vskip -0.2in
\end{figure}

\textit{Ablation Study.}
Our ablation study includes the following methods: (i) InCoAvg w/o Normalization (HeteroAvg with cross-layer gradients and optimization), (ii) InCoAvg w/o Optimization (HeteroAvg with normalized cross-layer gradients), (iii) InCoAvg w/o Normalization and Optimization (HeteroAvg with cross-layer gradients), and (iv) HeteroAvg (FedAvg with stage splitting). 
The ablation study of InCo Aggregation is depicted in Figure~\ref{fig:ablation_resnet}, demonstrating the efficiency of InCo Aggregation.

\subsection{The Reasons for the Improvements}
\label{sec:Reasons} 
We undertake a comprehensive analysis to gain deeper insights into the mechanisms underlying the efficacy of InCo Aggregation. Our analysis focuses on the following three key aspects:
(1) The investigation of important coefficients $\theta$ and $\beta$ in Theorem~\ref{thm:vec_solution}. (2) An examination of the feature spaces generated by different methods. (3) The evaluation of CKA similarity across various layers. 
Moreover, we discuss the differences between adding noises and InCo gradients in Appendix~\ref{apd:differences_noise_inco}.

\begin{figure}[!t]
\vskip -0.4in
\centering
\subfloat[$\theta$ in all layers.] {
    \includegraphics[width=0.15\columnwidth]{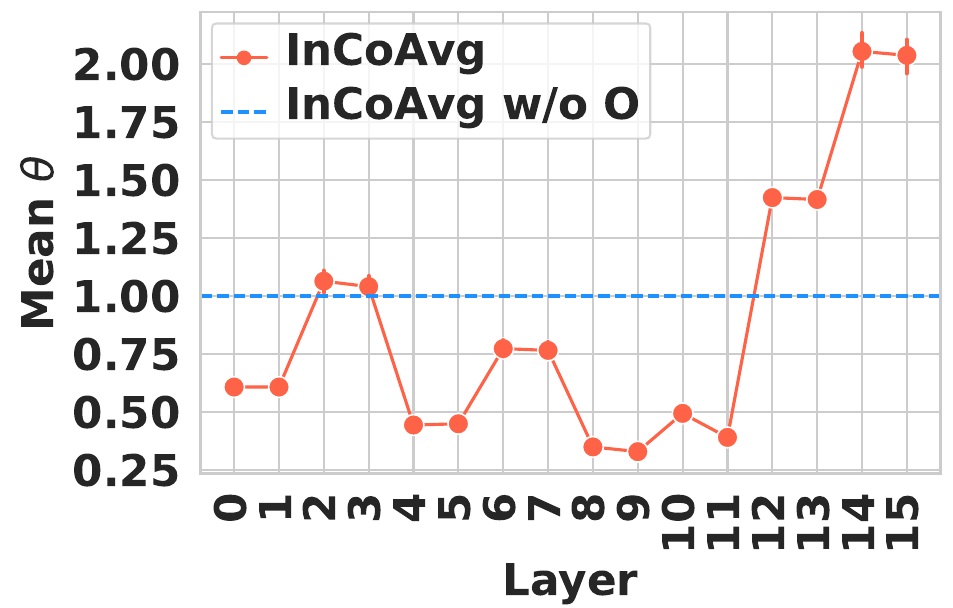}
    \label{fig:exp_theta_val}
}
\hfil
\subfloat[$\beta$ in Layer 11.] {
    \includegraphics[width=0.15\columnwidth]{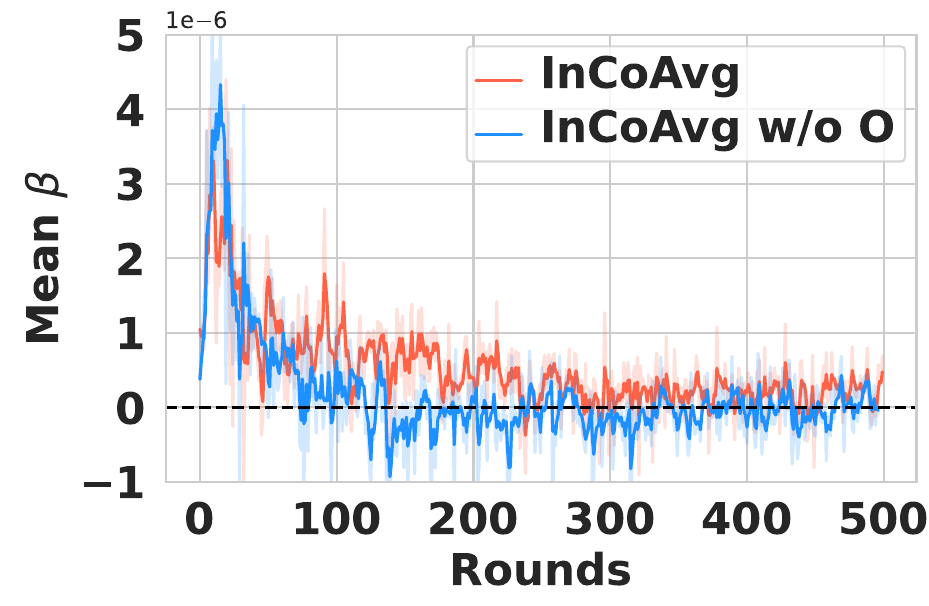}
    \label{fig:exp_beta_val_11}
}
\hfil
\subfloat[$\beta$ in Layer 13.] {
    \includegraphics[width=0.15\columnwidth]{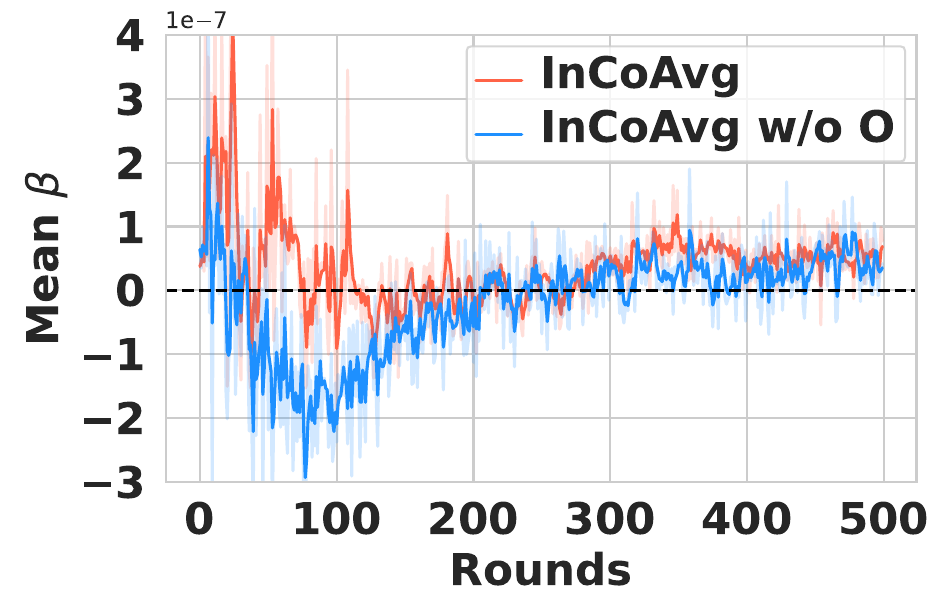}
    \label{fig:exp_beta_val_13}
}
\hfil
\subfloat[FedAvg.]
{\includegraphics[width=0.125\columnwidth]{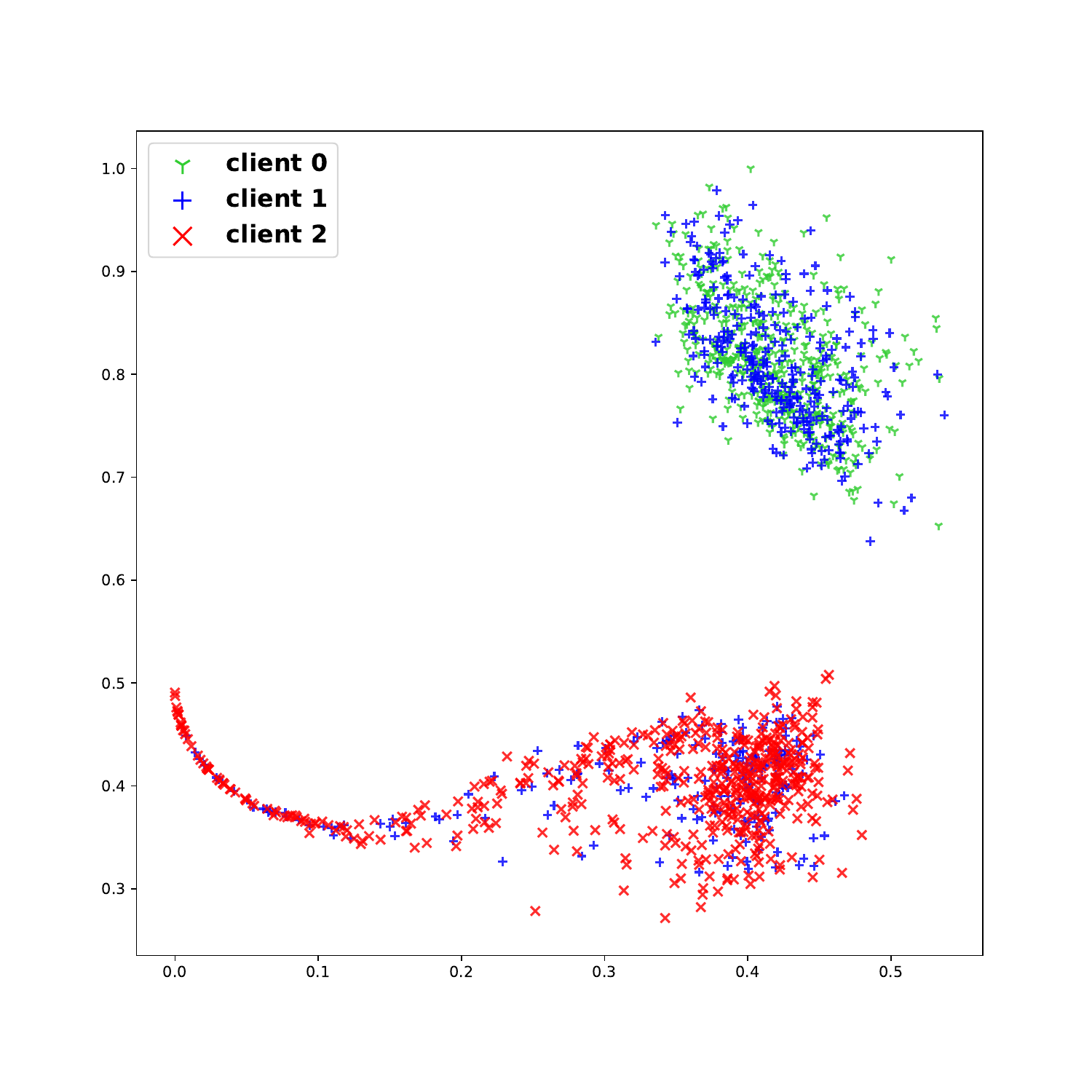}
\label{fig:exp_tsne_fedavg}}
\hfil
\subfloat[HeteroAvg.]
{\includegraphics[width=0.125\columnwidth]{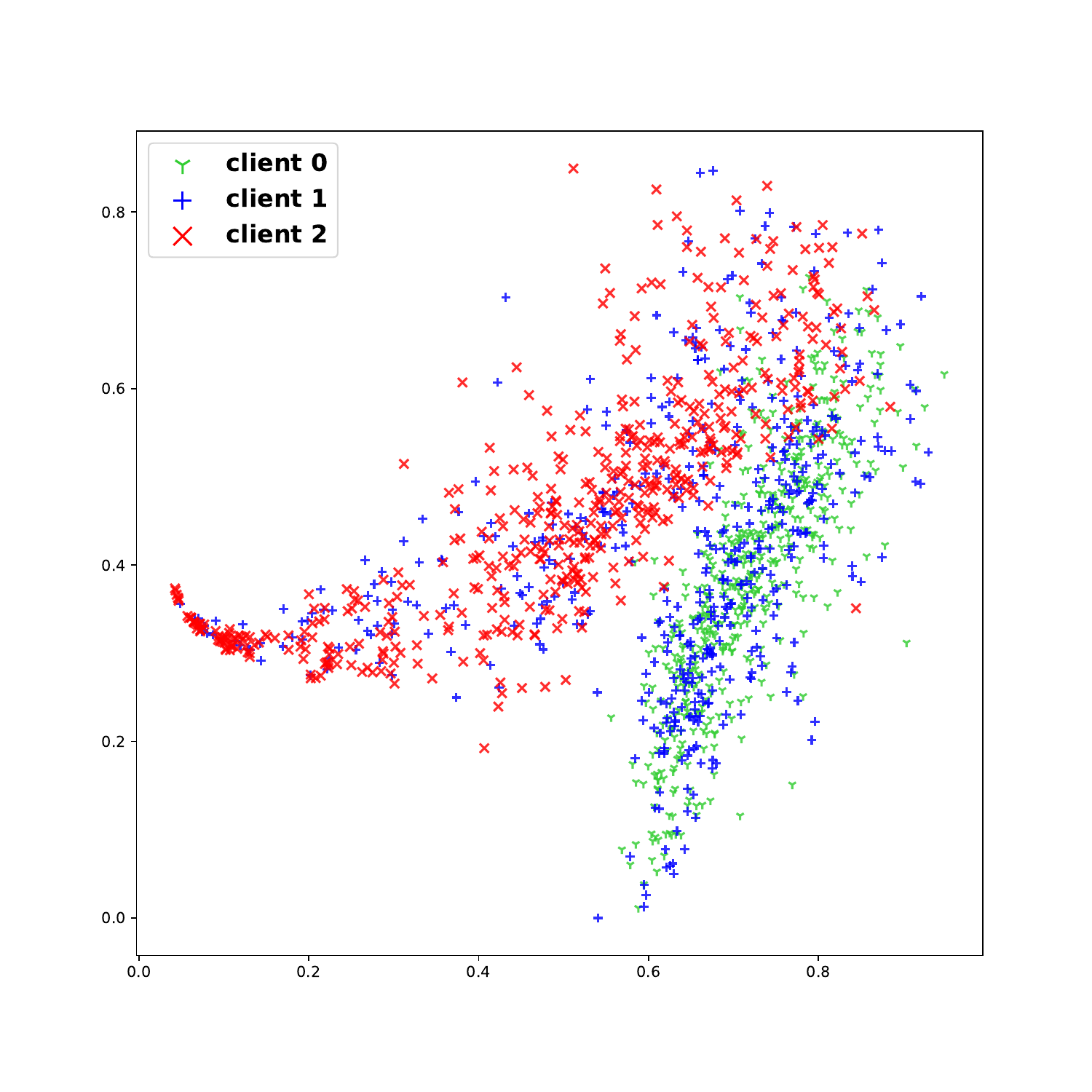}
\label{fig:exp_tsne_heteroavg}}
\hfil
\subfloat[InCoAvg.]
{\includegraphics[width=0.125\columnwidth]{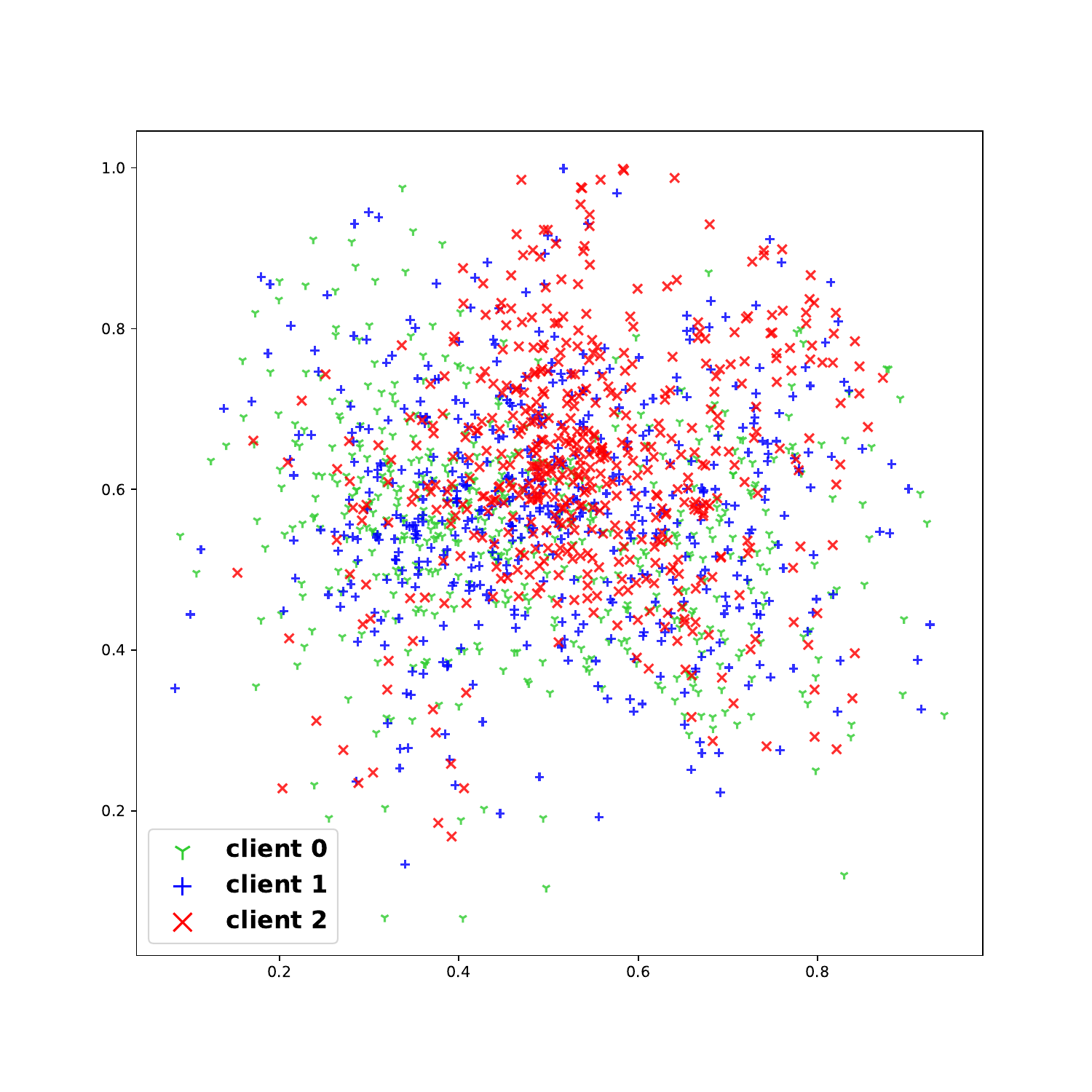}
\label{fig:exp_tsne_FedInCo}}
\vskip -0.07in
\caption{Important coefficients of Theorem~\ref{thm:vec_solution} and t-SNE visualization of features. (a): $\theta$ in all layers. (b): $\beta$ in Layer 11. (c): $\beta$ in Layer 13. (d) to (f): t-SNE visualization of features learned by different methods on CIFAR-10. We select data from one class and three clients (client 0: ResNet10, client 1: ResNet14, client 2: ResNet26) to simplify the notations in t-SNE figures.}
\label{fig:exp_coefficient_thm_analyses}
% \vskip -0.15in
\end{figure}

\begin{wraptable}{r}{0.35\columnwidth}
\vskip -0.16in
\tiny
    \caption{The Percentage of $\beta>0$}
    \label{tab:beta_statistics}
    \centering
    \def\arraystretch{1.3}
    \begin{tabular}{ccc}
    \toprule
    % \hline
        \multirow{2}*{Methods} & \multicolumn{2}{c}{Percentage of $\beta>0$} \\
        \cline{2-3}
         & Layer 11 & Layer 13 \\
      % \hline
      \midrule
       InCoAvg &  \textbf{83.8} & \textbf{74.4}  \\
       
       InCoAvg w/o O & 53.5 & 50.2  \\
    % \hline
    \bottomrule
    \end{tabular}
\vskip -0.15in
\end{wraptable}

\textit{Analysis for $\theta$ and $\beta$.} 
In our experiments, we set $\theta = 1$ for InCoAvg w/o Optimization, the blue dash line in Figure~\ref{fig:exp_theta_val}. However, under Theorem~\ref{thm:vec_solution}, we observe that the value of $\theta$ varies for different layers, indicating the effectiveness of the theorem in automatically determining the appropriate $\theta$ values.
$\beta > 0$ denotes the same direction between shallow layer gradients and the current layer gradients. 
Furthermore, Table~\ref{tab:beta_statistics} provides empirical evidence supporting the efficacy of Theorem~\ref{thm:vec_solution} in heterogeneous FL.

\textit{t-SNE Visualizations.}
Figure~\ref{fig:exp_tsne_fedavg} and Figure~\ref{fig:exp_tsne_heteroavg} provide visual evidence of bias stemming from model heterogeneity in the FedAvg and HeteroAvg.
In contrast, Figure~\ref{fig:exp_tsne_FedInCo} demonstrates that InCoAvg effectively addresses bias. These findings highlight the superior generalization capability of InCoAvg compared to HeteroAvg and FedAvg, indicating that InCoAvg mitigates bias issues in client models.

\begin{figure*}[!t]
\vskip -0.3in
\centering
    \subfloat[CIFAR10 with $\alpha=0.5$.]
    {\includegraphics[width=0.25\textwidth]{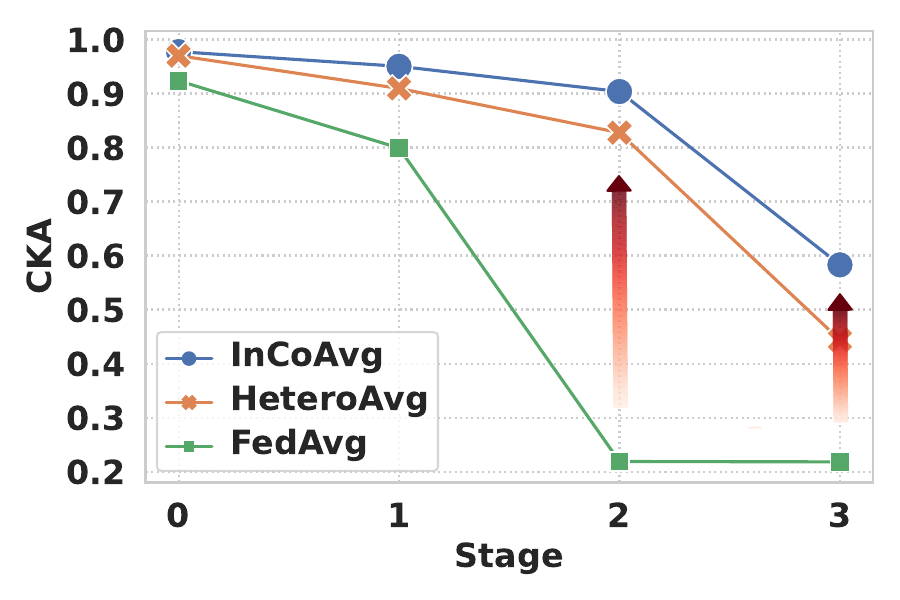}
    \label{fig:exp_resnet_05cifar}}
    \subfloat[InCoAvg.] 
    { 
    \includegraphics[width=0.16\textwidth]{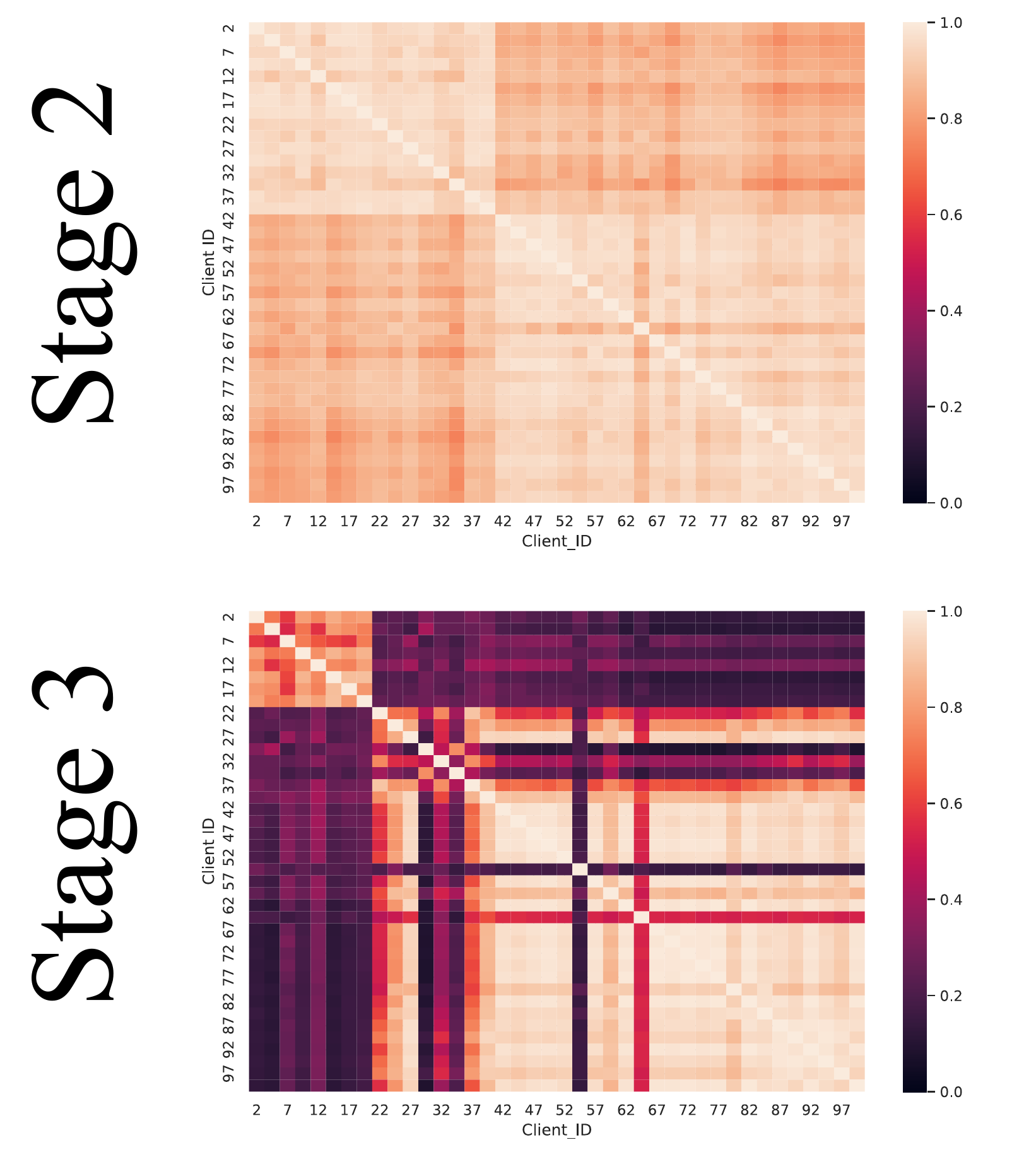}
    \label{fig:exp_resnet_FedInCo_block3_05cifar}}
    % \hspace{5pt}
    \subfloat[HeteroAvg.]
    {\includegraphics[width=0.133\textwidth]{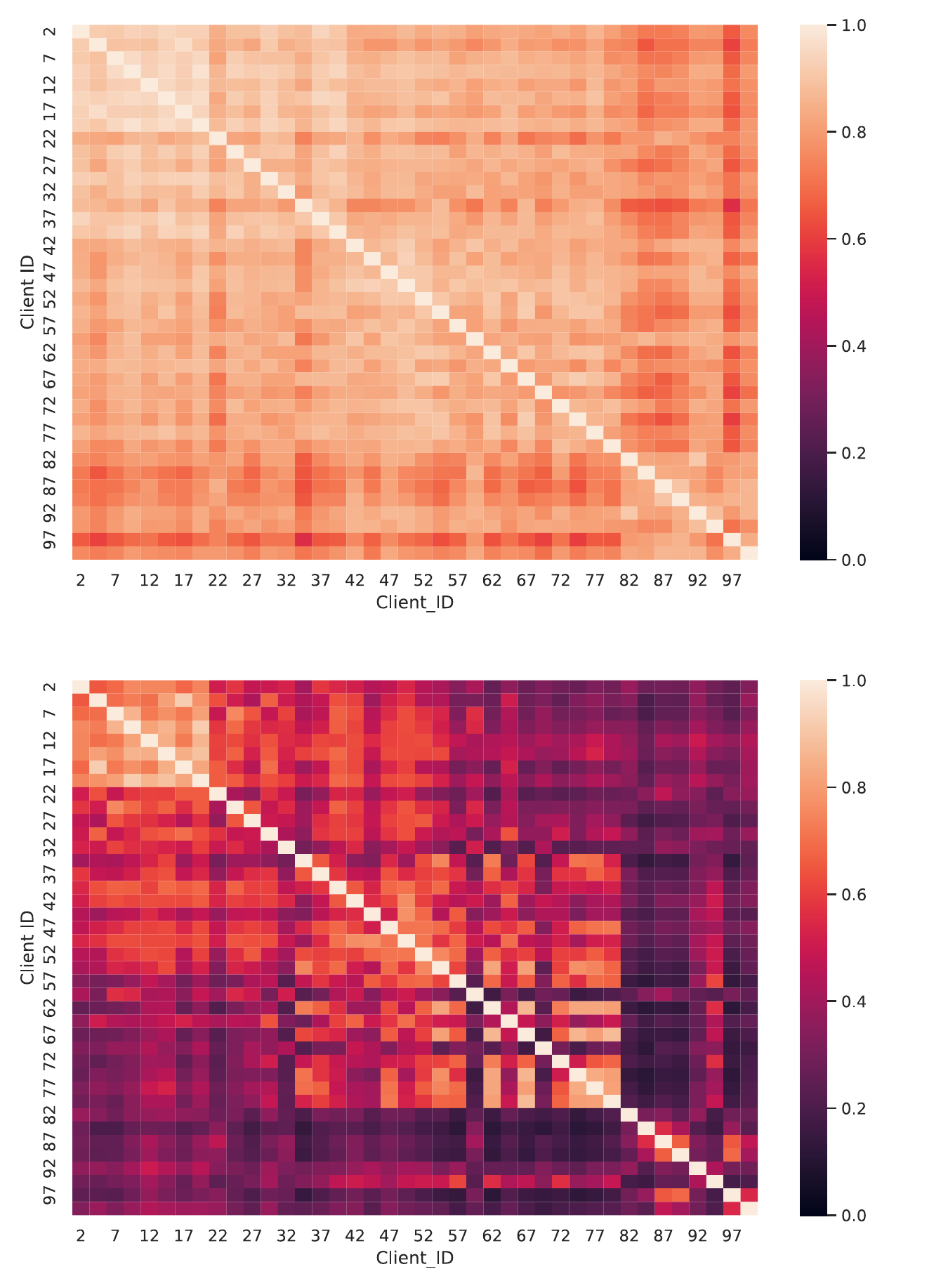}
    \label{fig:exp_resnet_heteroAvg_block3_05cifar}}
    % \hspace{5pt}
    \subfloat[FedAvg.]
    {\includegraphics[width=0.133\textwidth]{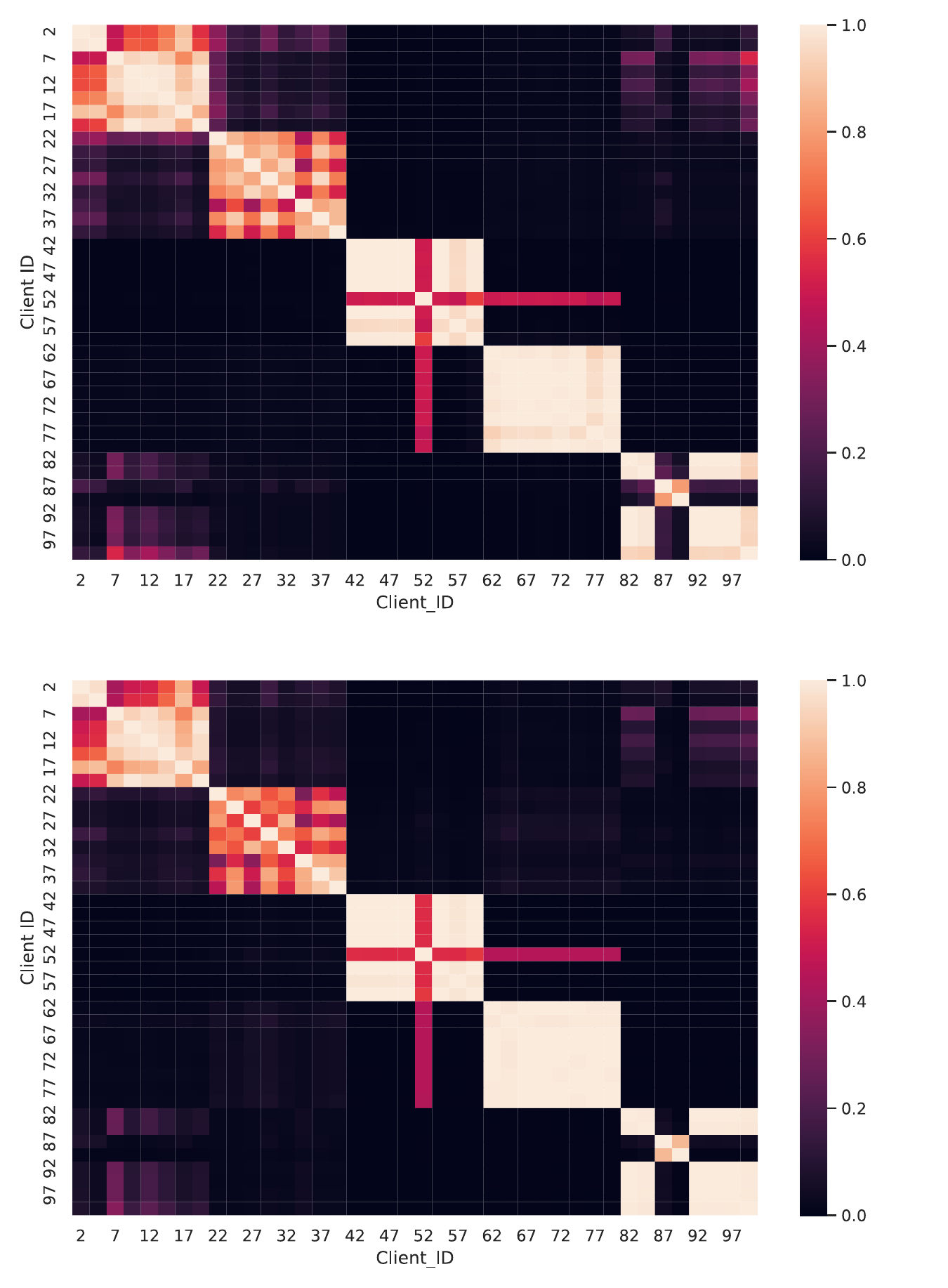}
    \label{fig:exp_resnet_fedAvg_block3_05cifar}
    }
    \subfloat[Client Accuracy.]{
     \raisebox{\height}{
        \def\arraystretch{1.3}
        \tiny
        \begin{tabular}{ccc}
        \toprule
           CID   & InCoAvg & HeteroAvg \\
          \midrule
           0 & 54.4 & 54.8  \\
           20 & \textbf{70.7} & 67.8  \\
           40 & \textbf{75.3} & 72.6  \\
           60 & \textbf{73.3} & 69.9  \\
           80 & \textbf{72.0} & 68.5  \\
        \bottomrule
        \end{tabular}
        \label{tab:client_acc}
        }}
\vskip -0.07in
\caption{CKA layer similarity, Heatmaps, and accuracy of different clients. (a): The layer similarity of different methods. (b) to (d): Heatmaps for different methods in stage 2 and stage 3. (e): Accuracy of each client group. (0: ResNet10, 20: ResNet14, 40: ResNet18, 60: ResNet22, 80: ResNet26)}
\label{fig:exp_cka_resnet}
\vskip -0.1in
\end{figure*}

\textit{Analysis for CKA Layer Similarity.}
Figure~\ref{fig:exp_resnet_05cifar} reveals that InCoAvg exhibits a significantly higher CKA layer similarity compared to FedAvg.
Consistent with the t-SNE visualization, FedAvg's heatmaps exhibit block-wise patterns in Figure~\ref{fig:exp_resnet_fedAvg_block3_05cifar} due to its inability to extract features from diverse model architectures. 
Notably, the smallest models in InCoAvg (top left corner) exhibit lower similarity (more black) with other clients compared to HeteroAvg in stage 3. 
This discrepancy arises because the accuracy of the smallest models in InCoAvg is similar to that of HeteroAvg, but the performance of larger models in InCoAvg surpasses that of HeteroAvg, as indicated in Figure~\ref{tab:client_acc}. Consequently, a larger similarity gap emerges between the smallest models and the other models. Addressing the performance of the smallest models in InCo Aggregation represents our future research direction.

\section{Conclusions}
We propose a novel FL training scheme called InCo Aggregation, which aims to enhance the capabilities of model-homogeneous FL methods in heterogeneous FL settings. 
Our approach leverages normalized cross-layer gradients to promote similarity among deep layers across different clients. Additionally, we introduce a convex optimization formulation to address the challenge of gradient divergence. 
Through extensive experimental evaluations, we demonstrate the effectiveness of InCo Aggregation in improving heterogeneous FL performance.

{
% \small
\bibliography{ref}

\begin{thebibliography}{62}
\providecommand{\natexlab}[1]{#1}
\providecommand{\url}[1]{\texttt{#1}}
\expandafter\ifx\csname urlstyle\endcsname\relax
  \providecommand{\doi}[1]{doi: #1}\else
  \providecommand{\doi}{doi: \begingroup \urlstyle{rm}\Url}\fi

\bibitem[Alam et~al.(2022)Alam, Liu, Yan, and Zhang]{alam2022fedrolex}
Samiul Alam, Luyang Liu, Ming Yan, and Mi~Zhang.
\newblock Fed{R}olex: Model-heterogeneous federated learning with rolling sub-model extraction.
\newblock \emph{Advances in Neural Information Processing Systems}, 35:\penalty0 29677--29690, 2022.

\bibitem[Alvarez(2023)]{9926163}
Sergio~A. Alvarez.
\newblock Gaussian rbf centered kernel alignment (cka) in the large-bandwidth limit.
\newblock \emph{IEEE Transactions on Pattern Analysis and Machine Intelligence}, 45\penalty0 (5):\penalty0 6587--6593, 2023.
\newblock \doi{10.1109/TPAMI.2022.3216518}.

\bibitem[Baek et~al.(2022)Baek, Yun, Kwak, Jung, Ji, Bennis, Park, and Kim]{baek2022joint}
Hankyul Baek, Won~Joon Yun, Yunseok Kwak, Soyi Jung, Mingyue Ji, Mehdi Bennis, Jihong Park, and Joongheon Kim.
\newblock Joint superposition coding and training for federated learning over multi-width neural networks.
\newblock In \emph{IEEE INFOCOM 2022-IEEE Conference on Computer Communications}, pp.\  1729--1738. IEEE, 2022.

\bibitem[Bot et~al.(2009)Bot, Grad, and Wanka]{bot2009duality}
Radu~Ioan Bot, Sorin-Mihai Grad, and Gert Wanka.
\newblock \emph{Duality in vector optimization}.
\newblock Springer Science \& Business Media, 2009.

\bibitem[Boyd et~al.(2004)Boyd, Boyd, and Vandenberghe]{boyd2004convex}
Stephen Boyd, Stephen~P Boyd, and Lieven Vandenberghe.
\newblock \emph{Convex optimization}.
\newblock Cambridge university press, 2004.

\bibitem[Caldas et~al.(2018)Caldas, Kone{\v{c}}ny, McMahan, and Talwalkar]{caldas2018expanding}
Sebastian Caldas, Jakub Kone{\v{c}}ny, H~Brendan McMahan, and Ameet Talwalkar.
\newblock Expanding the reach of federated learning by reducing client resource requirements.
\newblock \emph{arXiv preprint arXiv:1812.07210}, 2018.

\bibitem[Chan \& Ngai(2021)Chan and Ngai]{fedhe2021}
Yun~Hin Chan and Edith Ngai.
\newblock Fedhe: Heterogeneous models and communication-efficient federated learning.
\newblock \emph{IEEE International Confer- ence on Mobility, Sensing and Networking (MSN 2021)}, 2021.

\bibitem[Chan \& Ngai(2022)Chan and Ngai]{chan2022exploiting}
Yun-Hin Chan and Edith C-H Ngai.
\newblock Exploiting features and logits in heterogeneous federated learning.
\newblock \emph{arXiv preprint arXiv:2210.15527}, 2022.

\bibitem[Chen et~al.(2021)Chen, Xu, Wang, Li, Li, Chen, and Zhang]{chen2021communication}
Chen Chen, Hong Xu, Wei Wang, Baochun Li, Bo~Li, Li~Chen, and Gong Zhang.
\newblock Communication-efficient federated learning with adaptive parameter freezing.
\newblock In \emph{2021 IEEE 41st International Conference on Distributed Computing Systems (ICDCS)}, pp.\  1--11. IEEE, 2021.

\bibitem[Chen et~al.(2019)Chen, Wang, Xu, Yang, Liu, Shi, Xu, Xu, and Tian]{chen2019data}
Hanting Chen, Yunhe Wang, Chang Xu, Zhaohui Yang, Chuanjian Liu, Boxin Shi, Chunjing Xu, Chao Xu, and Qi~Tian.
\newblock Data-free learning of student networks.
\newblock In \emph{Proceedings of the IEEE/CVF International Conference on Computer Vision}, pp.\  3514--3522, 2019.

\bibitem[Cortes et~al.(2012)Cortes, Mohri, and Rostamizadeh]{cortes2012algorithms}
Corinna Cortes, Mehryar Mohri, and Afshin Rostamizadeh.
\newblock Algorithms for learning kernels based on centered alignment.
\newblock \emph{The Journal of Machine Learning Research}, 13\penalty0 (1):\penalty0 795--828, 2012.

\bibitem[Darlow et~al.(2018)Darlow, Crowley, Antoniou, and Storkey]{darlow2018cinic}
Luke~N Darlow, Elliot~J Crowley, Antreas Antoniou, and Amos~J Storkey.
\newblock Cinic-10 is not imagenet or cifar-10.
\newblock \emph{arXiv preprint arXiv:1810.03505}, 2018.

\bibitem[Diao et~al.(2021)Diao, Ding, and Tarokh]{diao2021heterofl}
Enmao Diao, Jie Ding, and Vahid Tarokh.
\newblock Hetero{FL}: Computation and communication efficient federated learning for heterogeneous clients.
\newblock In \emph{International Conference on Learning Representations}, 2021.

\bibitem[Dosovitskiy et~al.(2020)Dosovitskiy, Beyer, Kolesnikov, Weissenborn, Zhai, Unterthiner, Dehghani, Minderer, Heigold, Gelly, et~al.]{dosovitskiy2020image}
Alexey Dosovitskiy, Lucas Beyer, Alexander Kolesnikov, Dirk Weissenborn, Xiaohua Zhai, Thomas Unterthiner, Mostafa Dehghani, Matthias Minderer, Georg Heigold, Sylvain Gelly, et~al.
\newblock An image is worth 16x16 words: Transformers for image recognition at scale.
\newblock \emph{arXiv preprint arXiv:2010.11929}, 2020.

\bibitem[Fang \& Ye(2022)Fang and Ye]{fang2022robust}
Xiuwen Fang and Mang Ye.
\newblock Robust federated learning with noisy and heterogeneous clients.
\newblock In \emph{Proceedings of the IEEE/CVF Conference on Computer Vision and Pattern Recognition}, pp.\  10072--10081, 2022.

\bibitem[Gao et~al.(2022)Gao, Yao, and Yang]{Survey_on_Heterogeneous}
Dashan Gao, Xin Yao, and Qiang Yang.
\newblock A survey on heterogeneous federated learning, 2022.
\newblock URL \url{https://arxiv.org/abs/2210.04505}.

\bibitem[Goodfellow et~al.(2014)Goodfellow, Pouget-Abadie, Mirza, Xu, Warde-Farley, Ozair, Courville, and Bengio]{goodfellow2014generative}
Ian Goodfellow, Jean Pouget-Abadie, Mehdi Mirza, Bing Xu, David Warde-Farley, Sherjil Ozair, Aaron Courville, and Yoshua Bengio.
\newblock Generative adversarial nets.
\newblock \emph{Advances in neural information processing systems}, 27, 2014.

\bibitem[Gou et~al.(2021)Gou, Yu, Maybank, and Tao]{Gou_2021}
Jianping Gou, Baosheng Yu, Stephen~J. Maybank, and Dacheng Tao.
\newblock Knowledge distillation: A survey.
\newblock \emph{International Journal of Computer Vision}, 129\penalty0 (6):\penalty0 1789–1819, March 2021.
\newblock ISSN 1573-1405.
\newblock \doi{10.1007/s11263-021-01453-z}.
\newblock URL \url{http://dx.doi.org/10.1007/s11263-021-01453-z}.

\bibitem[He et~al.(2020)He, Annavaram, and Avestimehr]{he2020group}
Chaoyang He, Murali Annavaram, and Salman Avestimehr.
\newblock Group knowledge transfer: Federated learning of large cnns at the edge.
\newblock \emph{Advances in Neural Information Processing Systems}, 33:\penalty0 14068--14080, 2020.

\bibitem[He et~al.(2016)He, Zhang, Ren, and Sun]{he2016identity}
Kaiming He, Xiangyu Zhang, Shaoqing Ren, and Jian Sun.
\newblock Identity mappings in deep residual networks.
\newblock In \emph{Computer Vision--ECCV 2016: 14th European Conference, Amsterdam, The Netherlands, October 11--14, 2016, Proceedings, Part IV 14}, pp.\  630--645. Springer, 2016.

\bibitem[Hinton et~al.(2015)Hinton, Vinyals, and Dean]{hinton2015distilling}
Geoffrey Hinton, Oriol Vinyals, and Jeff Dean.
\newblock Distilling the knowledge in a neural network.
\newblock \emph{NIPS Deep Learning and Representation Learning Workshop}, 2015.

\bibitem[Hochreiter \& Schmidhuber(1997)Hochreiter and Schmidhuber]{hochreiter1997long}
Sepp Hochreiter and J{\"u}rgen Schmidhuber.
\newblock Long short-term memory.
\newblock \emph{Neural computation}, 9\penalty0 (8):\penalty0 1735--1780, 1997.

\bibitem[Horvath et~al.(2021)Horvath, Laskaridis, Almeida, Leontiadis, Venieris, and Lane]{horvath2021fjord}
Samuel Horvath, Stefanos Laskaridis, Mario Almeida, Ilias Leontiadis, Stylianos Venieris, and Nicholas Lane.
\newblock Fjord: Fair and accurate federated learning under heterogeneous targets with ordered dropout.
\newblock \emph{Advances in Neural Information Processing Systems}, 34:\penalty0 12876--12889, 2021.

\bibitem[Huang et~al.(2022)Huang, Ye, and Du]{huang2022learn}
Wenke Huang, Mang Ye, and Bo~Du.
\newblock Learn from others and be yourself in heterogeneous federated learning.
\newblock In \emph{Proceedings of the IEEE/CVF Conference on Computer Vision and Pattern Recognition}, pp.\  10143--10153, 2022.

\bibitem[Huang et~al.(2021)Huang, Gupta, Song, Li, and Arora]{huang2021evaluating}
Yangsibo Huang, Samyak Gupta, Zhao Song, Kai Li, and Sanjeev Arora.
\newblock Evaluating gradient inversion attacks and defenses in federated learning.
\newblock \emph{Advances in Neural Information Processing Systems}, 34:\penalty0 7232--7241, 2021.

\bibitem[Ilhan et~al.(2023)Ilhan, Su, and Liu]{ilhan2023scalefl}
Fatih Ilhan, Gong Su, and Ling Liu.
\newblock Scalefl: Resource-adaptive federated learning with heterogeneous clients.
\newblock In \emph{Proceedings of the IEEE/CVF Conference on Computer Vision and Pattern Recognition}, pp.\  24532--24541, 2023.

\bibitem[Jeong \& Hwang(2022)Jeong and Hwang]{jeong2022factorized}
Wonyong Jeong and Sung~Ju Hwang.
\newblock Factorized-fl: Personalized federated learning with parameter factorization \& similarity matching.
\newblock In \emph{Advances in Neural Information Processing Systems}, 2022.

\bibitem[Kairouz et~al.(2021)Kairouz, McMahan, Avent, Bellet, Bennis, Bhagoji, Bonawitz, Charles, Cormode, Cummings, et~al.]{kairouz2019advances}
Peter Kairouz, H~Brendan McMahan, Brendan Avent, Aur{\'e}lien Bellet, Mehdi Bennis, Arjun~Nitin Bhagoji, Kallista Bonawitz, Zachary Charles, Graham Cormode, Rachel Cummings, et~al.
\newblock Advances and open problems in federated learning.
\newblock 2021.

\bibitem[Karimireddy et~al.(2020)Karimireddy, Kale, Mohri, Reddi, Stich, and Suresh]{karimireddy2020scaffold}
Sai~Praneeth Karimireddy, Satyen Kale, Mehryar Mohri, Sashank Reddi, Sebastian Stich, and Ananda~Theertha Suresh.
\newblock Scaffold: Stochastic controlled averaging for federated learning.
\newblock In \emph{International Conference on Machine Learning}, pp.\  5132--5143. PMLR, 2020.

\bibitem[Kingma \& Ba(2015)Kingma and Ba]{KingmaB14}
Diederik~P. Kingma and Jimmy Ba.
\newblock Adam: {A} method for stochastic optimization.
\newblock In Yoshua Bengio and Yann LeCun (eds.), \emph{3rd International Conference on Learning Representations, {ICLR} 2015, San Diego, CA, USA, May 7-9, 2015, Conference Track Proceedings}, 2015.

\bibitem[Kornblith et~al.(2019)Kornblith, Norouzi, Lee, and Hinton]{kornblith2019similarity}
Simon Kornblith, Mohammad Norouzi, Honglak Lee, and Geoffrey Hinton.
\newblock Similarity of neural network representations revisited.
\newblock In \emph{International Conference on Machine Learning}, pp.\  3519--3529. PMLR, 2019.

\bibitem[Krizhevsky et~al.(2009)Krizhevsky, Hinton, et~al.]{krizhevsky2009learning}
Alex Krizhevsky, Geoffrey Hinton, et~al.
\newblock Learning multiple layers of features from tiny images.
\newblock 2009.

\bibitem[Li \& Wang(2019)Li and Wang]{li2019fedmd}
Daliang Li and Junpu Wang.
\newblock Fed{MD}: Heterogenous federated learning via model distillation.
\newblock \emph{NeurIPS Workshop on Federated Learning for Data Privacy and Confidentiality}, 2019.

\bibitem[Li et~al.(2021{\natexlab{a}})Li, He, and Song]{li2021model}
Qinbin Li, Bingsheng He, and Dawn Song.
\newblock Model-contrastive federated learning.
\newblock In \emph{Proceedings of the IEEE/CVF Conference on Computer Vision and Pattern Recognition}, pp.\  10713--10722, 2021{\natexlab{a}}.

\bibitem[Li et~al.(2022)Li, Diao, Chen, and He]{li2022federated}
Qinbin Li, Yiqun Diao, Quan Chen, and Bingsheng He.
\newblock Federated learning on non-iid data silos: An experimental study.
\newblock In \emph{2022 IEEE 38th International Conference on Data Engineering (ICDE)}, pp.\  965--978. IEEE, 2022.

\bibitem[Li et~al.(2020{\natexlab{a}})Li, Sahu, Talwalkar, and Smith]{li2020federated}
Tian Li, Anit~Kumar Sahu, Ameet Talwalkar, and Virginia Smith.
\newblock Federated learning: Challenges, methods, and future directions.
\newblock \emph{IEEE signal processing magazine}, 37\penalty0 (3):\penalty0 50--60, 2020{\natexlab{a}}.

\bibitem[Li et~al.(2020{\natexlab{b}})Li, Sahu, Zaheer, Sanjabi, Talwalkar, and Smith]{li2018federated}
Tian Li, Anit~Kumar Sahu, Manzil Zaheer, Maziar Sanjabi, Ameet Talwalkar, and Virginia Smith.
\newblock Federated optimization in heterogeneous networks.
\newblock \emph{Proceedings of the 3rd MLSys Conference}, 2020{\natexlab{b}}.

\bibitem[Li et~al.(2015)Li, Yosinski, Clune, Lipson, and Hopcroft]{li2015convergent}
Yixuan Li, Jason Yosinski, Jeff Clune, Hod Lipson, and John Hopcroft.
\newblock Convergent learning: Do different neural networks learn the same representations?
\newblock \emph{arXiv preprint arXiv:1511.07543}, 2015.

\bibitem[Li et~al.(2021{\natexlab{b}})Li, Zhou, Wang, Mi, and Hospedales]{li2021fedh2l}
Yiying Li, Wei Zhou, Huaimin Wang, Haibo Mi, and Timothy~M Hospedales.
\newblock Fed{H2L}: Federated learning with model and statistical heterogeneity.
\newblock \emph{arXiv preprint arXiv:2101.11296}, 2021{\natexlab{b}}.

\bibitem[Lin et~al.(2020)Lin, Kong, Stich, and Jaggi]{lin2020ensemble}
Tao Lin, Lingjing Kong, Sebastian~U Stich, and Martin Jaggi.
\newblock Ensemble distillation for robust model fusion in federated learning.
\newblock \emph{Advances in Neural Information Processing Systems}, 33:\penalty0 2351--2363, 2020.

\bibitem[Liu et~al.(2022)Liu, Wu, Wu, Wang, Lyu, Chen, and Xie]{liu2022no}
Ruixuan Liu, Fangzhao Wu, Chuhan Wu, Yanlin Wang, Lingjuan Lyu, Hong Chen, and Xing Xie.
\newblock No one left behind: Inclusive federated learning over heterogeneous devices.
\newblock In \emph{Proceedings of the 28th ACM SIGKDD Conference on Knowledge Discovery and Data Mining}, pp.\  3398--3406, 2022.

\bibitem[Luo et~al.(2021)Luo, Chen, Hu, Zhang, Liang, and Feng]{luo2021no}
Mi~Luo, Fei Chen, Dapeng Hu, Yifan Zhang, Jian Liang, and Jiashi Feng.
\newblock No fear of heterogeneity: Classifier calibration for federated learning with non-iid data.
\newblock \emph{Advances in Neural Information Processing Systems}, 34:\penalty0 5972--5984, 2021.

\bibitem[McMahan et~al.(2017)McMahan, Moore, Ramage, Hampson, and y~Arcas]{mcmahan2017communication}
Brendan McMahan, Eider Moore, Daniel Ramage, Seth Hampson, and Blaise~Aguera y~Arcas.
\newblock Communication-efficient learning of deep networks from decentralized data.
\newblock In \emph{Artificial intelligence and statistics}, pp.\  1273--1282. PMLR, 2017.

\bibitem[Mohri et~al.(2019)Mohri, Sivek, and Suresh]{mohri2019agnostic}
Mehryar Mohri, Gary Sivek, and Ananda~Theertha Suresh.
\newblock Agnostic federated learning.
\newblock In \emph{International Conference on Machine Learning}, pp.\  4615--4625. PMLR, 2019.

\bibitem[Morcos et~al.(2018)Morcos, Raghu, and Bengio]{morcos2018insights}
Ari Morcos, Maithra Raghu, and Samy Bengio.
\newblock Insights on representational similarity in neural networks with canonical correlation.
\newblock \emph{Advances in neural information processing systems}, 31, 2018.

\bibitem[Nayak et~al.(2019)Nayak, Mopuri, Shaj, Radhakrishnan, and Chakraborty]{nayak2019zero}
Gaurav~Kumar Nayak, Konda~Reddy Mopuri, Vaisakh Shaj, Venkatesh~Babu Radhakrishnan, and Anirban Chakraborty.
\newblock Zero-shot knowledge distillation in deep networks.
\newblock In \emph{International Conference on Machine Learning}, pp.\  4743--4751. PMLR, 2019.

\bibitem[Netzer et~al.(2011)Netzer, Wang, Coates, Bissacco, Wu, and Ng]{netzer2011reading}
Yuval Netzer, Tao Wang, Adam Coates, Alessandro Bissacco, Bo~Wu, and Andrew~Y Ng.
\newblock Reading digits in natural images with unsupervised feature learning.
\newblock 2011.

\bibitem[Raghu et~al.(2017)Raghu, Gilmer, Yosinski, and Sohl-Dickstein]{raghu2017svcca}
Maithra Raghu, Justin Gilmer, Jason Yosinski, and Jascha Sohl-Dickstein.
\newblock Svcca: Singular vector canonical correlation analysis for deep learning dynamics and interpretability.
\newblock \emph{Advances in neural information processing systems}, 30, 2017.

\bibitem[Raghu et~al.(2021)Raghu, Unterthiner, Kornblith, Zhang, and Dosovitskiy]{raghu2021vision}
Maithra Raghu, Thomas Unterthiner, Simon Kornblith, Chiyuan Zhang, and Alexey Dosovitskiy.
\newblock Do vision transformers see like convolutional neural networks?
\newblock \emph{Advances in Neural Information Processing Systems}, 34:\penalty0 12116--12128, 2021.

\bibitem[Ruder(2016)]{ruder2016overview}
Sebastian Ruder.
\newblock An overview of gradient descent optimization algorithms.
\newblock \emph{arXiv preprint arXiv:1609.04747}, 2016.

\bibitem[Sattler et~al.(2021)Sattler, Korjakow, Rischke, and Samek]{sattler2021fedaux}
Felix Sattler, Tim Korjakow, Roman Rischke, and Wojciech Samek.
\newblock Fedaux: Leveraging unlabeled auxiliary data in federated learning.
\newblock \emph{IEEE Transactions on Neural Networks and Learning Systems}, 2021.

\bibitem[Shen et~al.(2020)Shen, Zhang, Jia, Zhang, Huang, Zhou, Kuang, Wu, and Wu]{shen2020federated}
Tao Shen, Jie Zhang, Xinkang Jia, Fengda Zhang, Gang Huang, Pan Zhou, Kun Kuang, Fei Wu, and Chao Wu.
\newblock Federated mutual learning.
\newblock \emph{arXiv preprint arXiv:2006.16765}, 2020.

\bibitem[Tan et~al.(2022)Tan, Long, Liu, Zhou, Lu, Jiang, and Zhang]{tan2022fedproto}
Yue Tan, Guodong Long, Lu~Liu, Tianyi Zhou, Qinghua Lu, Jing Jiang, and Chengqi Zhang.
\newblock Fedproto: Federated prototype learning across heterogeneous clients.
\newblock In \emph{Proceedings of the AAAI Conference on Artificial Intelligence}, volume~36, pp.\  8432--8440, 2022.

\bibitem[Wang et~al.(2020)Wang, Liu, Liang, Joshi, and Poor]{wang2020tackling}
Jianyu Wang, Qinghua Liu, Hao Liang, Gauri Joshi, and H~Vincent Poor.
\newblock Tackling the objective inconsistency problem in heterogeneous federated optimization.
\newblock \emph{Advances in neural information processing systems}, 33:\penalty0 7611--7623, 2020.

\bibitem[Wang et~al.(2018)Wang, Hu, Gu, Hu, Wu, He, and Hopcroft]{wang2018towards}
Liwei Wang, Lunjia Hu, Jiayuan Gu, Zhiqiang Hu, Yue Wu, Kun He, and John Hopcroft.
\newblock Towards understanding learning representations: To what extent do different neural networks learn the same representation.
\newblock \emph{Advances in neural information processing systems}, 31, 2018.

\bibitem[Xiao et~al.(2017)Xiao, Rasul, and Vollgraf]{xiao2017/online}
Han Xiao, Kashif Rasul, and Roland Vollgraf.
\newblock Fashion-{MNIST}: a novel image dataset for benchmarking machine learning algorithms, 2017.

\bibitem[Xie et~al.(2020)Xie, Koyejo, and Gupta]{xie2019asynchronous}
Cong Xie, Sanmi Koyejo, and Indranil Gupta.
\newblock Asynchronous federated optimization.
\newblock \emph{12th Annual Workshop on Optimization for Machine Learning}, 2020.

\bibitem[Xiong et~al.(2020)Xiong, Yang, He, Zheng, Zheng, Xing, Zhang, Lan, Wang, and Liu]{xiong2020layer}
Ruibin Xiong, Yunchang Yang, Di~He, Kai Zheng, Shuxin Zheng, Chen Xing, Huishuai Zhang, Yanyan Lan, Liwei Wang, and Tieyan Liu.
\newblock On layer normalization in the transformer architecture.
\newblock In \emph{International Conference on Machine Learning}, pp.\  10524--10533. PMLR, 2020.

\bibitem[Yu \& Huang(2019)Yu and Huang]{yu2019universally}
Jiahui Yu and Thomas~S Huang.
\newblock Universally slimmable networks and improved training techniques.
\newblock In \emph{Proceedings of the IEEE/CVF international conference on computer vision}, pp.\  1803--1811, 2019.

\bibitem[Zhao et~al.(2023)Zhao, Yu, Zhao, and Ngai]{10.1145/3610873}
Running Zhao, Jiangtao Yu, Hang Zhao, and Edith~C.H. Ngai.
\newblock Radio2text: Streaming speech recognition using mmwave radio signals.
\newblock 7\penalty0 (3), sep 2023.
\newblock \doi{10.1145/3610873}.
\newblock URL \url{https://doi.org/10.1145/3610873}.

\bibitem[Zhao et~al.(2018)Zhao, Li, Lai, Suda, Civin, and Chandra]{zhao2018federated}
Yue Zhao, Meng Li, Liangzhen Lai, Naveen Suda, Damon Civin, and Vikas Chandra.
\newblock Federated learning with non-iid data.
\newblock \emph{arXiv preprint arXiv:1806.00582}, 2018.

\bibitem[Zhu et~al.(2021)Zhu, Hong, and Zhou]{zhu2021data}
Zhuangdi Zhu, Junyuan Hong, and Jiayu Zhou.
\newblock Data-free knowledge distillation for heterogeneous federated learning.
\newblock In \emph{International Conference on Machine Learning}, pp.\  12878--12889. PMLR, 2021.

\end{thebibliography}
\bibliographystyle{iclr2024_conference}

}

\newpage
\appendix
\onecolumn

\section{Centered Kernel Alignment}
\label{sec:CKA}
Centered Kernel Alignment (CKA) originally serves as a similarity measure for different kernel functions \citep{cortes2012algorithms}. Later, its purpose has been extended to discovering meaningful similarities between internal representations of neural networks \citep{kornblith2019similarity}. 
Compared with alternative methods to monitor representation learning, such as Canonical Correlation Analysis-based methods \citep{raghu2017svcca, morcos2018insights} and neuron alignment methods \citep{li2015convergent, wang2018towards}, CKA achieves the state-of-the-art performance in measuring the difference between representations of neural network. This is based on the fact that CKA reliably identifies correspondences between representations from architecturally corresponding layers in two networks trained with different initializations.\footnote{We refer the readers to Section 6.1. in \cite{kornblith2019similarity} for a complete sanity check of representational similarity measures.}

Denote $X\in \mathbb{R}^{n\times p}$ and $Y\in \mathbb{R}^{n\times q}$ as two representations of $n$ data points with possibly different dimensions (i.e., $p\neq q$). These two representations fall into the following three categories: (1) internal outputs at two different layers of an individual network, (2) internal layer outputs of two architecturally identical networks trained from different initialization or by different datasets, or (3) internal layer outputs of two networks with different architectures possibly trained by different datasets. The application of CKA in our paper belongs to the third category, where we examine the CKA similarities of a corresponding layer output between every pair of local client models in the context of federated learning.

Let $k_x(\cdot,\cdot)$ and $k_y(\cdot,\cdot)$ be the kernel functions for $X$ and $Y$ respectively. Then the resulted kernel matrices of $k_x$ and $k_y$ with respect to $\mathbf{x}_1,\ldots,\mathbf{x}_n$ and $\mathbf{y}_1,\ldots,\mathbf{y}_n$ are $K_x$ and $K_y$, whose $(i,j)$-entries are $K_x(i,j)=k_x(\mathbf{x}_i,\mathbf{x}_j)$ and $K_y(i,j)=k_y(\mathbf{y}_i,\mathbf{y}_j)$. Then CKA is defined as
\begin{equation}
    \text{CKA}(K_x,K_y)\coloneqq\frac{\text{tr}(K_x H K_y H)}{\sqrt{\text{tr}(K_x H K_x H)\text{tr}(K_y H K_x H)}},
\end{equation}
where $H = I_n-\frac{1}{n}\mathbf{1}\mathbf{1}^\text{T}$ is the centering matrix.

As for the kernels in CKA, we select linear kernel (i.e., $K_x=XX^\text{T}$, $K_y=YY^\text{T}$) \footnote{Linear kernel $k(\mathbf{x}_i,\mathbf{x}_j)=\mathbf{x}_i^\text{T}\mathbf{x}_j$} over Radial Basis Function (RBF) kernel \footnote{Radial Basis Function (RBF) kernel $k(\mathbf{x}_i,\mathbf{x}_j)=\exp(-\gamma||\mathbf{x}_i-\mathbf{x}_j||^2_2)$ with $\gamma>0$} from common kernels for the following reasons. First, experiments in \cite{kornblith2019similarity} manifest that linear and RBF kernels work equally well in similarity measurement of feature representations. Furthermore, it is recently validated that CKA based on an RBF kernel converges to linear CKA in the large-bandwidth limit \citep{9926163}. Hence, in our investigation we stick with linear CKA for computational efficiency, where the resulting linear CKA is 
\begin{equation}
    \begin{aligned}[b]
    \text{CKA}_{\text{linear}}(X,Y)&=\text{CKA}(XX^\text{T},YY^\text{T})\\
        &=\frac{\text{tr}(XX^\text{T}HYY^\text{T}H)}{\sqrt{\text{tr}(XX^\text{T}HXX^\text{T}H)\text{tr}(YY^\text{T}HYY^\text{T}H)}}\\
        &=\frac{\text{tr}(Y^\text{T}HXX^\text{T}HY)}{\sqrt{\text{tr}(X^\text{T}HXX^\text{T}HX)\text{tr}(Y^\text{T}HYY^\text{T}HY)}}\\
        &=\frac{||Y^\text{T}HX||^2_\text{F}}
        {||X^\text{T}HX||_\text{F}||Y^\text{T}HY||_\text{F}}.
\end{aligned}
\end{equation}

In our design, we measure the averaged CKA similarities according to the outputs from the same batch of test data. The range of CKA is between 0 and 1, and a higher CKA score means more similar paired features.

\section{Comparisons with Residual Connections}
\label{sec:apd_comparisons_rc}
\textbf{Remarks on self-mixture approaches in neural networks.} 
The goal of residual connections is to avoid exploding and vanishing gradients to facilitate the training of a single model \citep{he2016identity}, while cross-layer gradients aim to increase the layer similarities across a group of models that are jointly optimized in federated learning. Specifically, residual connections modify forward passes by adding the shallow-layer outputs to those of the deep layers. In contrast, cross-layer gradients operate on the gradients calculated by back-propagation. We present the distinct gradient outcomes of the two methods in the following.

\begin{figure}[htbp]
% \vskip 0.2in
\centering
\subfloat[Cross-layer gradients]
{\includegraphics[width=0.3\columnwidth]{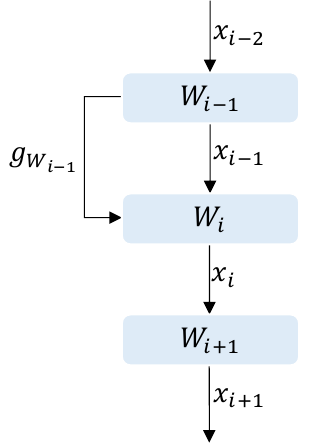}
\label{pic:cross_layer_gradient}}
\hfil
\subfloat[Residual connections]
{\includegraphics[width=0.3\columnwidth]{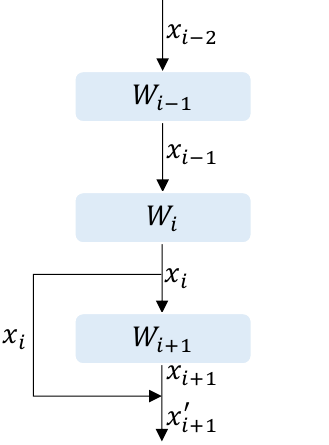}
\label{pic:residual_connection}}
\caption{Comparison of cross-layer gradients and residual connections}
\label{fig:cf_res_grad}
\vskip -0.2in
\end{figure}

Consider three consecutive layers of a feedforward neural network indexed by $i-1, i, i+1$. With a slight abuse of symbols, we use $f(\,\cdot\,;W_k)$ to denote the calculation in the $k$-th layer. Given the input $\mathbf{x_{i-2}}$ to layer $i-1$, the output from the previous layer becomes the input to the next layer, thus generating $\mathbf{x_{i-1}}, \mathbf{x_{i}}, \mathbf{x_{i+1}}$ sequentially.
\begin{equation}
\begin{aligned}
\mathbf{x_{i-1}}&=f(\mathbf{x_{i-2}}; W_{i-1})\\
\mathbf{x_{i}}&=f(\mathbf{x_{i-1}}; W_{i})\\
\mathbf{x_{i+1}}&=f(\mathbf{x_{i}}; W_{i+1})
\end{aligned}
\end{equation}
In the case of residual connections, there is an additional operation that directs $\mathbf{x_{i}}$ to $\mathbf{x_{i+1}}$, formulated as $\mathbf{x_{i+1}'}=\mathbf{x_{i+1}}+\mathbf{x_{i}}$. The gradient of $W_{i}$ is 
\begin{equation}
    \begin{aligned}
        g_{W_i}&=\frac{\partial loss}{\partial W_i}\\
        &=\frac{\partial loss}{\partial \mathbf{x_{i+1}'}}
        \cdot
        \frac{\partial \mathbf{x_{i+1}'}}{\partial \mathbf{x_i}}
        \cdot
        \frac{\partial \mathbf{x_i}}{\partial W_i}\\
        &=\frac{\partial loss}{\partial (\mathbf{x_{i+1}}+\mathbf{x_{i}})}
        \cdot \left(\frac{\partial \mathbf{x_{i+1}}}{\partial \mathbf{x_{i}}}+ \mathbb{I} \right)\cdot 
        \frac{\partial \mathbf{x_i}}{\partial W_i}\\
        &=\frac{\partial loss}{\partial (\mathbf{x_{i+1}}+\mathbf{x_{i}})}
        \cdot \left(\frac{\partial \mathbf{x_{i+1}}}{\partial W_i}+\frac{\partial \mathbf{x_{i}}}{\partial W_i}\right)
    \end{aligned}
\end{equation}

In the case of cross-layer gradients, the gradient of $W_{i}$ is 
\begin{equation}
    \begin{aligned}
        g_{W_i}&=\frac{\partial loss}{\partial W_i}+\frac{\partial loss}{\partial W_{i-1}}\\
        &=\frac{\partial loss}{\partial \mathbf{x}_{i+1}}\cdot\left(\frac{\partial\mathbf{x_{i+1}}}{\partial W_i}+\frac{\partial\mathbf{x_{i+1}}}{\partial W_{i-1}}\right)
    \end{aligned}
\end{equation}

We note that both residual connections and cross-layer gradients are subject to certain constraints. Residual connections require identical shapes for the layer outputs, while cross-layer gradients operate on the layer weights with the same shape.

\section{Proof of Theorem~\ref{thm:vec_solution}}
\label{sec:apd_proof}
This section demonstrates the details of the proof of Theorem~\ref{thm:vec_solution}. We will present the proof of Theorem~\ref{thm:vec_solution} in the vector form and the matrix form.

\subsection{Vector Form}
\label{subsec:vector_form}
We state the convex optimization problem Theorem~\ref{eq:primal_optim} in the vector form in the following,
\begin{equation}
\label{eq:apd_primal_optim}
\begin{aligned}
 \min_{g_{opt}} && ||g_{k}-g_{opt}||_{2}^{2}, \\
 s.t.\ && \langle g_{opt},g_{0}\rangle \geq 0.
\end{aligned}
\end{equation}

Because the superscript $t$ would not influence the proof of the theorem, we simplify the notation $g^t$ to $g$. We use Equation~\ref{eq:apd_primal_optim} instead of Equation~\ref{eq:primal_optim} to complete this proof. The Lagrangian of Equation~\ref{eq:apd_primal_optim} is shown as,
\begin{equation}
\label{eq:Lagrangian}
\begin{aligned}
 L(g_{opt},\lambda)=&(g_{k}-g_{opt})^T(g_{k}-g_{opt})-\lambda g_{opt}^Tg_0 \\
 =&g_{k}^Tg_{k}-g_{opt}^Tg_{k}-g_{k}^Tg_{opt}+g_{opt}^Tg_{opt}-\lambda g_{opt}^Tg_0 \\
 =&g_{k}^Tg_{k}-2g_{opt}^Tg_{k}+g_{opt}^Tg_{opt}-\lambda g_{opt}^Tg_0.
\end{aligned}
\end{equation}

Let $\frac{\partial L(g_{opt},\lambda)}{\partial g_{opt}} = 0$, we have 
\begin{equation}
\label{eq:primal_optim_point}
\begin{aligned}
g_{opt}=g_k+\lambda g_0/2,
\end{aligned}
\end{equation}
which is the optimum point for the primal problem Equation~\ref{eq:apd_primal_optim}. To get the Lagrange dual function $L(\lambda)=\inf_{g_{opt}} L(g_{opt},\lambda)$, we substitute $g_{opt}$ by $g_k+\lambda g_0/2$ in $L(g_{opt},\lambda)$. We have 
\begin{equation}
\label{eq:Lagrange_dual}
\begin{aligned}
 L(\lambda)=&g_{k}^Tg_{k}-2(g_k+\frac{\lambda g_0}{2})^Tg_{k}+(g_k+\frac{\lambda g_0}{2})^T(g_k+\frac{\lambda g_0}{2})-\lambda (g_k+\frac{\lambda g_0}{2})^Tg_0 \\
 =&g_{k}^Tg_{k}-2g_k^Tg_k-\lambda g_0^Tg_{k}+g_{k}^Tg_k+\frac{\lambda g_0^Tg_k}{2}+\frac{\lambda g_k^Tg_0}{2} + \frac{\lambda^2g_0^Tg_0}{4}-\lambda g_k^Tg_0-\frac{\lambda^2 g_0^Tg_0}{2} \\
 =&g_{k}^Tg_{k}-2g_k^Tg_k+g_{k}^Tg_k-\lambda g_0^Tg_{k}+\lambda g_0^Tg_k+\frac{\lambda^2g_0^Tg_0}{4}-\frac{\lambda^2 g_0^Tg_0}{2}-\lambda g_k^Tg_0 \\
 =&-\frac{g_0^Tg_0}{4}\lambda^2-g_k^Tg_0\lambda .
\end{aligned}
\end{equation}

Thus, the Lagrange dual problem is described as follows,
\begin{equation}
\label{eq:dual_optim}
\begin{aligned}
 \max_\lambda\ &L(\lambda)=-\frac{g_0^Tg_0}{4}\lambda^2-g_k^Tg_0\lambda,\\
 s.t.\ &\lambda \geq 0.
\end{aligned}
\end{equation}
$L(\lambda)$ is a quadratic function. Because $g_0^Tg_0\geq 0$, the maximum of $L(\lambda)$ is at the point $\lambda=-\frac{2b}{a}$ where $a=g_0^Tg_0$ and $b=g_k^Tg_0$ if we do not consider the constraint. It is clear that this convex optimization problem holds strong duality because it satisfies Slater's constraint qualification\citep{boyd2004convex}, which indicates that the optimum point of the dual problem Equation~\ref{eq:dual_optim} is also the optimum point for the primal problem Equation~\ref{eq:apd_primal_optim}. We substitute $\lambda$ by $-\frac{2b}{a}$ in Equation~\ref{eq:primal_optim_point}, and we have
\begin{equation}
\label{eq:apd_analytic_sol}
    g_{opt}=
\begin{cases}
g_k,& \text{if } b\geq 0, \\
g_k-\theta^t g_0, & \text{if } b<0,
\end{cases}
\end{equation}
where $\theta^t=\frac{b}{a}$, $a=(g_0)^Tg_0$ and $b=g_k^Tg_0$. We add the superscript $t$ to all gradients, and we finish the proof of Theorem~\ref{thm:vec_solution}.

\subsection{Matrix Form}
The proof of the matrix form is similar to Appendix~\ref{subsec:vector_form}. We update Equation~\ref{eq:apd_primal_optim} to the matrix form as follows,
\begin{equation}
\label{eq:matrix_primal_optim}
\begin{aligned}
 \min_{G_{opt}} && ||G_{k}-G_{opt}||_{F}^{2}, \\
 s.t.\ && \langle G_{opt},G_{0}\rangle \geq 0.
\end{aligned}
\end{equation}
Similar to Equation~\ref{eq:Lagrangian}, the Lagragian of Equation~\ref{eq:matrix_primal_optim} is,
\begin{equation}
\label{eq:matrix_Lagrangian}
\begin{aligned}
 L(G_{opt},\lambda)=tr(G_{k}^TG_{k})-tr(G_{opt}^TG_{k})
 -tr(G_{k}^TG_{opt})+tr(G_{opt}^TG_{opt})-\lambda tr(G_{opt}^TG_{0}),
\end{aligned}
\end{equation}
where $tr(A)$ means the trace of the matrix $A$. 
We can obtain the optimum point for Equation~\ref{eq:matrix_primal_optim} according to $\frac{\partial L(G_{opt},\lambda)}{\partial G_{opt}} = 0$. We have 
\begin{equation}
\label{eq:matrix_primal_optim_point}
\begin{aligned}
G_{opt}=G_k+\lambda G_0/2.
\end{aligned}
\end{equation}

Similar to the analysis in Appendix~\ref{subsec:vector_form} and Equation~\ref{eq:Lagrange_dual}, we get the Lagrange dual problem as follows,
\begin{equation}
\label{eq:matirx_dual_optim}
\begin{aligned}
 \max_\lambda\ &L(\lambda)=-\frac{tr(G_{0}^TG_{0})}{4}\lambda^2 -tr(G_{k}^TG_{0})\lambda, \\
 s.t.\ &\lambda \geq 0,
\end{aligned}
\end{equation}
where $tr(G_{0}^TG_{0})\geq 0$. Following the same analysis in Appendix~\ref{subsec:vector_form}, we have
\begin{equation}
\label{eq:matrix_analytic_sol}
    G_{opt}=
\begin{cases}
G_{k},& \text{if } b\geq 0, \\
G_{k}-\theta^t G_{0}, & \text{if } b<0,
\end{cases}
\end{equation}
where $\theta^t=\frac{b}{a}$, $a=tr(G_0^TG_0)$ and $b=tr(G_k^TG_0)$. At last, we have finished the proof of the matrix form of Theorem~\ref{thm:vec_solution}.

\section{Proof of Convergence Analysis}
\label{subsec:apd_convergence_proof}
We show the details of convergence analysis for cross-layer gradients. $W^{l_i}_{t,e}$ are the weights from the layers which need cross-layer gradients at round $t$ of the local step $e$. To simplify the notations, we use $W_{t,e}$ instead of $W^{l_i}_{t,e}$.

\begin{lemma}
\label{lem:per_round_progress}
\text{(Per Round Progress.)}
Suppose our functions satisfy Assumption~\ref{ass:L-smooth} and Assumption~\ref{ass:unbiased_gradients}. The expectation of a loss function of any arbitrary clients at communication round $t$ after $E$ local steps are bounded as,
\begin{equation}
\label{eq:per_round_progress_2}
\begin{aligned}
\mathbb{E}[L_{t,E-1}] \leq \mathbb{E}[L_{t, 0}] -(\eta-\frac{L\eta^2}{2})\sum_{e=0}^{E-1} ||\nabla L_{t, e}||^2 + \frac{LE\eta^2}{2}\sigma^2.
\end{aligned}
\end{equation}
\end{lemma}

\textit{Proof.}

Considering an arbitrary client, we omit the client index $i$ in this lemma. Let $W_{t,e+1}=W_{t,e}-\eta g_{t,e}$, we have 
\begin{equation}
\label{eq:per_round_progress_1}
\begin{aligned}
L_{t,e+1}&\leq L_{t, e} + \langle \nabla L_{t, e}, W_{t, e+1}-W_{t, e}\rangle + \frac{L}{2}||W_{t, e+1}-W_{t, e}||^2 \\
&\leq L_{t, e} -\eta \langle \nabla L_{t, e}, g_{t,e}\rangle + \frac{L}{2}||\eta g_{t,e}||^2,
\end{aligned}
\end{equation}
where Equation~\ref{eq:per_round_progress_1} follows Assumption~\ref{ass:L-smooth}. We take expectation on both sides of Equation~\ref{eq:per_round_progress_1}, then
\begin{equation}
\label{eq:per_round_progress_3}
\begin{aligned}
\mathbb{E}[L_{t,e+1}] &\leq \mathbb{E}[L_{t, e}] -\eta \mathbb{E}[\langle \nabla L_{t, e}, g_{t,e}\rangle] + \frac{L}{2}\mathbb{E}[||\eta g_{t,e}||^2] \\ 
&= \mathbb{E}[L_{t, e}] -\eta ||\nabla L_{t, e}||^2 + \frac{L\eta^2}{2}\mathbb{E}[||g_{t,e}||^2] \\
&\overset{(a)}{=} \mathbb{E}[L_{t, e}] -\eta ||\nabla L_{t, e}||^2 + \frac{L\eta^2}{2}(\mathbb{E}[||g_{t,e}||]^2 + Var(||g_{t,e}||)) \\
&= \mathbb{E}[L_{t, e}] -\eta ||\nabla L_{t, e}||^2 + \frac{L\eta^2}{2}(||\nabla L_{t, e}||^2 + Var(||g_{t,e}||)) \\
&\overset{(b)}{\leq} \mathbb{E}[L_{t, e}] -(\eta-\frac{L\eta^2}{2}) ||\nabla L_{t, e}||^2 + \frac{L\eta^2}{2}\sigma^2,
\end{aligned}
\end{equation}
where (a) follows $Var(X)=\mathbb{E}[X^2]-\mathbb{E}^2[X]$, and (b) is Assumption~\ref{ass:unbiased_gradients}. Telescoping local step 0 to $E-1$, we have
\begin{equation}
\label{eq:per_round_progress_4}
\begin{aligned}
\mathbb{E}[L_{t,E-1}] \leq \mathbb{E}[L_{t, 0}] -(\eta-\frac{L\eta^2}{2})\sum_{e=0}^{E-1} ||\nabla L_{t, e}||^2 + \frac{LE\eta^2}{2}\sigma^2,
\end{aligned}
\end{equation}
then we finish the proof of Lemma~\ref{lem:per_round_progress}.

\begin{lemma}
\label{lem:Bound_client_drift}
\text{(Bound Client Dirft.)}
Suppose our functions satisfy Assumption~\ref{ass:unbiased_gradients}, Assumption~\ref{ass:Bounded_gradients} and Assumption~\ref{ass:Bounded_covar}. After each aggregation, the updates, $\Delta W$, for the layers need cross-layer gradients have bounded drift:
\begin{equation}
\label{eq:bound_client_drift}
\begin{aligned}
\mathbb{E}[||\Delta W||^2] \leq 2\eta^2(2\rho^2+\sigma^2+\Gamma).
\end{aligned}
\end{equation}
\end{lemma}

\textit{Proof.}

We have $W_{t+1,0}-W_{t,E-1}=\Delta W=\eta (g_{l_0}+g_{l_i}), \forall l_i$ need cross-layer gradients. Because all gradients are in the same aggregation round, we omit the time subscript in this proof process. Since $\eta$ is a constant, we also simplify it. $g_{l_0}$ and $g_{l_i}$ the gradients from the same client, indicating that they are dependent, then
\begin{equation}
\label{eq:bound_client_drift_1}
\begin{aligned}
||\Delta W||^2&=||g_{l_0}+g_{l_i}||^2 \\
&\overset{(c)}{\leq} ||g_{l_0}||^2 + 2||\langle g_{l_0},g_{l_i}\rangle|| + ||g_{l_i}||^2,
\end{aligned}
\end{equation}
where (c) is Cauchy–Schwarz inequality. We take the expectation on both sides, then
\begin{equation}
\label{eq:bound_client_drift_2}
\begin{aligned}
\mathbb{E}[||\Delta W||^2] &\leq \mathbb{E}[||g_{l_0}||^2] + 2\mathbb{E}[||\langle g_{l_0},g_{l_i}\rangle||] + \mathbb{E}[||g_{l_i}||^2] \\
&\overset{(a)}{=} \mathbb{E}[||g_{l_0}||]^2 + Var(||g_{l_0}||) + \mathbb{E}[||g_{l_i}||]^2 + Var(||g_{l_i}||) + 2\mathbb{E}[||\langle g_{l_0},g_{l_i}\rangle||] \\
&\overset{(d)}{\leq} 2(\rho^2+\sigma^2) + 2\mathbb{E}[||g_{l_0},g_{l_i}||] \\
&\overset{(e)}{=} 2(\rho^2+\sigma^2) + 2(Cov(g_{l_0},g_{l_i})+\mathbb{E}[||g_{l_0}||]\mathbb{E}[||g_{l_i}||]) \\
&\overset{(f)}{\leq} 2(\rho^2+\sigma^2) + 2(\Gamma+\rho^2) \\
&=4\rho^2+2\sigma^2+2\Gamma,
\end{aligned}
\end{equation}
where (d) follows assumption Assumption~\ref{ass:unbiased_gradients} and Assumption~\ref{ass:Bounded_gradients}, (e) follows the covariance formula, and (f) follows assumption Assumption~\ref{ass:Bounded_covar}. We put back $\eta^2$ to the final step of Equation~\ref{eq:bound_client_drift_2}. At last, we complete the proof of Lemma~\ref{lem:Bound_client_drift}.

\subsection{Proof of Theorem~\ref{thm:per_round_drift} and Theorem~\ref{thm:non_convex_convergence}}
We state Theorem~\ref{thm:per_round_drift} again in the following,

\textit{(Per round drift) Supposed Assumption~\ref{ass:L-smooth} to Assumption~\ref{ass:Bounded_covar} are satisfied, the loss function of an arbitrary client at round $t+1$ is bounded by,}
\begin{equation}
\label{eq:per_round_drift_apd}
\begin{aligned}
\mathbb{E}[L_{t+1,0}]\leq \mathbb{E}[L_{t,0}]-(\eta-\frac{L\eta^2}{2})\sum_{e=0}^{E-1}||\nabla L_{t,e}||^2 + \\
\frac{LE\eta^2}{2}\sigma^2 + 2\eta(\Gamma+\rho^2)+L\eta^2(2\rho^2+\sigma^2+\Gamma).
\end{aligned}
\end{equation}

\textit{Proof.}

Following the Assumption~\ref{ass:L-smooth}, we have
\begin{equation}
\label{eq:per_round_drift_proof_1}
\begin{aligned}
L_{t+1,0}&\leq L_{t, E-1} + \langle \nabla L_{t, E-1}, W_{t+1, 0}-W_{t, E-1}\rangle + \frac{L}{2}||W_{t+1, 0}-W_{t, E-1}||^2 \\
&= L_{t, E-1} +\eta \langle \nabla L_{t, E-1}, g_{l_0}+g_{l_1}\rangle + \frac{L}{2}\eta^2||\Delta W||^2.
\end{aligned}
\end{equation}
Taking the expectation on both sides, we obtain
\begin{equation}
\label{eq:per_round_drift_proof_2}
\begin{aligned}
\mathbb{E}[L_{t+1,0}] = \mathbb{E}[L_{t, E-1}] +\eta \mathbb{E}[\langle \nabla L_{t, E-1}, g_{l_0}+g_{l_1}\rangle] + \frac{L}{2}\eta^2\mathbb{E}[||\Delta W||^2].
\end{aligned}
\end{equation}
The first item is Lemma~\ref{lem:per_round_progress}, and the third item is Lemma~\ref{lem:Bound_client_drift}. We consider the second item $\mathbb{E}[\langle \nabla L_{t, E-1}, g_{l_0}+g_{l_1}\rangle]$ in the following, then,
\begin{equation}
\label{eq:per_round_drift_proof_3}
\begin{aligned}
\mathbb{E}[\langle \nabla L_{t, E-1}, g_{l_0}+g_{l_1}\rangle] &= \mathbb{E}[\nabla L_{t, E-1} g_{l_0}] + \mathbb{E}[\nabla L_{t, E-1} g_{l_1}] \\ 
&\overset{(e)}{=} Cov(\nabla L_{t, E-1}, g_{l_0}) + \mathbb{E}[||\nabla L_{t, E-1}||]\mathbb{E}[||g_{l_0}||] + \\
& Cov(\nabla L_{t, E-1}, g_{l_1}) + \mathbb{E}[||\nabla L_{t, E-1}||]\mathbb{E}[||g_{l_1}||] \\
&\overset{(f)}{\leq} 2(\Gamma+\rho^2).
\end{aligned}
\end{equation}
Combining two lemmas and Equation~\ref{eq:per_round_drift_proof_3}, we have
\begin{equation}
\label{eq:per_round_drift_proof_4}
\begin{aligned}
\mathbb{E}[L_{t+1,0}] \leq \mathbb{E}[L_{t, 0}] -(\eta-\frac{L\eta^2}{2})\sum_{e=0}^{E-1} ||\nabla L_{t, e}||^2 + \frac{LE\eta^2}{2}\sigma^2 + 2\eta(\Gamma + \rho^2)+L\eta^2(2\rho^2+\sigma^2+\Gamma),
\end{aligned}
\end{equation}
then we finish the proof of Theorem~\ref{thm:per_round_drift}.

For Theorem~\ref{thm:non_convex_convergence}, we consider the sum of the second term to the last term in Equation~\ref{eq:per_round_drift_proof_4} to be smaller than 0, i.e.,
\begin{equation}
\label{eq:non_convex_convergence_proof}
\begin{aligned}
-(\eta-\frac{L\eta^2}{2})\sum_{e=0}^{E-1} ||\nabla L_{t, e}||^2 + \frac{LE\eta^2}{2}\sigma^2 + 2\eta(\Gamma + \rho^2)+L\eta^2(2\rho^2+\sigma^2+\Gamma) < 0,
\end{aligned}
\end{equation}
then, we have
\begin{equation}
\label{eq:non_convex_convergence_proof_1}
\begin{aligned}
\eta < \frac{2\sum_{e=0}^{E-1}||\nabla L_{t,e}||^2-4(\Gamma+\rho^2)}{L(\sum_{e=0}^{E-1}||\nabla L_{t,e}||^2+E\rho^2+2(2\rho^2+\sigma^2+\Gamma)}.
\end{aligned}
\end{equation}
We finish the proof of Theorem~\ref{thm:non_convex_convergence}.

\subsection{Proof of Theorem~\ref{thm:Non_convex_rate}}
Telescoping the communication rounds from $t=0$ to $t=T-1$ with the local step from $e=0$ to $e=E-1$ on the expectation on both sides of Equation~\ref{eq:per_round_drift_proof_4}, we have
\begin{equation}
\label{eq:non_convex_rate_proof}
\begin{aligned}
\frac{1}{TE}\sum_{t=0}^{T-1}\sum_{e=0}^{E-1}||\nabla L_{t,e}||^2 \leq \frac{\frac{2}{TE}\sum_{t=0}^{T-1}(\mathbb{E}[L_{t,0}]-\mathbb{E}[L_{t+1,0}])+L\eta^2\sigma^2 + 4\eta(\Gamma+\rho^2) +2L\eta^2(2\rho^2+\sigma^2+\Gamma)}{2\eta-L\eta^2}.
\end{aligned}
\end{equation}

Given any $\epsilon > 0$, let 
\begin{equation}
\label{eq:non_convex_rate_proof_1}
\begin{aligned}
\frac{\frac{2}{TE}\sum_{t=0}^{T-1}(\mathbb{E}[L_{t,0}]-\mathbb{E}[L_{t+1,0}])+L\eta^2\sigma^2 + 4\eta(\Gamma+\rho^2) +2L\eta^2(2\rho^2+\sigma^2+\Gamma)}{2\eta-L\eta^2} \leq \epsilon,
\end{aligned}
\end{equation}
and we denote $\kappa=L_0-L^*$, then Equation~\ref{eq:non_convex_rate_proof_1} becomes
\begin{equation}
\label{eq:non_convex_rate_proof_2}
\begin{aligned}
\frac{\frac{2\kappa}{TE}+L\eta^2\sigma^2 + 4\eta(\Gamma+\rho^2) +2L\eta^2(2\rho^2+\sigma^2+\Gamma)}{2\eta-L\eta^2} \leq \epsilon,
\end{aligned}
\end{equation}
because $\sum_{t=0}^{T-1}(\mathbb{E}[L_{t,0}]-\mathbb{E}[L_{t+1,0}]) \leq \kappa$. We consider $T$ in Equation~\ref{eq:non_convex_rate_proof_2}, i.e.,
\begin{equation}
\label{eq:non_convex_rate_proof_3}
\begin{aligned}
T\geq \frac{2\kappa}{E\eta((2-L\eta)\epsilon-3L\eta\sigma^2-
2(2+L\eta)\Gamma-4(1+L\eta)\rho^2)},
\end{aligned}
\end{equation}
then, we have 
\begin{equation}
\label{eq:non_convex_rate_proof_4}
\begin{aligned}
\frac{1}{TE}\sum_{t=0}^{T-1}\sum_{e=0}^{E-1}||\nabla L_{t,e}||^2\leq \epsilon,
\end{aligned}
\end{equation}
when
\begin{equation}
\label{eq:non_convex_rate_proof_5}
\begin{aligned}
\eta < \frac{2\epsilon-4(\Gamma+\rho^2)}{L(\epsilon+E\rho^2+2(2\rho^2+\sigma^2+\Gamma)}.
\end{aligned}
\end{equation}
We complete the proof of Theorem~\ref{thm:Non_convex_rate}.

\section{More Related Works}

% FedAvg, FedProx, FedAsyn (from FedHe related work)
\subsection{Federated Learning.}
In 2017, Google proposed a novel machine learning technique, i.e., Federated Learning (FL), to organize collaborative computing among edge devices or servers \citep{mcmahan2017communication}. It enables multiple clients to collaboratively train models while keeping training data locally, facilitating privacy protection. Various synchronous or asynchronous FL schemes have been proposed and achieved good performance in different scenarios. For example, FedAvg \citep{mcmahan2017communication} takes a weighted average of the models trained by local clients and updates the local models iteratively. FedProx \citep{li2018federated} generalized and re-parametrized FedAvg, guaranteeing the convergence when learning over non-iid data. FedAsyn \citep{xie2019asynchronous} employed coordinators and schedulers to achieve an asynchronous training process.

\subsection{Heterogeneous Models.}
The clients in homogeneous federated learning frameworks have identical neural network architectures, while the edge devices or servers in real-world settings show great diversity. They usually have different memory and computation capabilities, making it difficult to deploy the same machine-learning model in all the clients. Therefore, researchers have proposed various methods supporting heterogeneous models in the FL environment.

\textbf{Knowledge Distillation.} Knowledge distillation (KD) \citep{hinton2015distilling} was proposed by Hinton et al., aiming to train a student model with the knowledge distilled from a teacher model, which becomes an important research area in Machine Learning \citep{Gou_2021, 10.1145/3610873}. 
Inspired by the knowledge distillation, several studies\citep{li2019fedmd, li2021fedh2l, he2020group} are proposed to address the system heterogeneity problem. In FedMD\citep{li2019fedmd}, the clients distill and transmit logits from a large public dataset, which helps them learn from both logits and private local datasets. In RHFL \citep{fang2022robust}, the knowledge is distilled from the unlabeled dataset and the weights of clients are computed by the symmetric cross-entropy loss function.
Unlike the aforementioned methods, data-free KD is a new approach to completing the knowledge distillation process without the training data. The basic idea is to optimize noise inputs to minimize the distance to prior knowledge\citep{nayak2019zero}. Chen et al.\citep{chen2019data} train Generative Adversarial Networks (GANs)\citep{goodfellow2014generative} to generate training data for the entire KD process utilizing the knowledge distilled from the teacher model. In FedHe\citep{fedhe2021}, a server directly averages the logits transmitted from clients. FedGen\citep{zhu2021data} adopts a generator to simulate the prior knowledge from all the clients, which is used along with the private data from clients in local training. In FedGKT\citep{he2020group}, a neural network is separated into two segments, one held by clients, the other preserved in a server, in which the features and logits from clients are sent to the server to train the large model. In Felo \citep{chan2022exploiting}, the representations from the intermediate layers are the knowledge instead of directly using the logits.

\textbf{Public or Generated Data.} In FedML \citep{shen2020federated}, latent information from homogeneous models is applied to train heterogeneous models. FedAUX \citep{sattler2021fedaux} initialized heterogeneous models by unsupervised pre-training and unlabeled auxiliary data. 
FCCL \citep{huang2022learn} calculate a cross-correlation matrix according to the global unlabeled dataset to exchange knowledge.
However, these methods require a public dataset. The server might not be able to collect sufficient data due to data availability and privacy concerns.

\textbf{Model Compression.} Although HeteroFL \citep{diao2021heterofl} derives local models with different sizes from one large model, the architectures of local and global models still have to share the same model architecture, and it is inflexible that all models have to be retrained when the best participant joins or leaves the FL training process. Federated Dropout \citep{caldas2018expanding} randomly selects sub-models from the global models following the dropout way.
SlimFL \citep{baek2022joint} incorporated width-adjustable slimmable neural network (SNN) architectures\citep{yu2019universally} into FL which can tune the widths of local neural networks. FjORD \citep{horvath2021fjord} tailored model widths to clients' capabilities by leveraging Ordered Dropout and a self-distillation methodology. 
FedRoLex \citep{alam2022fedrolex} proposes a rolling sub-model extraction scheme to adapt to the heterogeneous model environment.
However, similar to HeteroFL, they only vary the number of parameters for each layer.

\section{Problem Formulation}
In this section, we introduce federated learning with model heterogeneity. Federated learning aims to foster collaboration with clients to jointly train a shared global model while preserving the privacy of their local data. However, in the context of model heterogeneity, it becomes challenging to maintain the same architectures across all clients. Specifically, we consider a set of physical resources denoted as $\{R_i\}_{i=1}^n$, where $R_i$ represents the available resources of client $i$. For each local client model $\{M_i\}_{i=1}^n$, the resource requirement $R(M_i)$ must be smaller than or equal to the available resources of client $i$, i.e. $\{R(M_i)\leq R_i\}_{i=1}^n$. To satisfy this constraint, the client models have varying sizes and architectures. Let $w_i$ denote the weights of the client model $M_i$, and $f(x,w_i)$ represent the forward function of model $M_i$ with input $x$. Moreover, each client has a local dataset $D_i=\{(x_{k,i}, y_{k,i})|k\in \{1,2,...,|D_i|\}\}$, where $|D_i|$ signifies the size of a dataset $D_i$. The loss function $l_i$ of client $i$ is shown as follows,
\begin{equation}
\label{eq:apd_FL_local_objective}
\begin{aligned}
 \min_w&\ & l_i(w_i)=\frac{1}{|D_i|}\sum_{k=1}^{|D_i|}{l_{CSE}(f(x_k;w_i), y_k)},
\end{aligned}
\end{equation}
where $l_{CSE}$ is a cross-entropy function. Moreover, if we denote $K=\sum_{i=1}^n |D_i|$ as the total size of all local datasets, the global optimization problem is,
\begin{equation}
\label{eq:apd_FL_global_objective}
\begin{aligned}
 \min_{w_1, w_2, ..., w_n}\ L(w_1, ..., w_n)=\sum_{i=1}^n\frac{|D_i|}{K}{l_i(w_i)},
\end{aligned}
\end{equation}
where the optimized model weights $\{w_1, w_2, ..., w_n\}$ are the parameters from ${M_i}_{i=1}^n$. In our case, $\{w_1, w_2, ..., w_n\}$ are split from the server model weight $w_s$ from the server model $M_s$. The goal of our paper is to propose a method that can effectively optimize Equation~\ref{eq:apd_FL_global_objective}.

\section{Configurations and More Results of the Case Study}
\label{apd:case_study}
\subsection{Configurations}
In the case study, we have five ResNet models which are stage splitting from ResNet26, resulting in ResNet10, ResNet14, ResNet18, ResNet22, and ResNet26. Five ViTs models are ViT-S/8, ViT-S/9, ViT-S/10, ViT-S/11, ViT-S/12, the results from the layer splitting of ViT-S/12. The model prototypes are the same as the experiment settings. To quantify a model's degree of bias towards its local dataset, we use CKA similarities among the clients based on the outputs from the same stages in ResNet (ResNets of different sizes always contain four stages) and the outputs from the same layers in Vision Transformers (ViTs) \citep{dosovitskiy2020image}. Specifically, we measure the averaged CKA similarities according to the outputs from the same batch of test data. The range of CKA is between 0 and 1, and a higher CKA score means more similar paired features. We train FedAvg under three settings: IID with the homogeneous setting, Non-IID with the homogeneous setting, and Non-IID with the heterogeneous setting. FedAvg only aggregates gradients from the models sharing the same architectures under the heterogeneous model setting \citep{lin2020ensemble}.
For ResNets, we conduct training 100 communication rounds, while only 20 rounds for ViTs. The local training epochs for clients are five for all settings. We use Adam\citep{KingmaB14} optimizer with default parameter settings for all client models, and the batch size is 64.
We use two small federated scales. One is ten clients deployed the same model architecture (ResNet18 for ResNets and ViT-S/12 for ViTs), which is called a homogeneous setting. The other is ten clients with five different model architectures, which is a heterogeneous setting. This setting means that we have five groups whose architectures are heterogeneous, but the clients belonging to the same group have the same architectures.

\begin{figure}[htbp]
\vskip -0.2in
\centering
\subfloat[Similarity of gradients in Block 7.]
{\includegraphics[width=0.23\columnwidth]{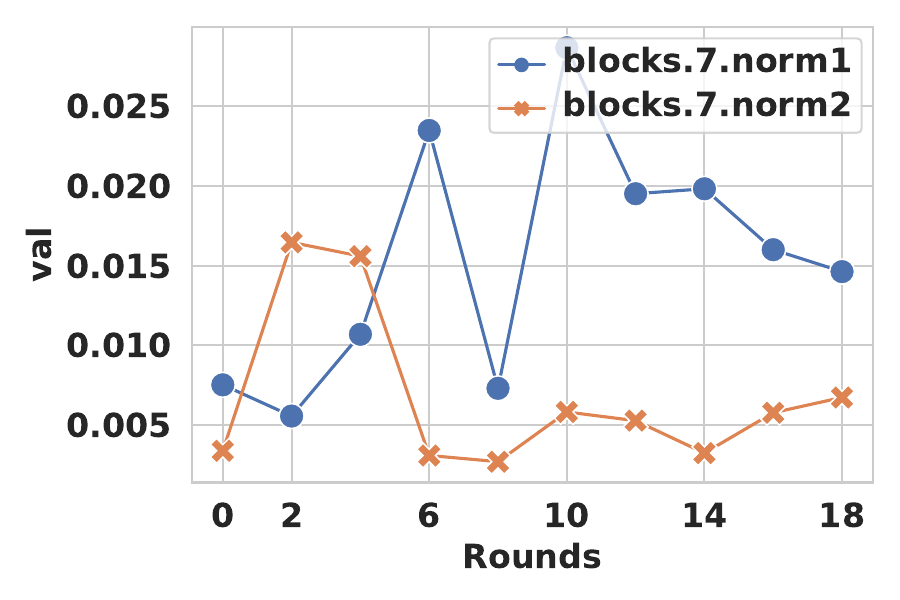}
\label{fig:apd_Analysis_Gradient_cross_environments_layer7}}
\hfil
\subfloat[Similarity of gradients in Block 11.]
{\includegraphics[width=0.23\columnwidth]{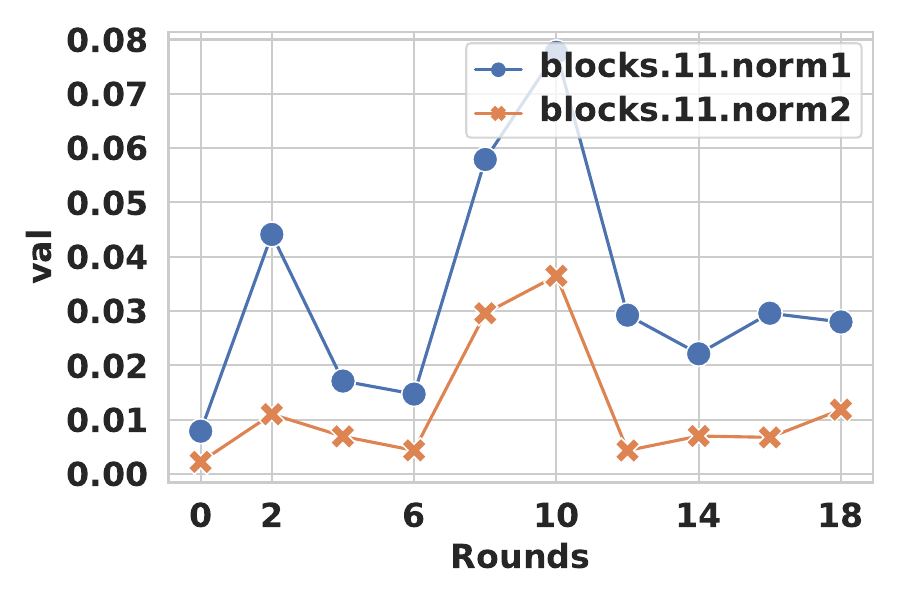}
\label{fig:apd_Analysis_Gradient_cross_environments_layer11}}
\hfil
\subfloat[The relations between CKA and accuracy in stage1.]
{\includegraphics[width=0.23\columnwidth]{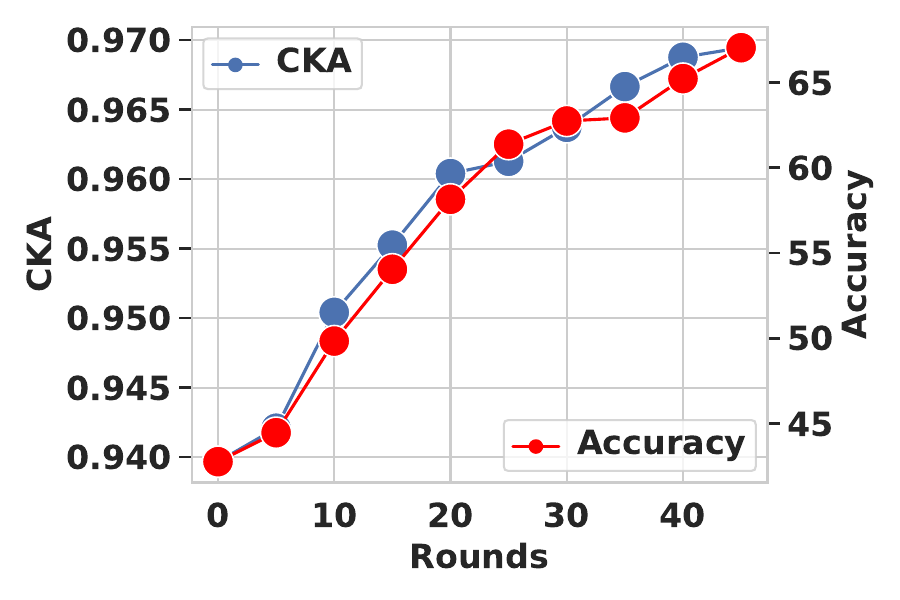}
\label{fig:apd_Analysis_CKA_acc_relation_stage1}}
\hfil
\subfloat[The relations between CKA and accuracy in stage2.]
{\includegraphics[width=0.23\columnwidth]{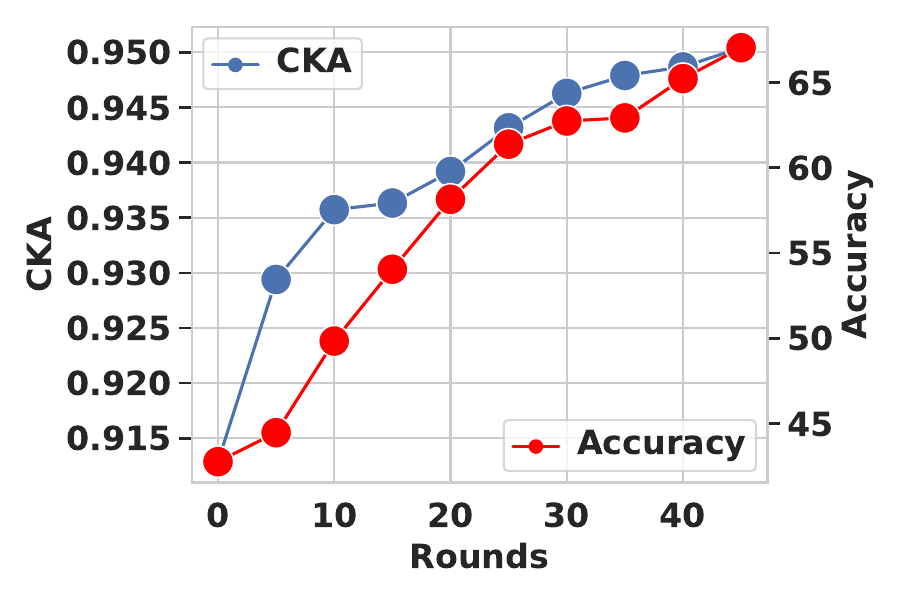}
\label{fig:apd_Analysis_CKA_acc_relation_stage2}}
% \vskip -0.07in
\caption{Cross-environment similarity and more results between accuracy and CKA similarity. (a) and (b): Cross-environment similarity from Block 7 and Block 11 from ViTs. (c) and (d): The positive relation between stage1 and stage2.}
\label{fig:apd_Analysis_Gradient_cross_environments_vit_CKA_accuracy_stage12}
\vskip -0.1in
\end{figure}

\begin{figure}[htbp]
\vskip -0.2in
\centering
\subfloat[Block7 norm1 in IID with homo.]
{\includegraphics[width=0.23\columnwidth]{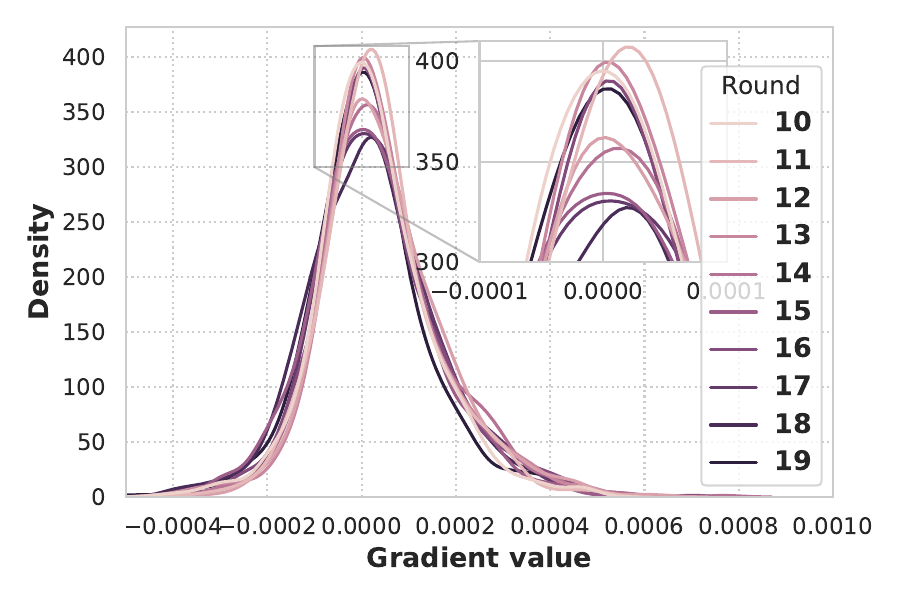}
\label{fig:apd_Analysis_Gradient_Dis_IID_FedAvg_layer7_norm1_1020}}
\hfil
\subfloat[Block7 norm2 in IID with homo.]
{\includegraphics[width=0.23\columnwidth]{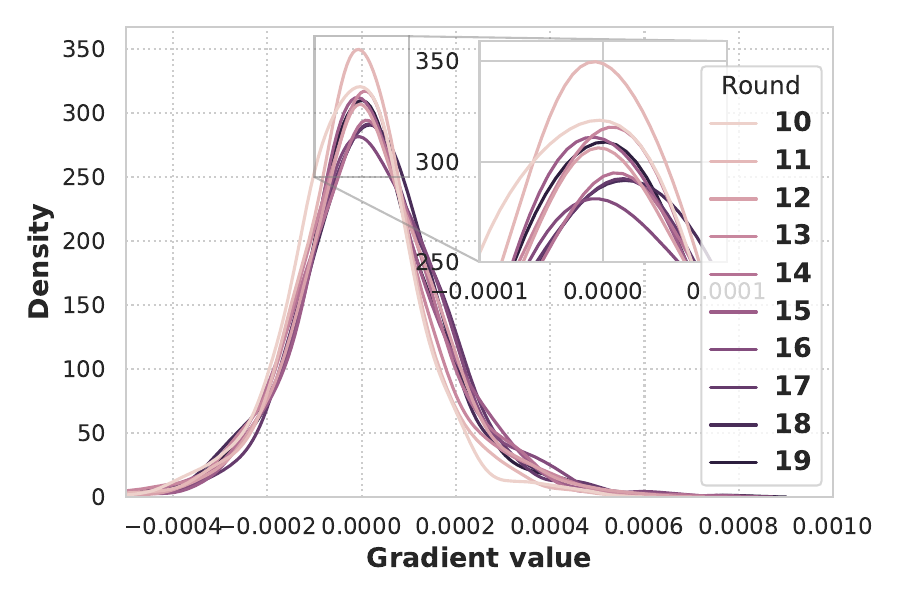}
\label{fig:apd_Analysis_Gradient_Dis_IID_FedAvg_layer7_norm2_1020}}
\hfil
\subfloat[Block11 norm1 in IID with homo.]
{\includegraphics[width=0.23\columnwidth]{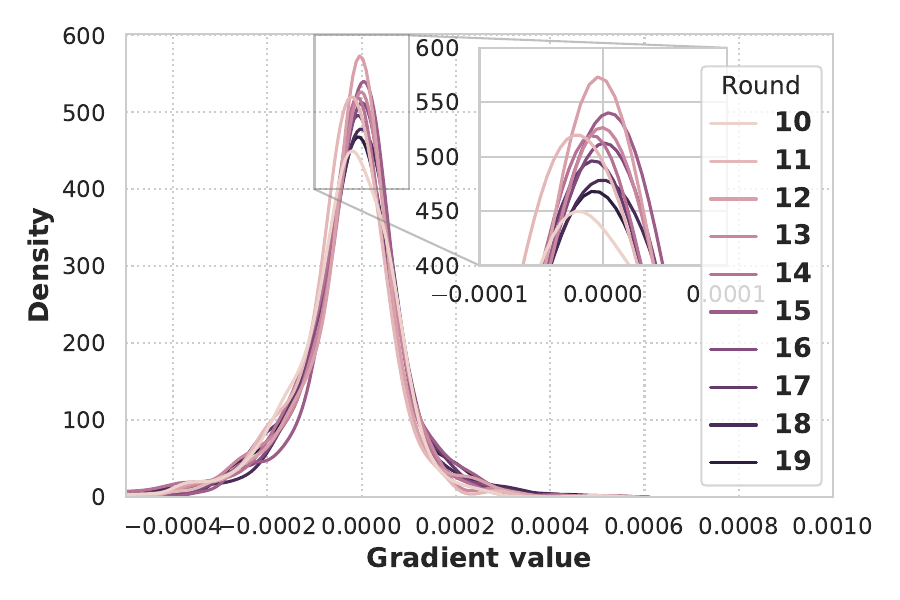}
\label{fig:apd_Analysis_Gradient_Dis_IID_FedAvg_layer11_norm1_1020}}
\hfil
\subfloat[Block11 norm2 in IID with homo.]
{\includegraphics[width=0.23\columnwidth]{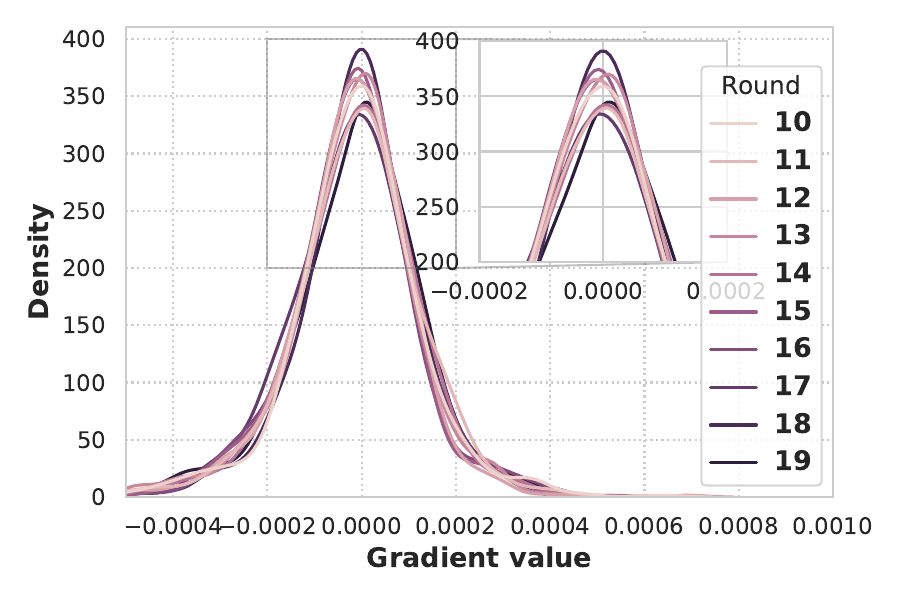}
\label{fig:apd_Analysis_Gradient_Dis_IID_FedAvg_layer11_norm2_1020}}
\vskip -0.07in
\caption{The gradient distributions from round 10 to 20 of ViTs in IID with homo.}
\label{fig:apd_Analysis_Gradient_Dis_IID_FedAvg_vit_1020}
\vskip -0.1in
\end{figure}

\begin{figure}[htbp]
\vskip -0.2in
\centering
\subfloat[Block7 norm1 in Non-IID with hetero.]
{\includegraphics[width=0.23\columnwidth]{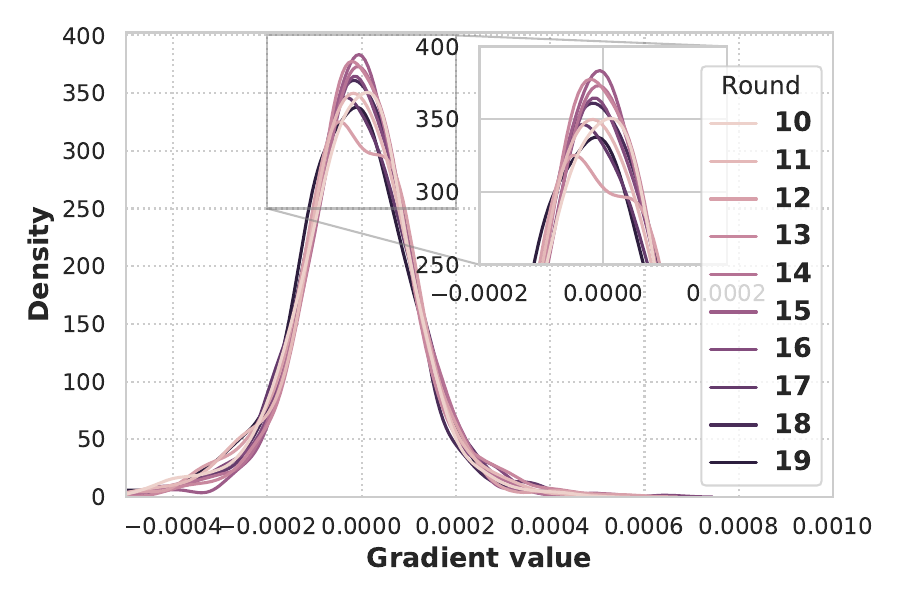}
\label{fig:apd_Analysis_Gradient_Dis_Hetero_FedAvg_layer7_norm1_1020}}
\hfil
\subfloat[Block7 norm2 in Non-IID with hetero.]
{\includegraphics[width=0.23\columnwidth]{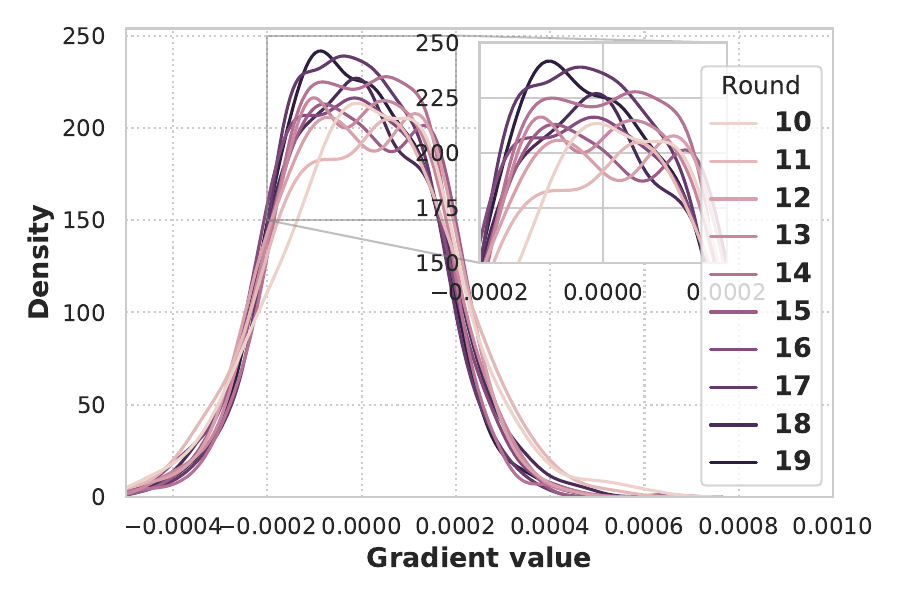}
\label{fig:apd_Analysis_Gradient_Dis_Hetero_FedAvg_layer7_norm2_1020}}
\hfil
\subfloat[Block11 norm1 in Non-IID with hetero.]
{\includegraphics[width=0.23\columnwidth]{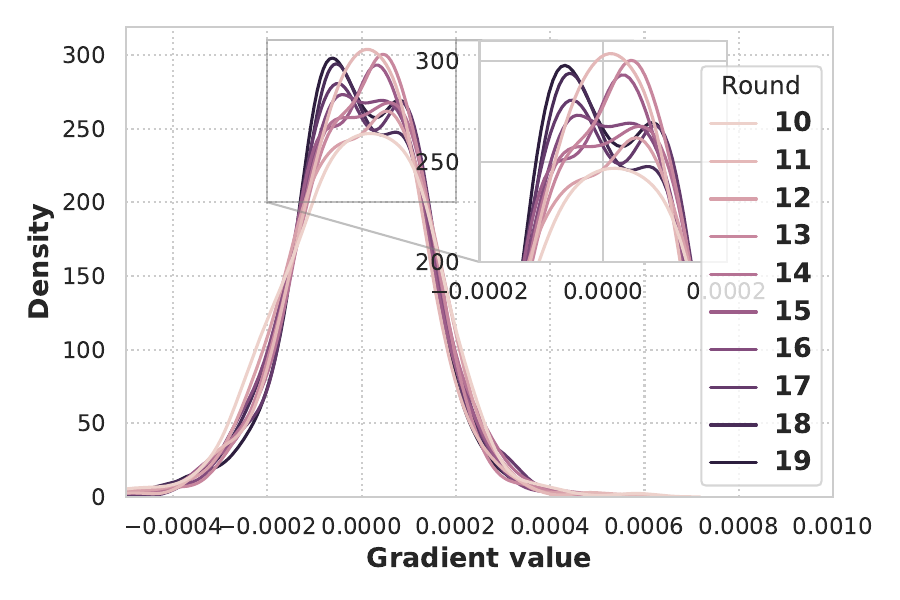}
\label{fig:apd_Analysis_Gradient_Dis_Hetero_FedAvg_layer11_norm1_1020}}
\hfil
\subfloat[Block11 norm2 in Non-IID with hetero.]
{\includegraphics[width=0.23\columnwidth]{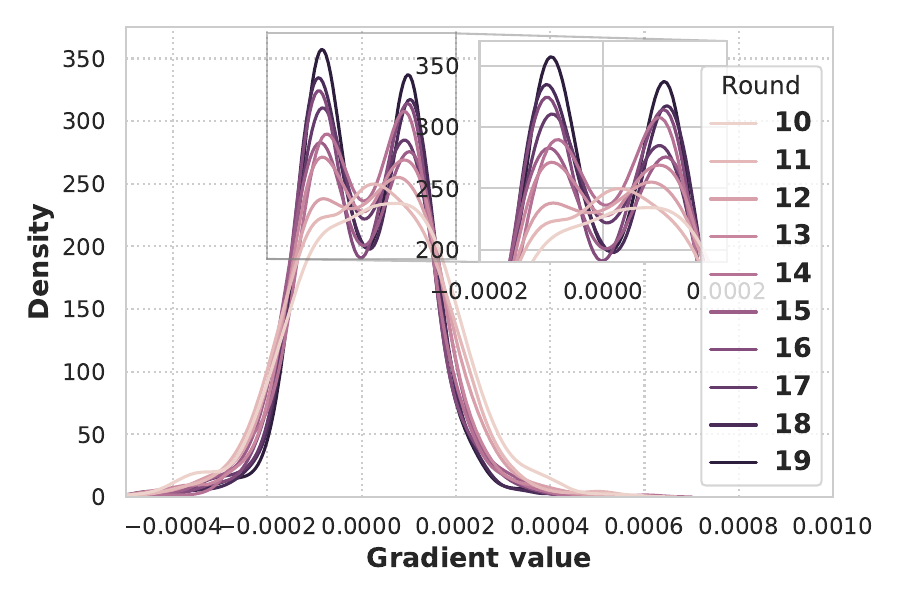}
\label{fig:apd_Analysis_Gradient_Dis_Hetero_FedAvg_layer11_norm2_1020}}
\vskip -0.07in
\caption{The gradient distributions from round 10 to 20 of ViTs in Non-IID with hetero.}
\label{fig:apd_Analysis_Gradient_Dis_Hetero_FedAvg_vit_1020}
\vskip -0.1in
\end{figure}

\subsection{Relations between CKA and Accuracy.}
In this subsection, we continue to describe more results about the relations between CKA similarity and accuracy. Similar to Figure~\ref{fig:Analysis_CKA_acc_relation} from stage0, Figure~\ref{fig:apd_Analysis_CKA_acc_relation_stage1} and Figure~\ref{fig:apd_Analysis_CKA_acc_relation_stage2} show the positive relations between CKA and accuracy from stage1 and stage2.

\subsection{Gradient Analysis for ViTs}
\label{subsec:apd_gradients_analysis_ViTs}
Similar to the gradient analyses conducted for ResNets, we have performed the analysis of gradient distributions for ViTs. In our investigation, we have analyzed the outputs from the norm1 and norm2 layers within the ViT blocks and have also applied InCo Aggregation to these layers. The selection of norm1 and norm2 layers is motivated by the significance of Layer Norm in the architecture of transformers
\citep{xiong2020layer}. Additionally, we have chosen Block7 and Block11 for analysis as, in the context of heterogeneous models, Block7 is the final layer of the smallest ViTs, while Block11 represents the final layer of the largest ViTs.

From Figure~\ref{fig:apd_Analysis_Gradient_cross_environments_layer7} and Figure~\ref{fig:apd_Analysis_Gradient_cross_environments_layer11}, we observe that the cross-environment similarities derived from the shallow layer norm (norm1) are higher compared to those from the deep layer norm (norm2). Moreover, similar to the analysis conducted for ResNets, we discover that the distributions of norm1 in ViTs exhibit greater smoothness compared to norm2, as depicted in Figure~\ref{fig:apd_Analysis_Gradient_Dis_IID_FedAvg_vit_1020} and Figure~\ref{fig:apd_Analysis_Gradient_Dis_Hetero_FedAvg_vit_1020}. These findings reinforce the notion that InCo Aggregation is indeed suitable for ViTs.

\begin{figure}[htbp]
\vskip -0.2in
\centering
\subfloat[Stage2 conv0 in Non-IID with hetero.]
{\includegraphics[width=0.23\columnwidth]{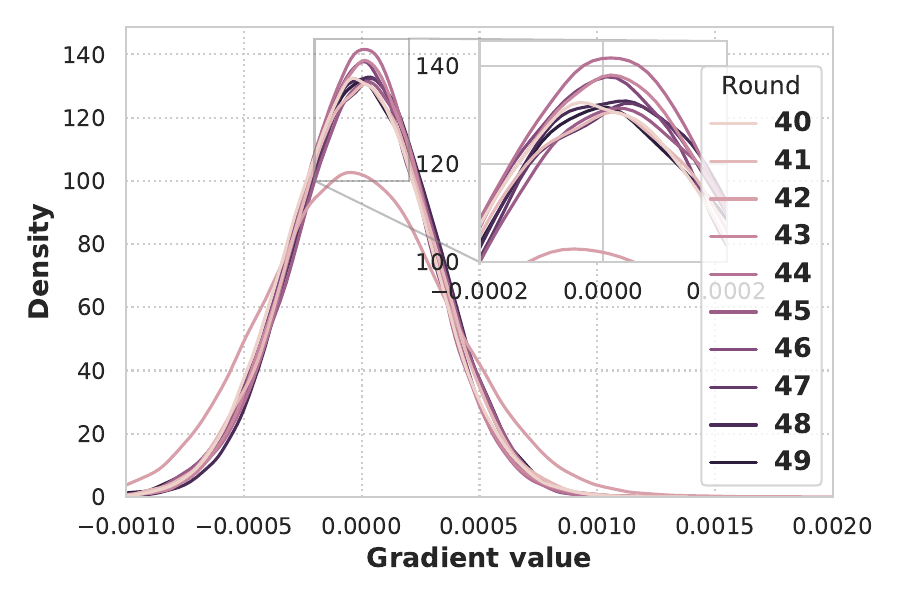}
\label{fig:apd_Analysis_Gradient_Dis_Hetero_FedAvg_stage2_conv0_4050}}
\hfil
\subfloat[Stage2 conv1 in Non-IID with hetero.]
{\includegraphics[width=0.23\columnwidth]{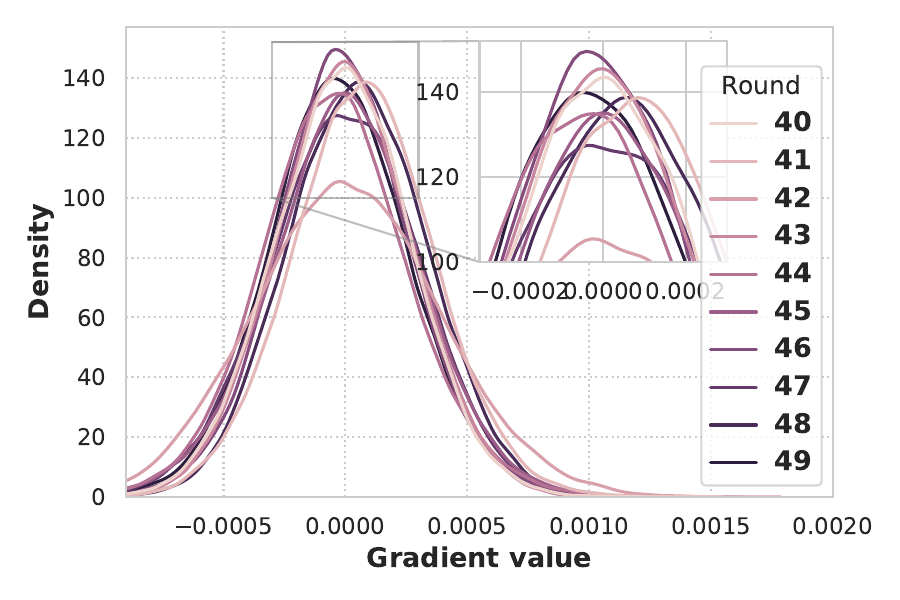}
\label{fig:apd_Analysis_Gradient_Dis_Hetero_FedAvg_stage2_conv1_4050}}
\hfil
\subfloat[Stage3 conv0 in Non-IID with hetero with different seed.]
{\includegraphics[width=0.23\columnwidth]{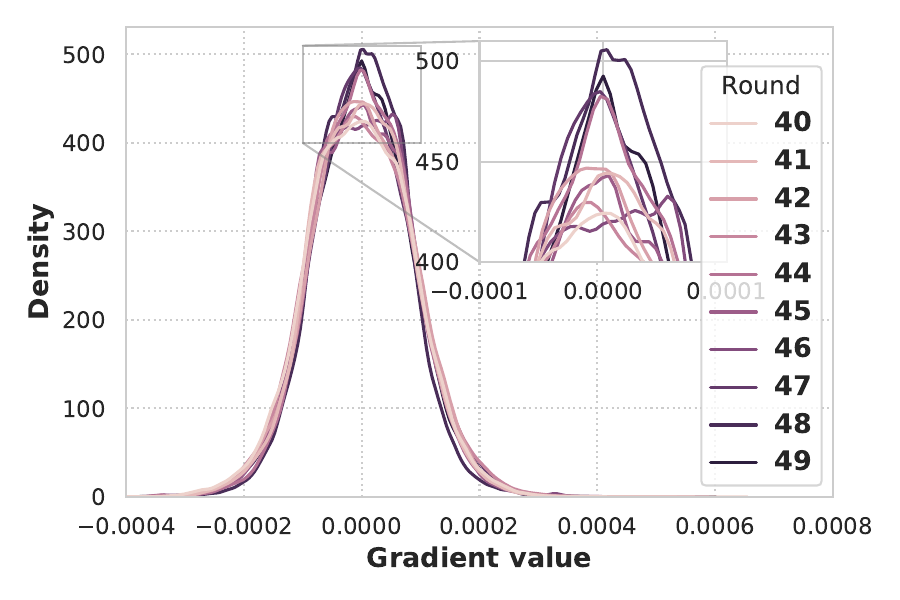}
\label{fig:apd_Analysis_Gradient_Dis_Hetero_FedAvg_stage3_conv0_4050_seed42}}
\hfil
\subfloat[Stage3 conv1 in Non-IID with hetero with different seed.]
{\includegraphics[width=0.23\columnwidth]{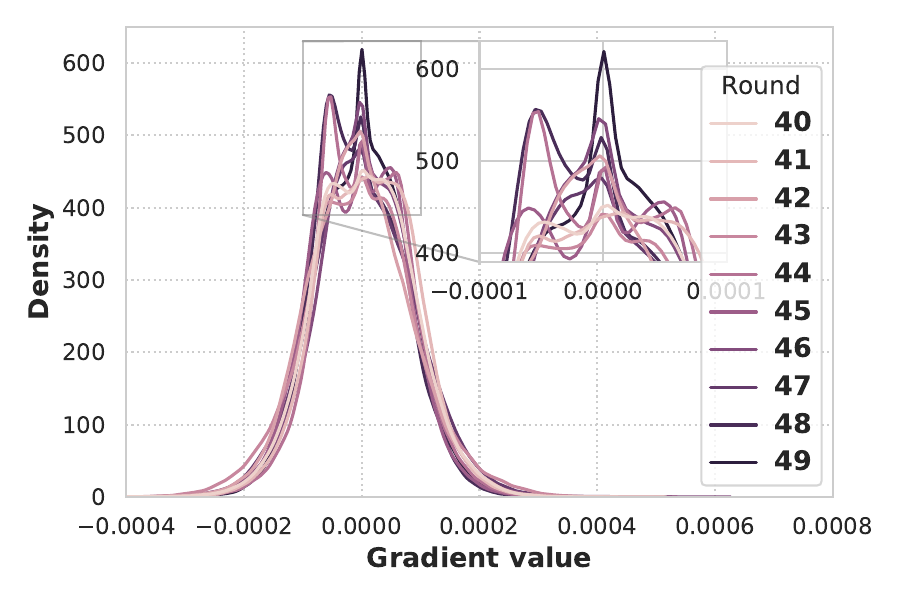}
\label{fig:apd_Analysis_Gradient_Dis_Hetero_FedAvg_stage3_conv1_4050_seed42}}
\vskip -0.07in
\caption{The gradient distributions from round 40 to 50 of ResNets in Non-IID with hetero. (a) and (b) Stage2 conv0 and conv1. (c) and (d) Stage3 conv0 and conv1 with different seed.}
\label{fig:apd_Analysis_Gradient_Dis_Hetero_FedAvg_resnet_stage2_diff_stage3_4050}
\vskip -0.1in
\end{figure}

\subsection{Gradient Distributions from Stage 2 and Different Seed.}
Figure~\ref{fig:apd_Analysis_Gradient_Dis_Hetero_FedAvg_stage2_conv0_4050} and Figure~\ref{fig:apd_Analysis_Gradient_Dis_Hetero_FedAvg_stage2_conv1_4050} demonstrate the gradient distributions in Stage 2. In contrast to the gradient distributions of Stage 3, the differences in gradient distributions across different layers are less evident for Stage 2. This can be observed from Figure~\ref{fig:Analysis_resnet_CKA}, where the CKA similarity for Stage 2 is considerably higher than that of Stage 3. The higher similarity indicates that Stage 2 is relatively less biased and more generalized compared to Stage 3, resulting in less noticeable differences in gradient distributions. This observation further supports the relationship between similarity and smoothness, as higher similarity leads to smoother distributions. Moreover, Figure~\ref{fig:apd_Analysis_Gradient_Dis_Hetero_FedAvg_stage3_conv0_4050_seed42} and Figure~\ref{fig:apd_Analysis_Gradient_Dis_Hetero_FedAvg_stage3_conv1_4050_seed42} illustrate that the gradient distributions still keep the same properties in different random seed, indicating that the relations between similarity and smooth gradients are not affected by SGD noise.

\begin{figure}[htbp]
\vskip -0.2in
\centering
\subfloat[CKA with mean.]
{\includegraphics[width=0.23\columnwidth]{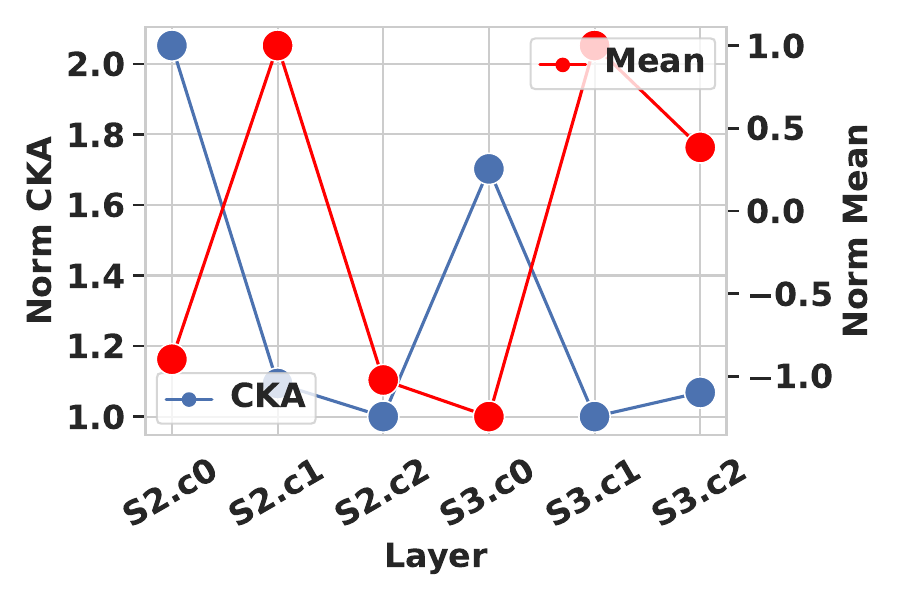}
\label{fig:apd_CKA_mean_relation}}
\hfil
\subfloat[CKA with variance.]
{\includegraphics[width=0.23\columnwidth]{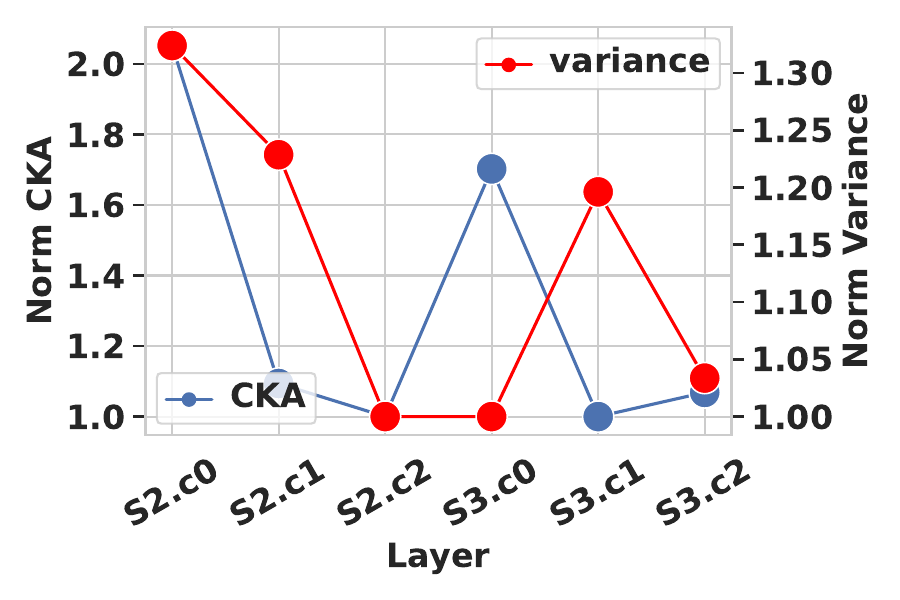}
\label{fig:apd_CKA_var_relation}}
\hfil
\subfloat[CKA with covariance.]
{\includegraphics[width=0.23\columnwidth]{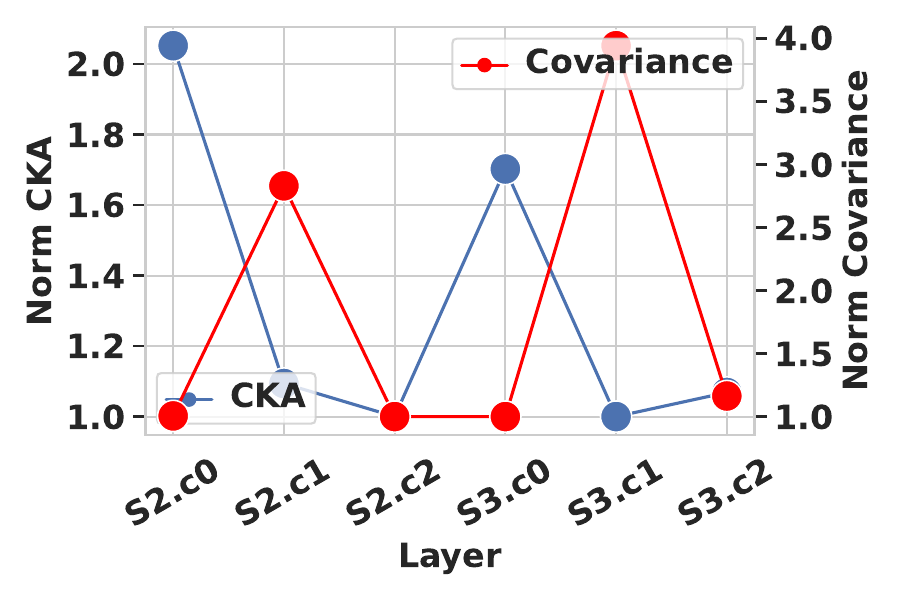}
\label{fig:apd_CKA_cov_relation}}
\hfil
\subfloat[CKA with difference.]
{\includegraphics[width=0.23\columnwidth]{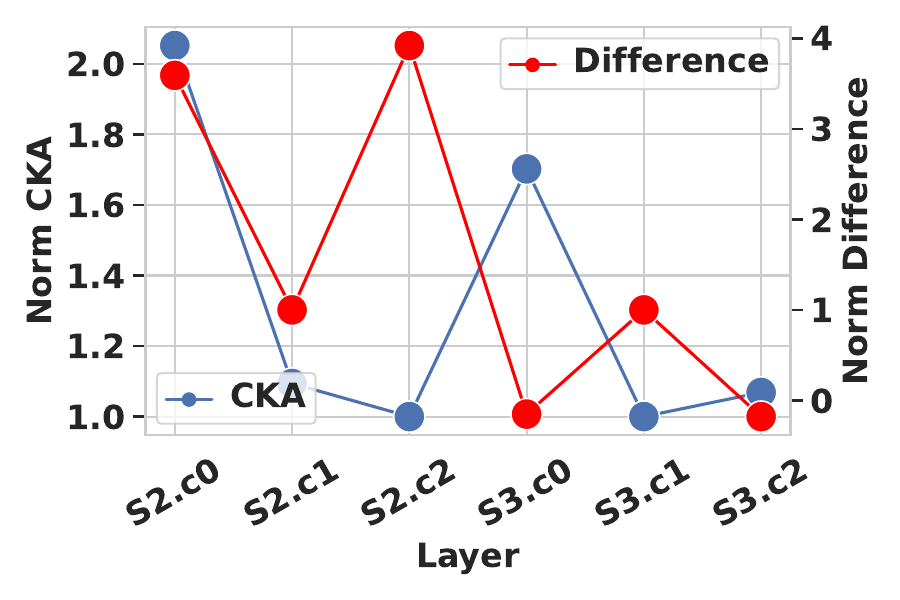}
\label{fig:apd_CKA_diff_relation}}
\vskip -0.07in
\caption{Other relationships between CKA and the cross-environment statistics of gradients from IID with homo and Non-IID with hetero. We use abbreviations for "stage" and "conv," represented as "s" and "c" respectively. For example, "s2c0" represents stage 2, conv0.}
\label{fig:apd_CKA_other_relations}
\vskip -0.1in
\end{figure}

\subsection{Other Relations between CKA and the Statistics of Gradients.}
Figure~\ref{fig:apd_CKA_other_relations} provides an overview of additional relationships between layer similarity and the cross-environment gradient statistics derived from IID with homo and Non-IID with hetero.  We calculate the difference between gradients from the same layer across these two environments. To clarify the tendency of similarity for each stage, we normalize the results according to the smallest value within each stage. As shown in Figure~\ref{fig:apd_CKA_other_relations}, none of these gradient statistics exhibit stronger correlations with the similarity of gradients compared to the smoothness, discussed in Section~\ref{sec:deep_insights_of_grads} and Appendix~\ref{subsec:apd_gradients_analysis_ViTs}.

\subsection{Heatmaps for the Case Study}
\label{subsec:apd_heatmaps_case_study}
\begin{figure*}[!t]
% \vskip 0.2in
\centering
\subfloat[The CKA similarity of IID with homo for stage 0.]
{\includegraphics[width=0.22\textwidth]{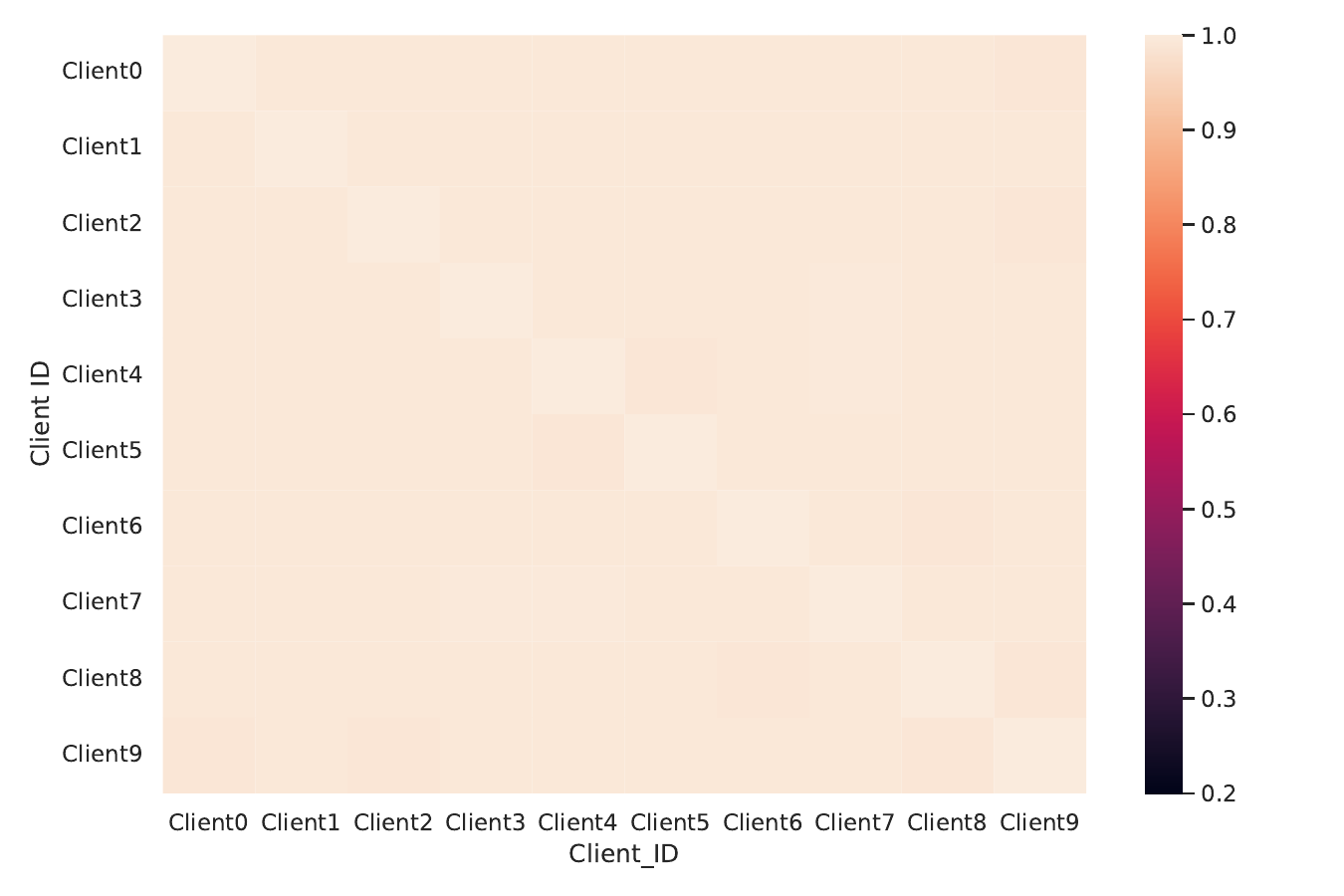}
\label{fig:apd_iid_homo_0}}
\hfil
\subfloat[The CKA similarity of IID with homo for stage 1.]
{\includegraphics[width=0.22\textwidth]{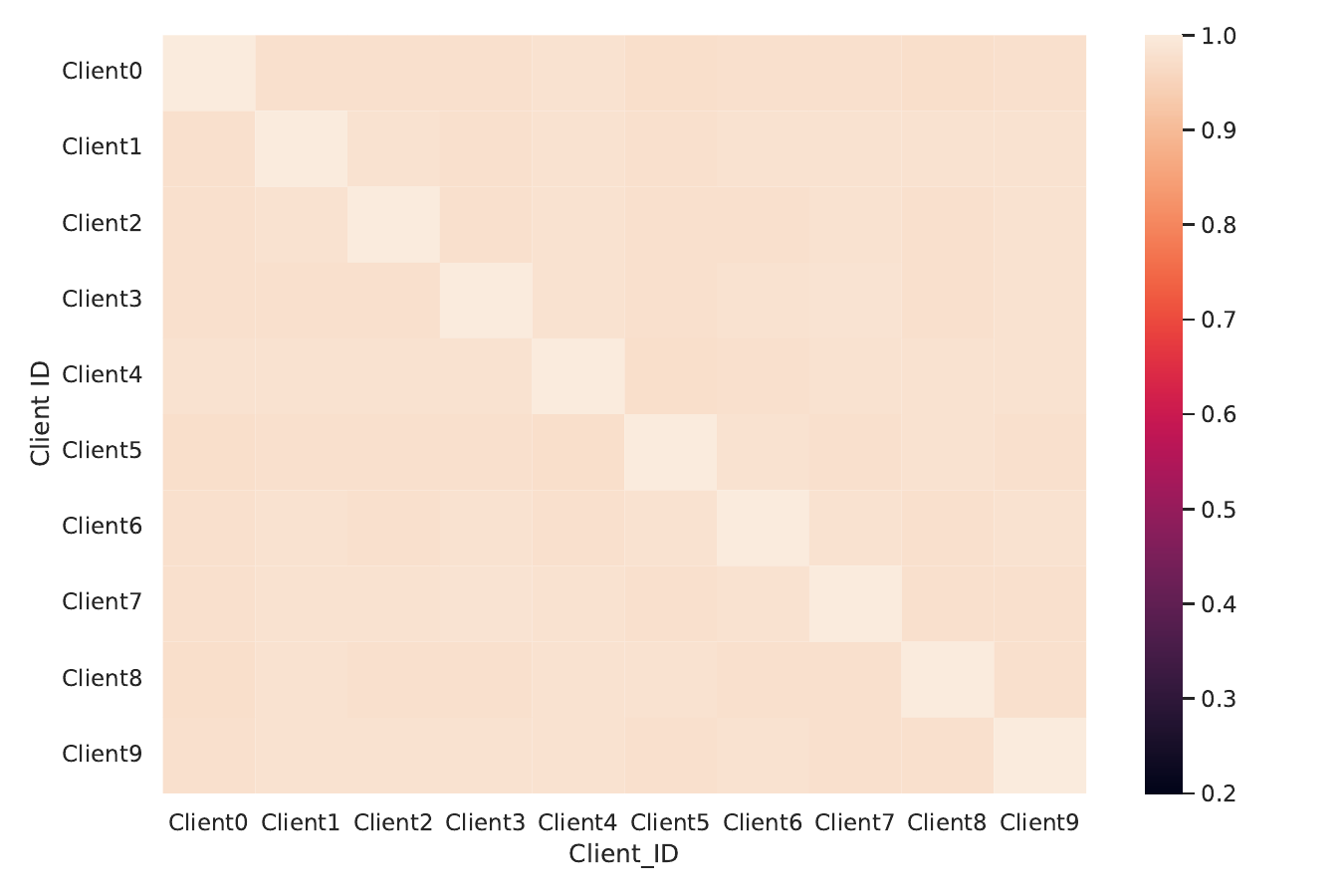}
\label{fig:apd_iid_homo_1}}
\hfil
\subfloat[The CKA similarity of IID with homo for stage 2.]
{\includegraphics[width=0.22\textwidth]{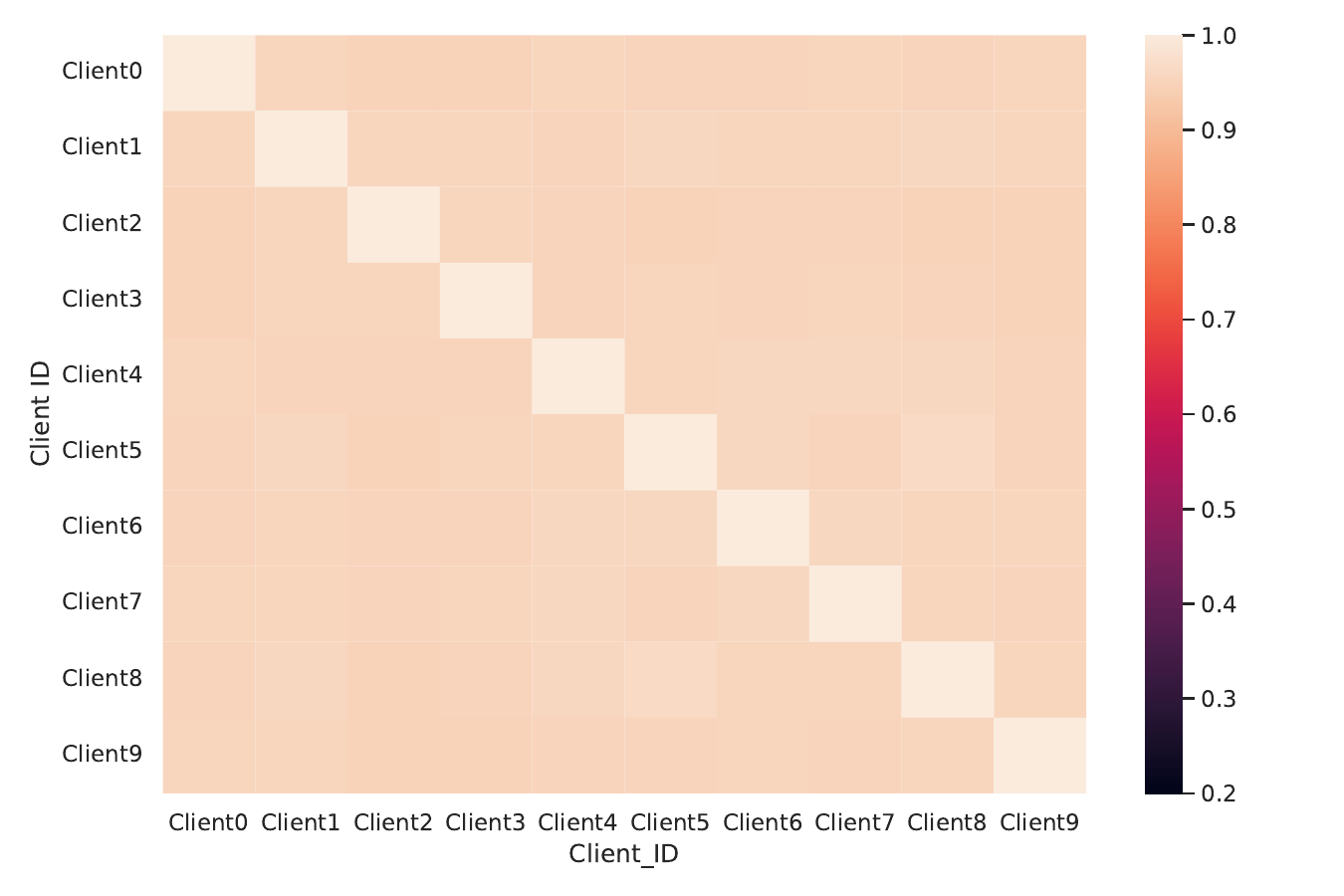}
\label{fig:apd_iid_homo_2}}
\hfil
\subfloat[The CKA similarity of IID with homo for stage 3.]
{\includegraphics[width=0.22\textwidth]{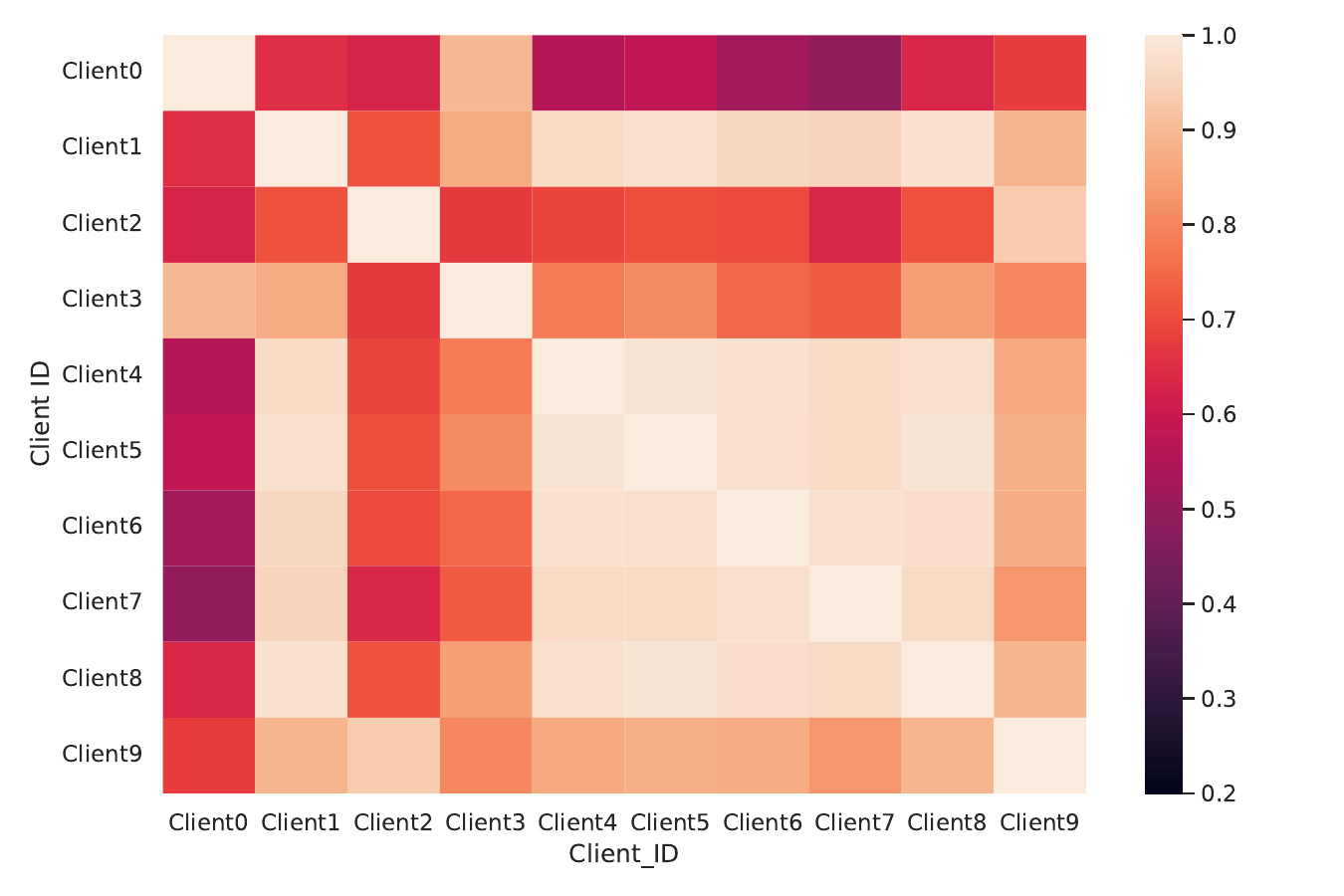}
\label{fig:apd_iid_homo_3}}

\subfloat[The CKA similarity of Non-IID with homo for stage 0.]
{\includegraphics[width=0.22\textwidth]{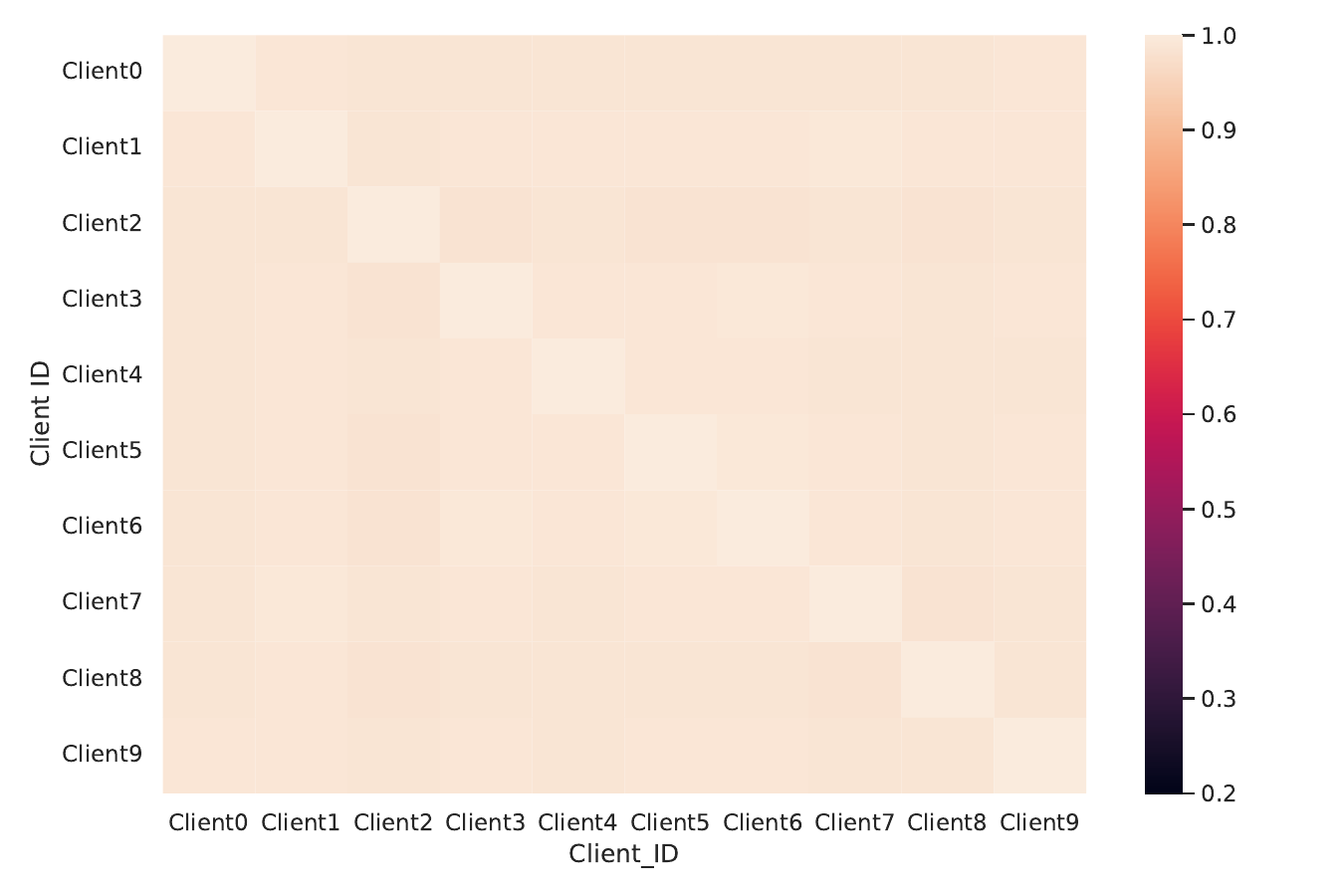}
\label{fig:apd_noniid_homo_0}}
\hfil
\subfloat[The CKA similarity of Non-IID with homo for stage 1.]
{\includegraphics[width=0.22\textwidth]{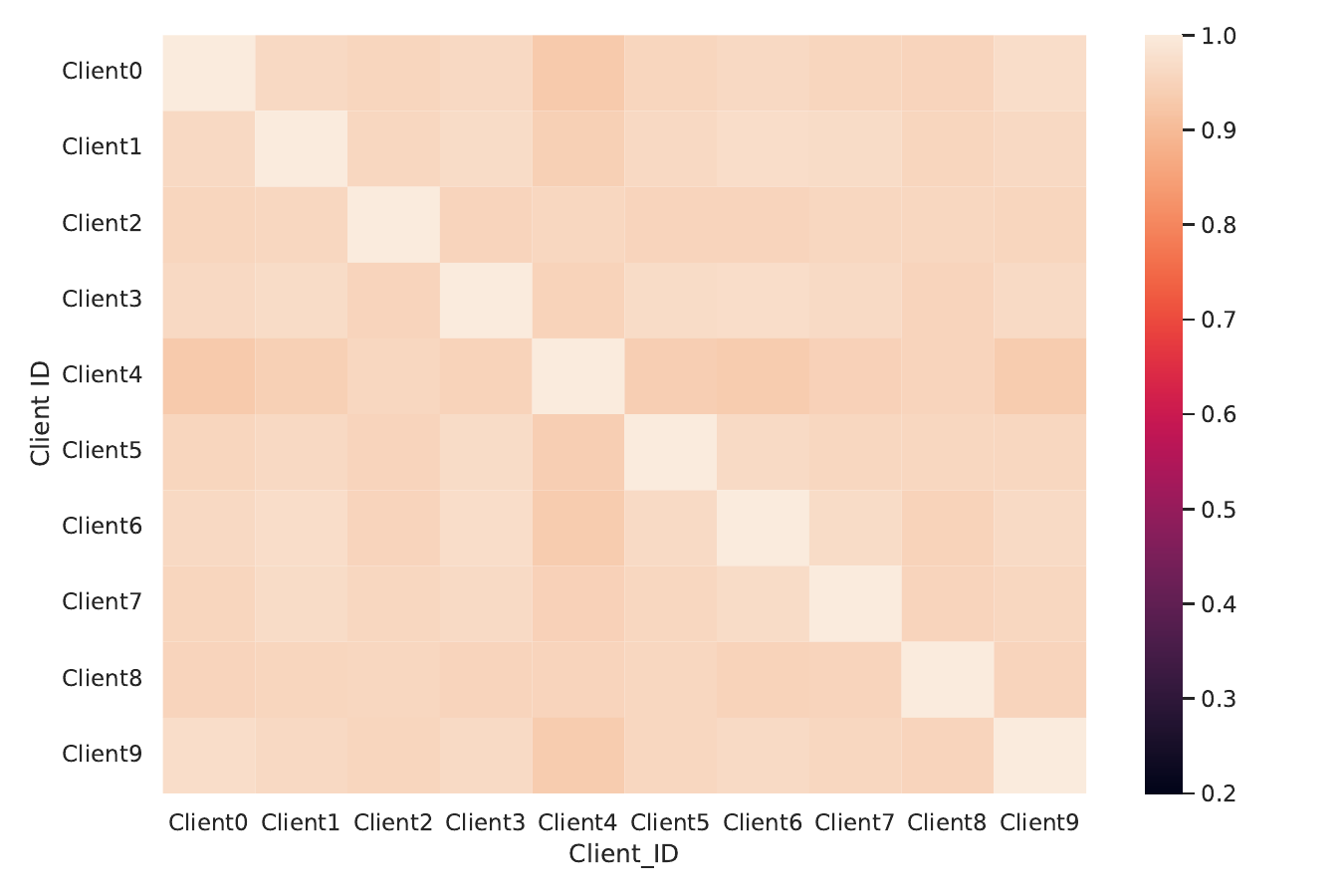}
\label{fig:apd_noniid_homo_1}}
\hfil
\subfloat[The CKA similarity of Non-IID with homo for stage 2.]
{\includegraphics[width=0.22\textwidth]{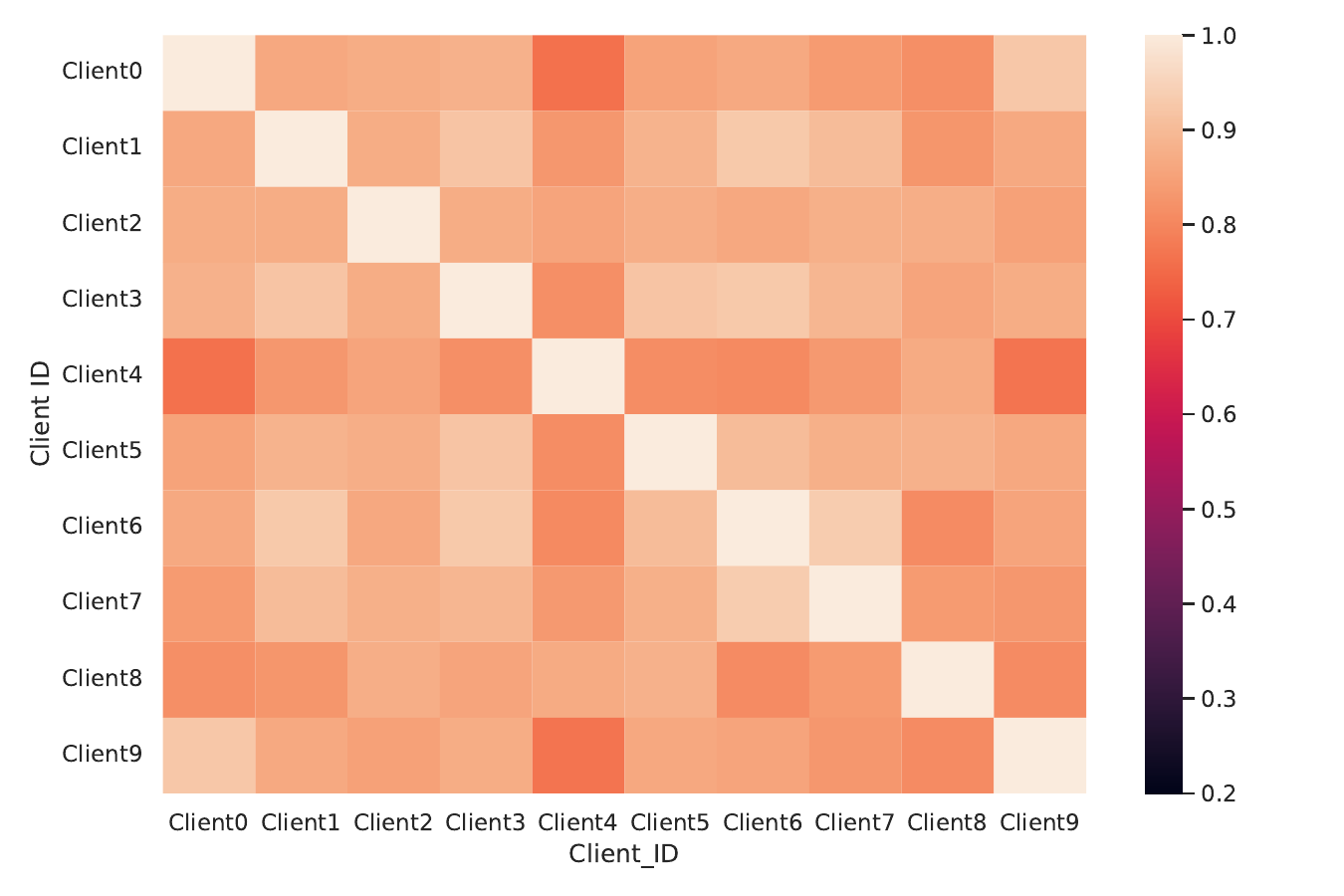}
\label{fig:apd_noniid_homo_2}}
\hfil
\subfloat[The CKA similarity of Non-IID with homo for stage 3.]
{\includegraphics[width=0.22\textwidth]{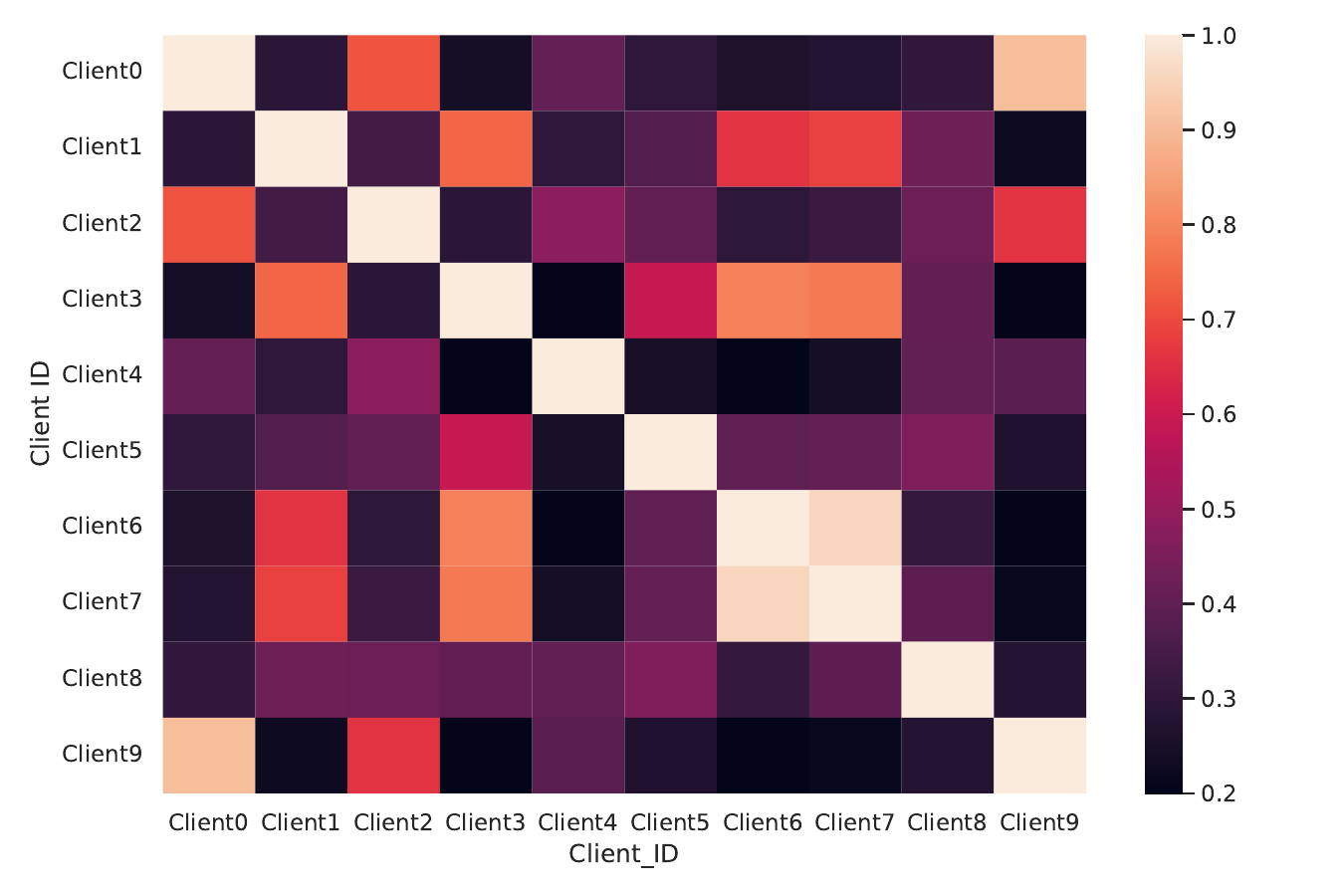}
\label{fig:apd_noniid_homo_3}}

\subfloat[The CKA similarity of Non-IID with hetero for stage 0.]
{\includegraphics[width=0.22\textwidth]{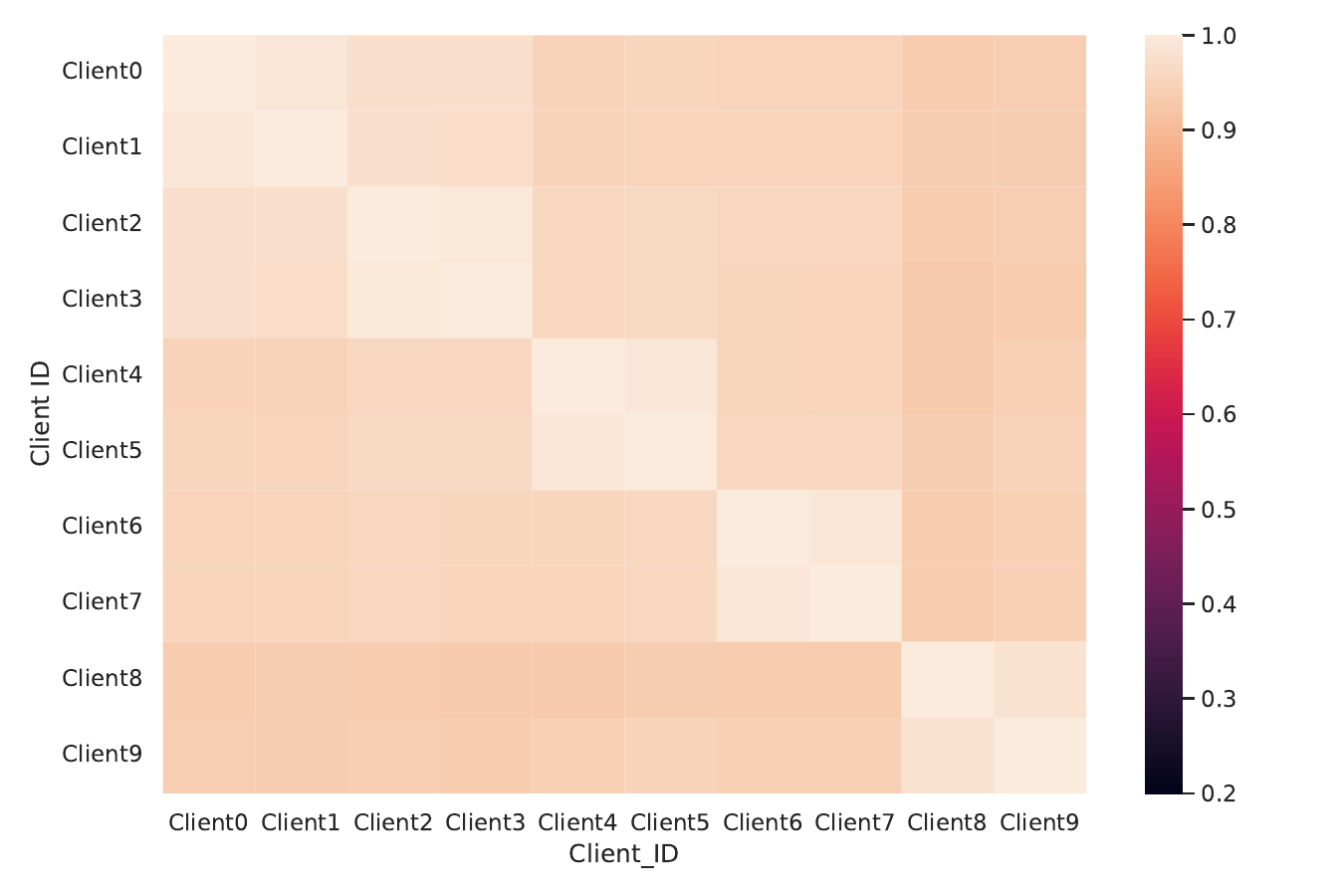}
\label{fig:apd_noniid_hetero_0}}
\hfil
\subfloat[The CKA similarity of Non-IID with hetero for stage 1.]
{\includegraphics[width=0.22\textwidth]{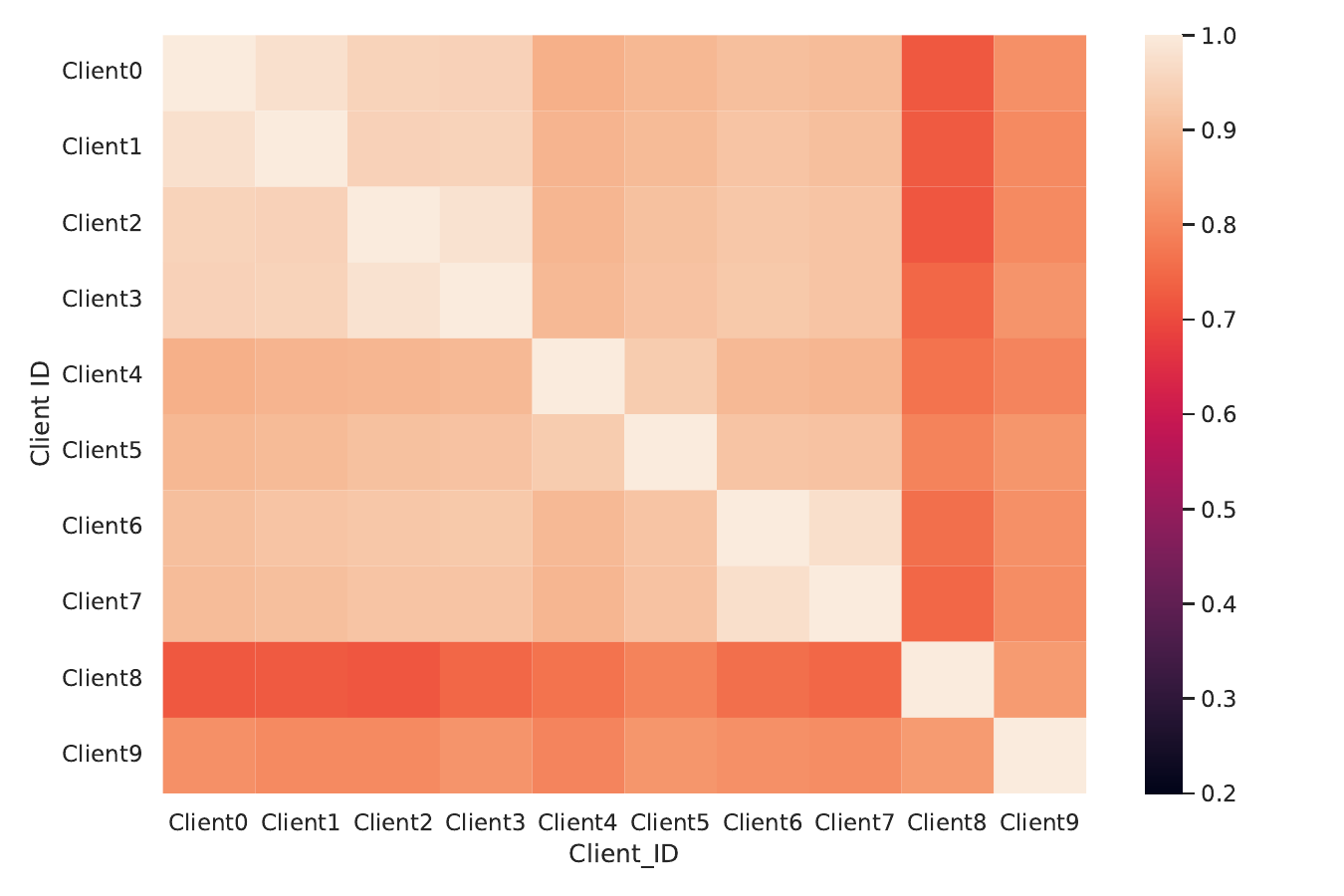}
\label{fig:apd_noniid_hetero_1}}
\hfil
\subfloat[The CKA similarity of Non-IID with hetero for stage 2.]
{\includegraphics[width=0.22\textwidth]{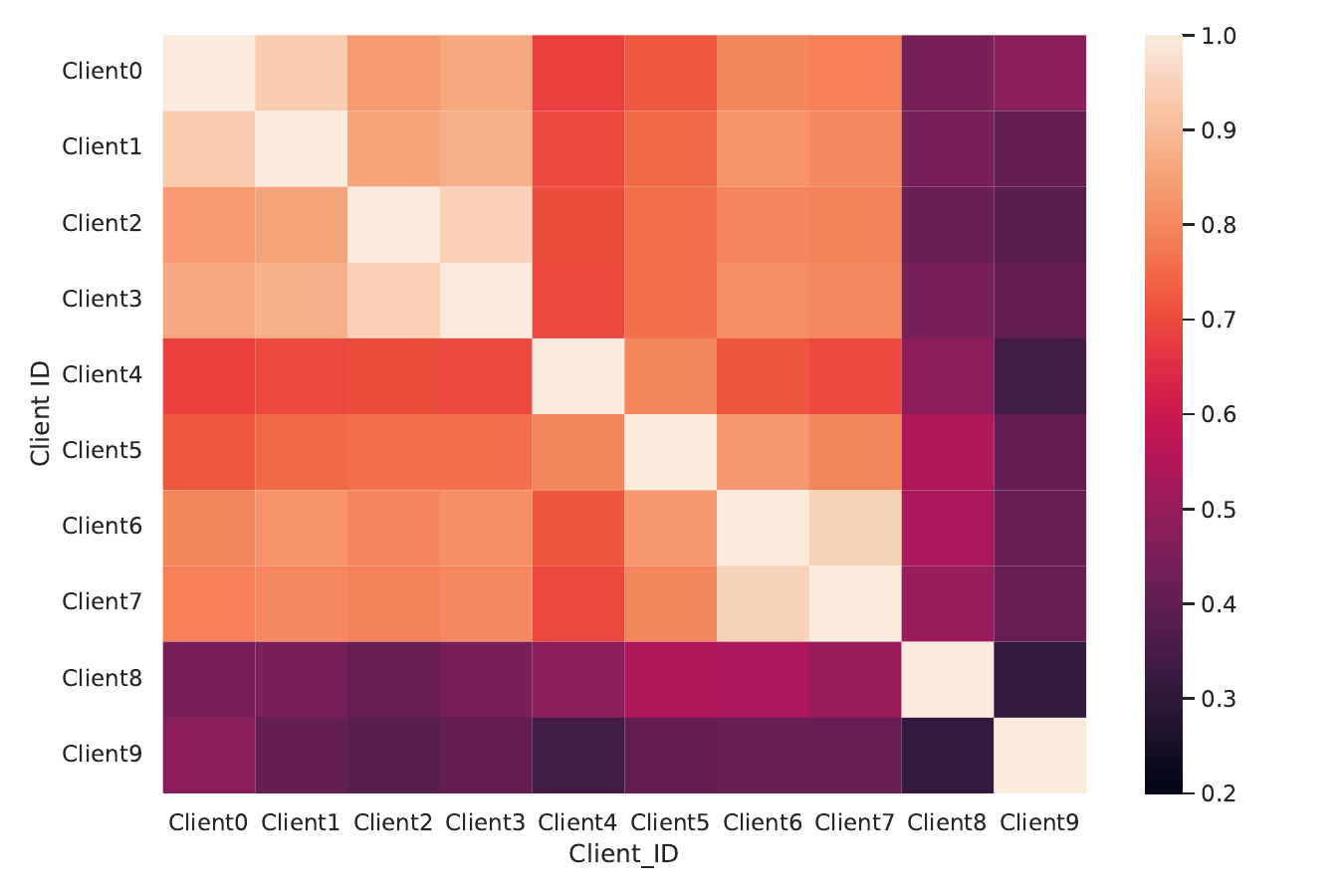}
\label{fig:apd_noniid_hetero_2}}
\hfil
\subfloat[The CKA similarity of Non-IID with hetero for stage 3.]
{\includegraphics[width=0.22\textwidth]{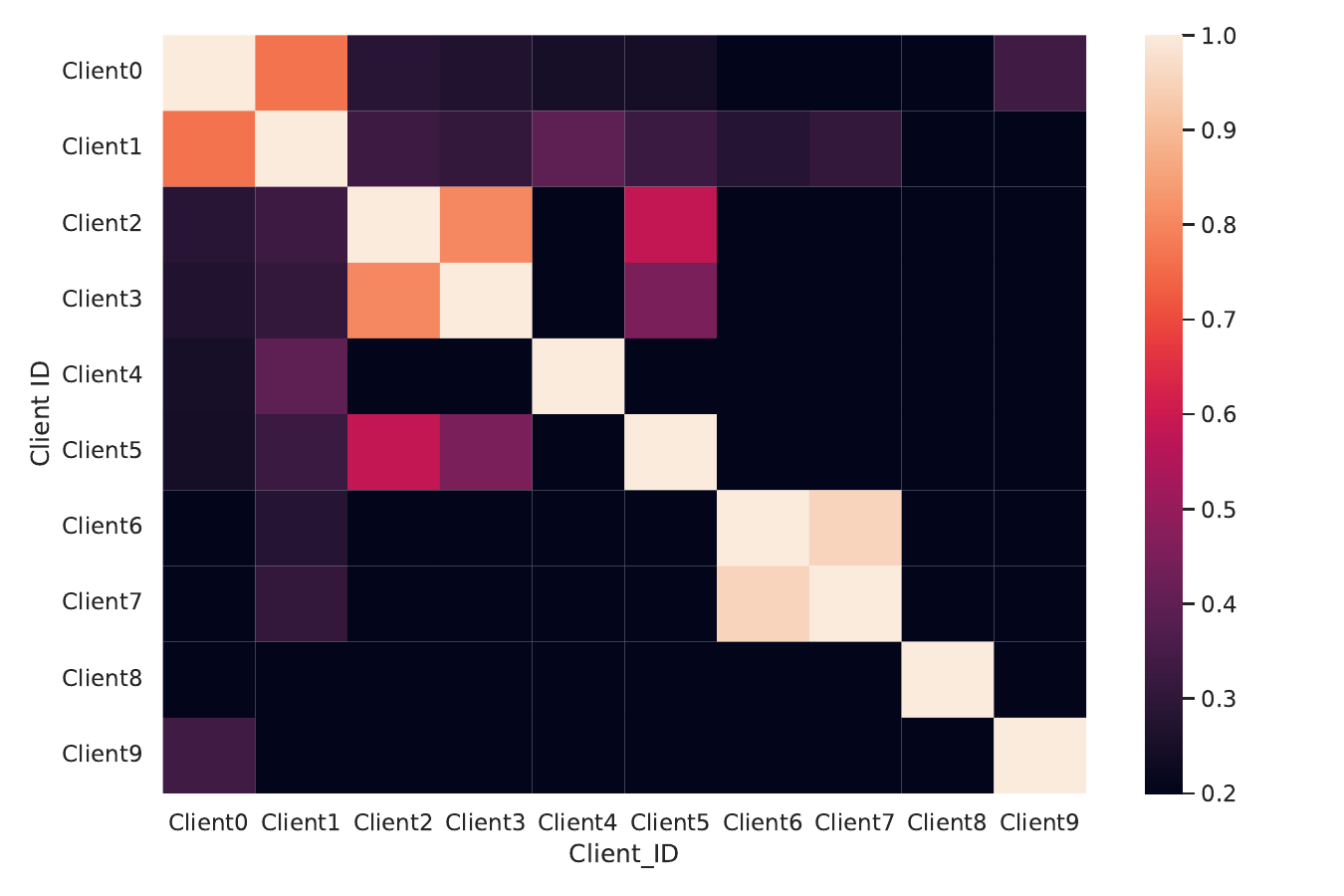}
\label{fig:apd_noniid_hetero_3}}
\caption{The CKA similarity of IID with homo, Non-IID with homo and Non-IID with hetero for ResNets.}
\label{fig:apd_heatmaps_resnets}
\vskip -0.2in
\end{figure*}

\begin{figure*}[!t]
% \vskip 0.2in
\centering
\subfloat[The CKA similarity of IID with homo for layer 4.]
{\includegraphics[width=0.22\textwidth]{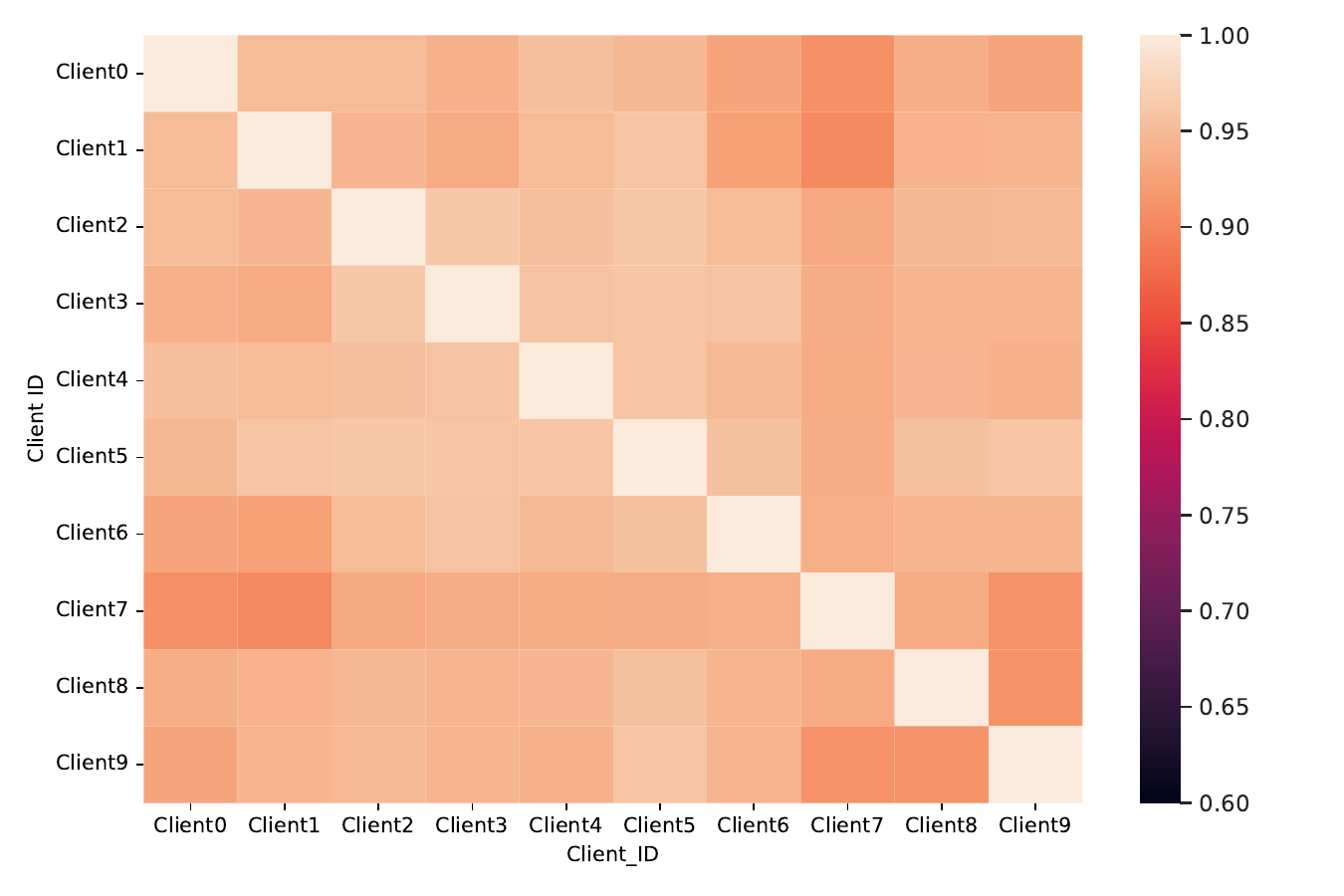}
\label{fig:apd_iid_homo_L4_vit}}
\hfil
\subfloat[The CKA similarity of IID with homo for layer 5.]
{\includegraphics[width=0.22\textwidth]{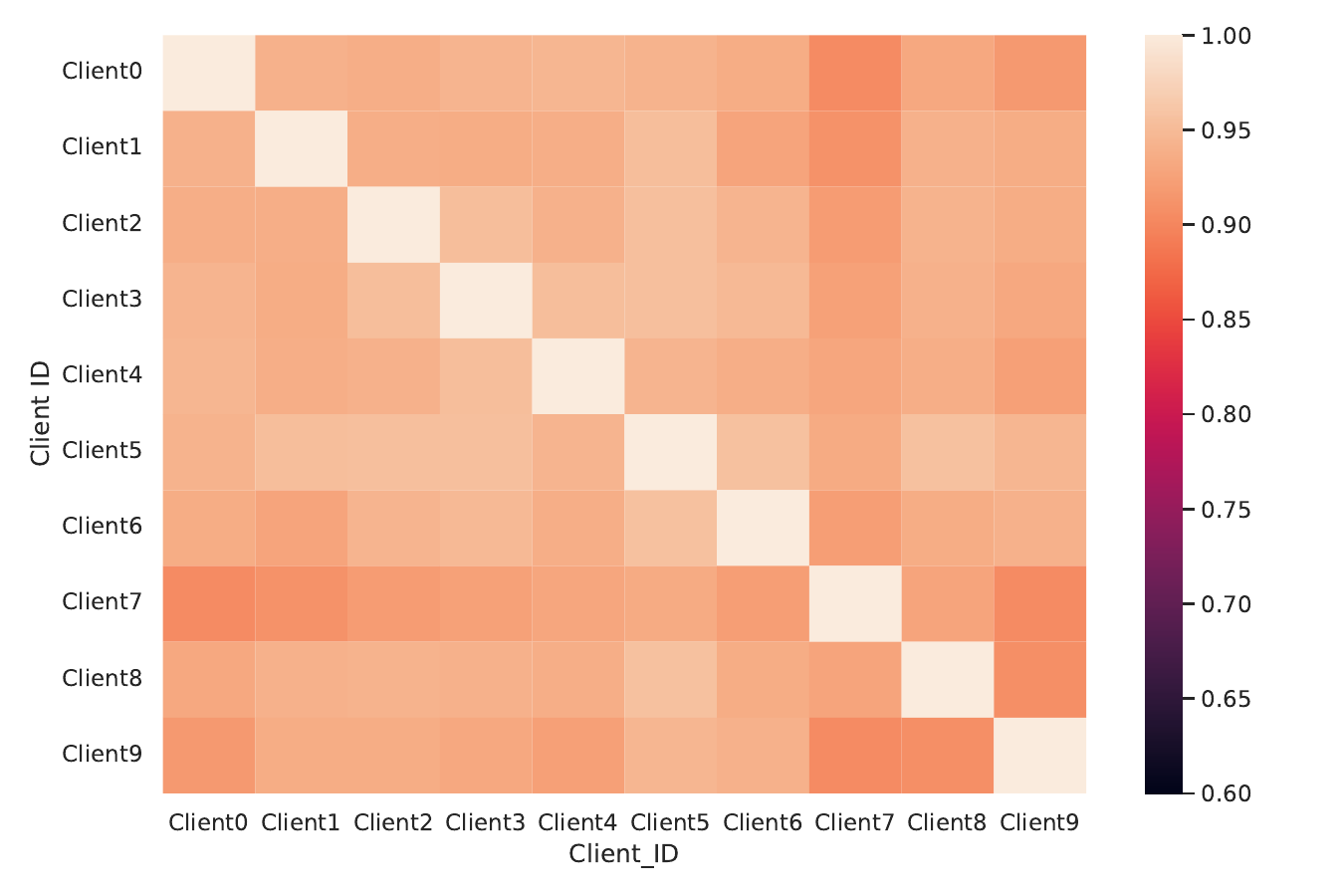}
\label{fig:apd_iid_homo_L5_vit}}
\hfil
\subfloat[The CKA similarity of IID with homo for layer 6.]
{\includegraphics[width=0.22\textwidth]{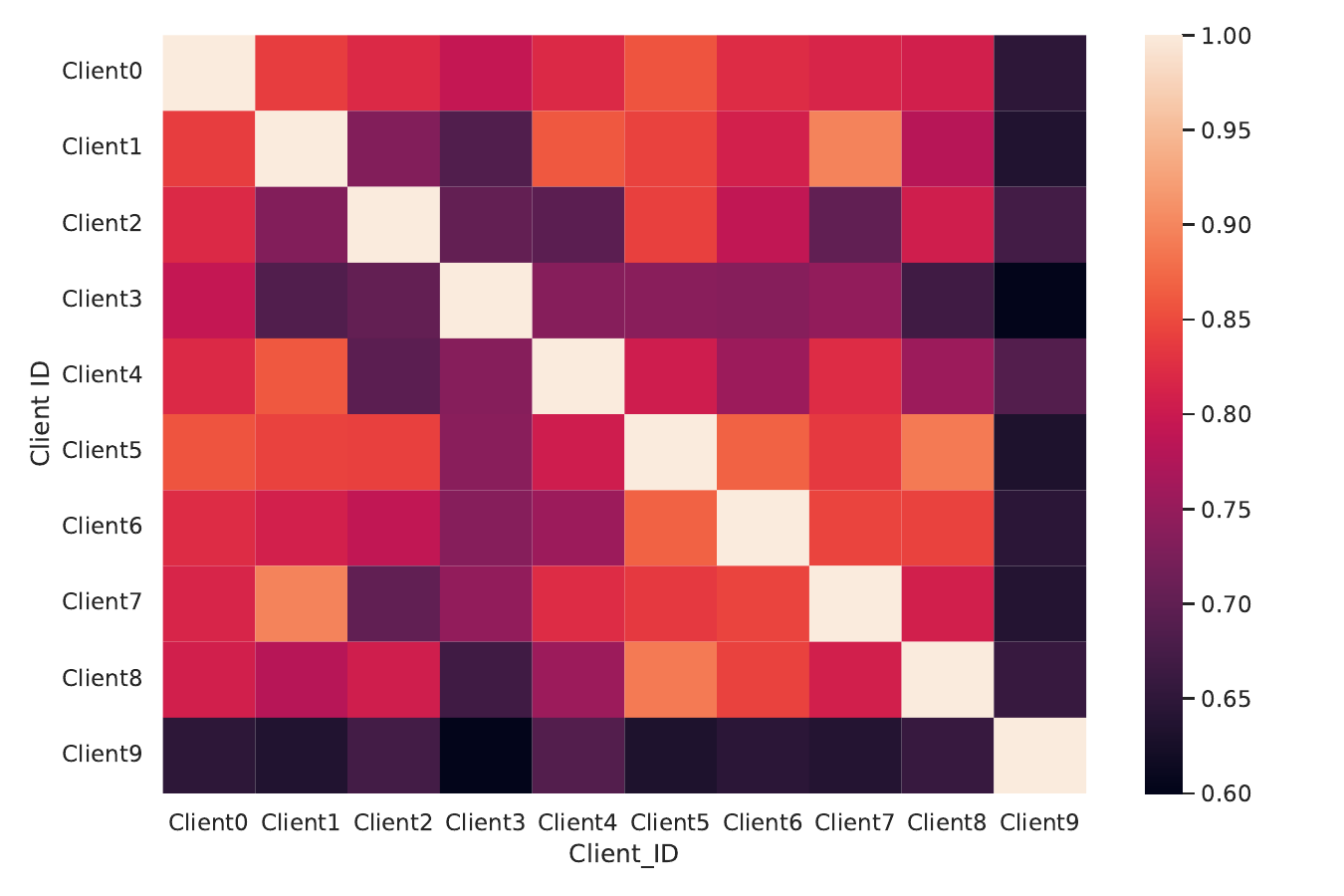}
\label{fig:apd_iid_homo_L6_vit}}
\hfil
\subfloat[The CKA similarity of IID with homo for layer 7.]
{\includegraphics[width=0.22\textwidth]{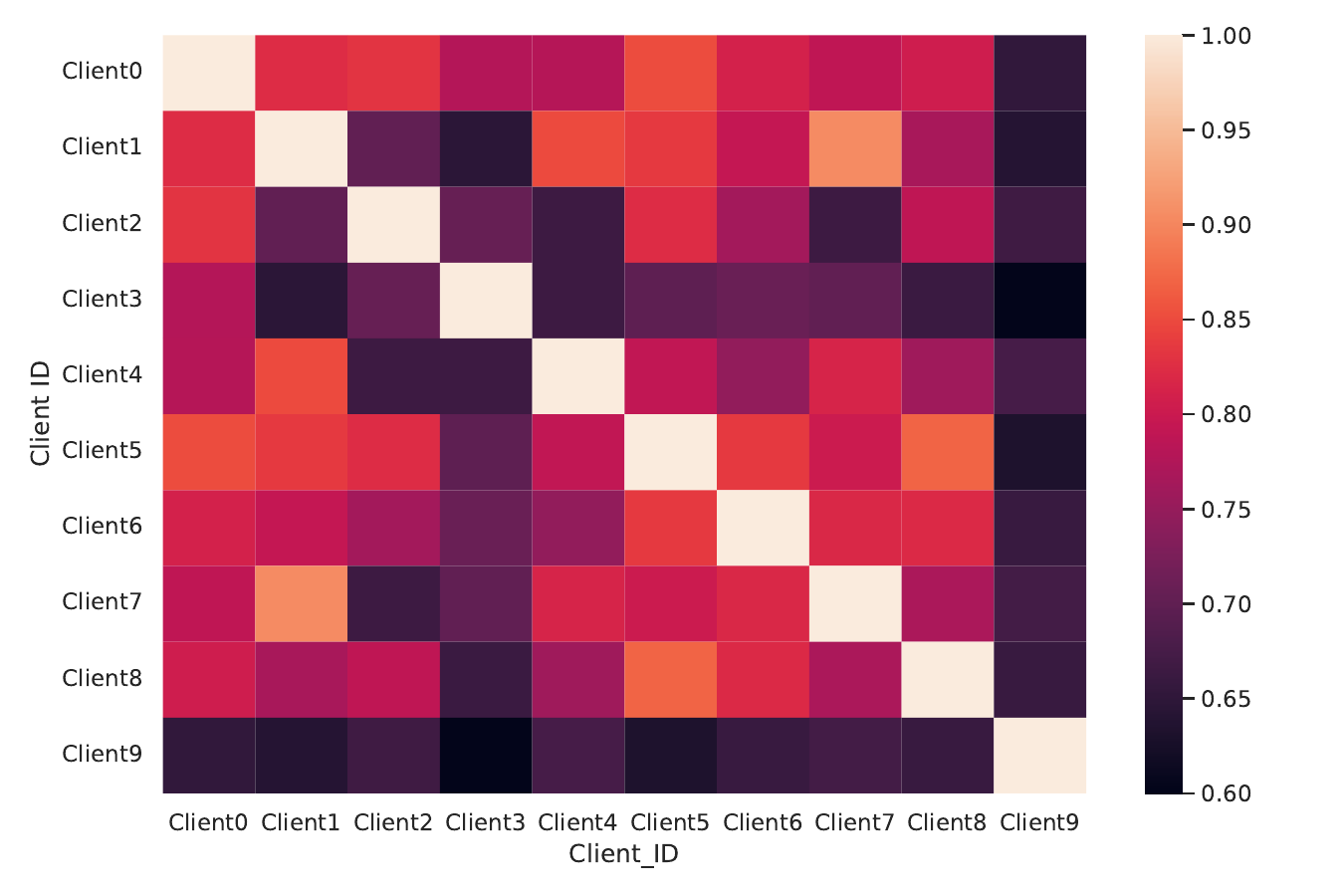}
\label{fig:apd_iid_homo_L7_vit}}

\subfloat[The CKA similarity of Non-IID with homo for layer 4.]
{\includegraphics[width=0.22\textwidth]{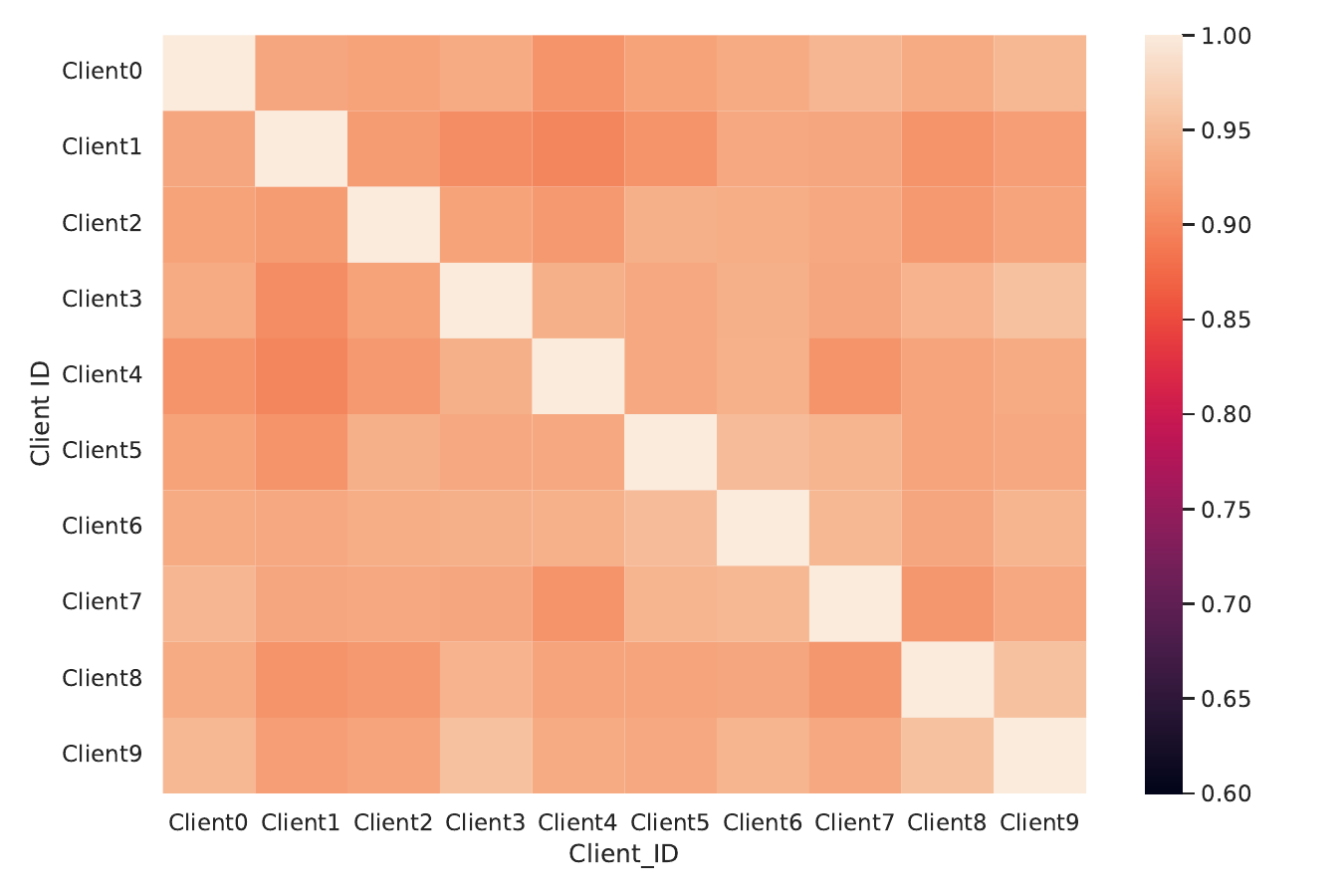}
\label{fig:apd_noniid_homo_L4_vit}}
\hfil
\subfloat[The CKA similarity of Non-IID with homo for layer 5.]
{\includegraphics[width=0.22\textwidth]{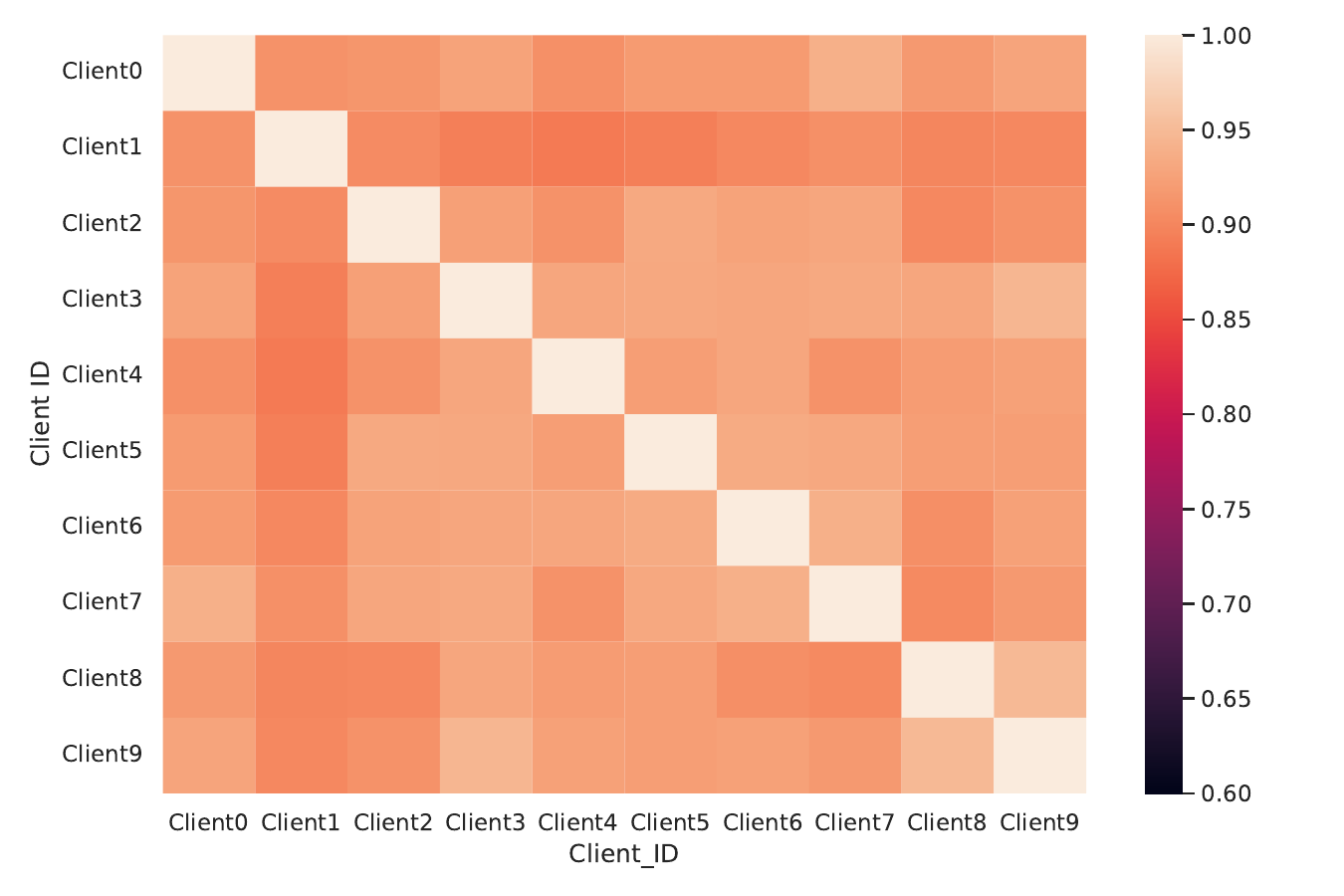}
\label{fig:apd_noniid_homo_L5_vit}}
\hfil
\subfloat[The CKA similarity of Non-IID with homo for layer 6.]
{\includegraphics[width=0.22\textwidth]{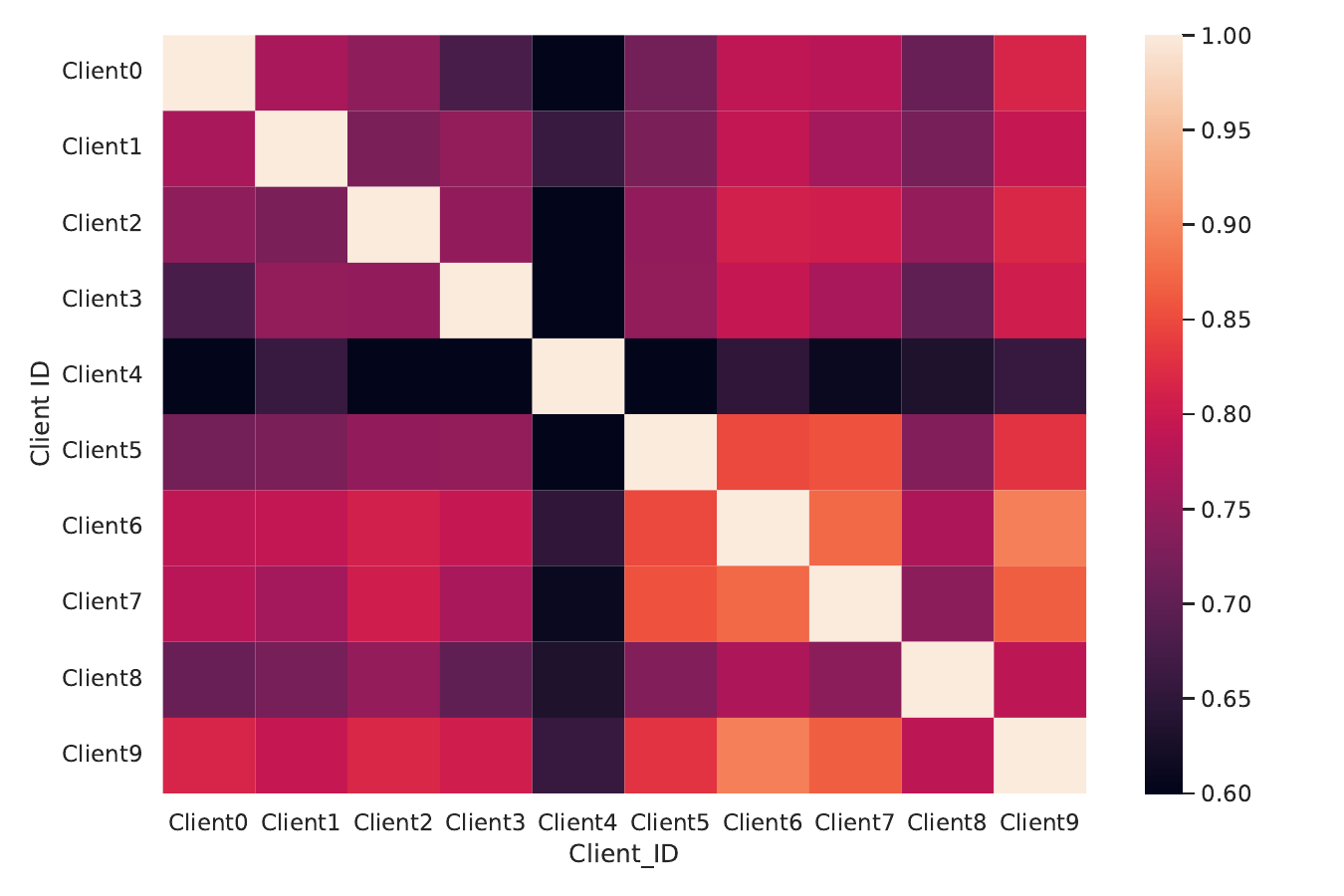}
\label{fig:apd_noniid_homo_L6_vit}}
\hfil
\subfloat[The CKA similarity of Non-IID with homo for layer 7.]
{\includegraphics[width=0.22\textwidth]{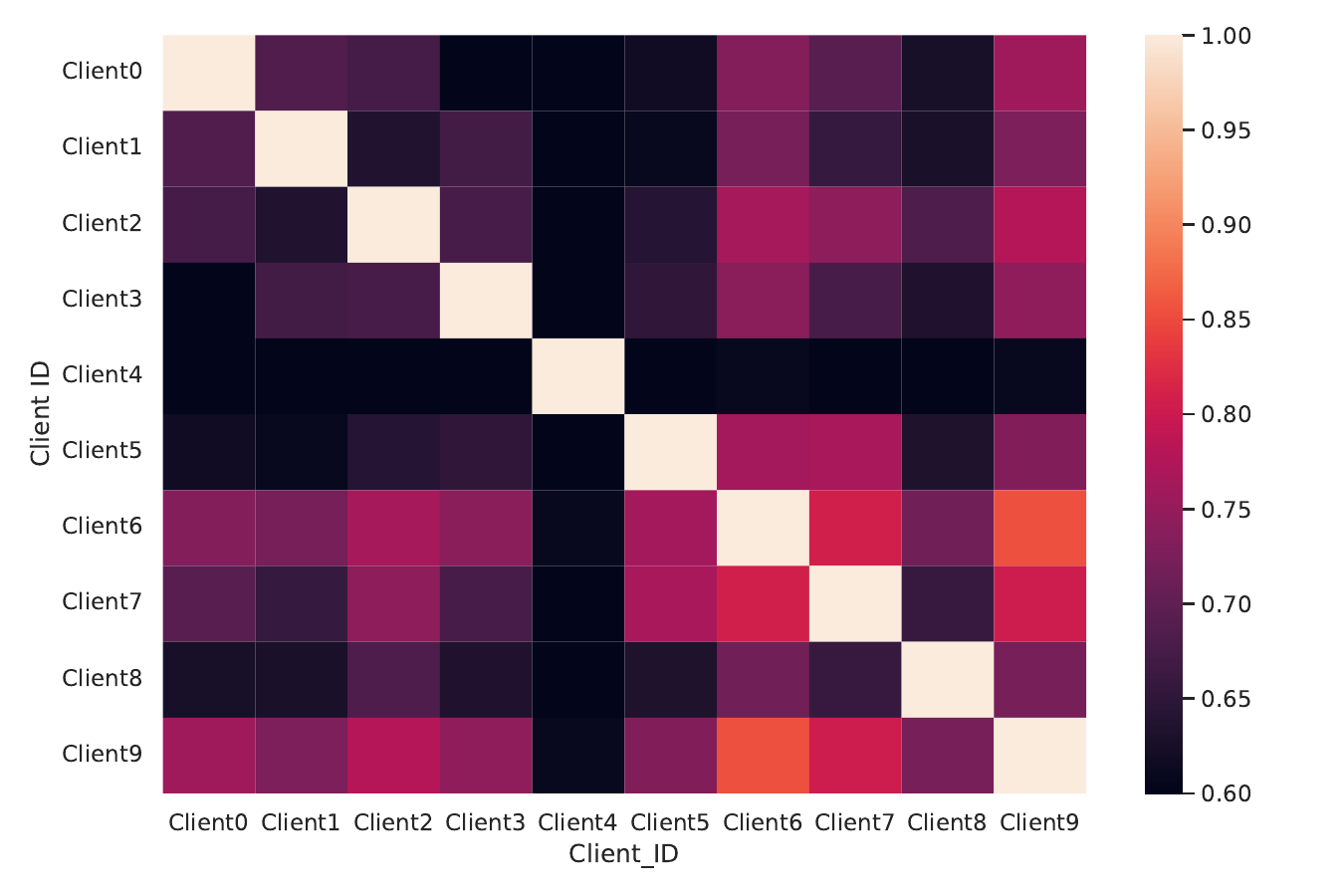}
\label{fig:apd_noniid_homo_L7_vit}}

\subfloat[The CKA similarity of Non-IID with hetero for layer 4.]
{\includegraphics[width=0.22\textwidth]{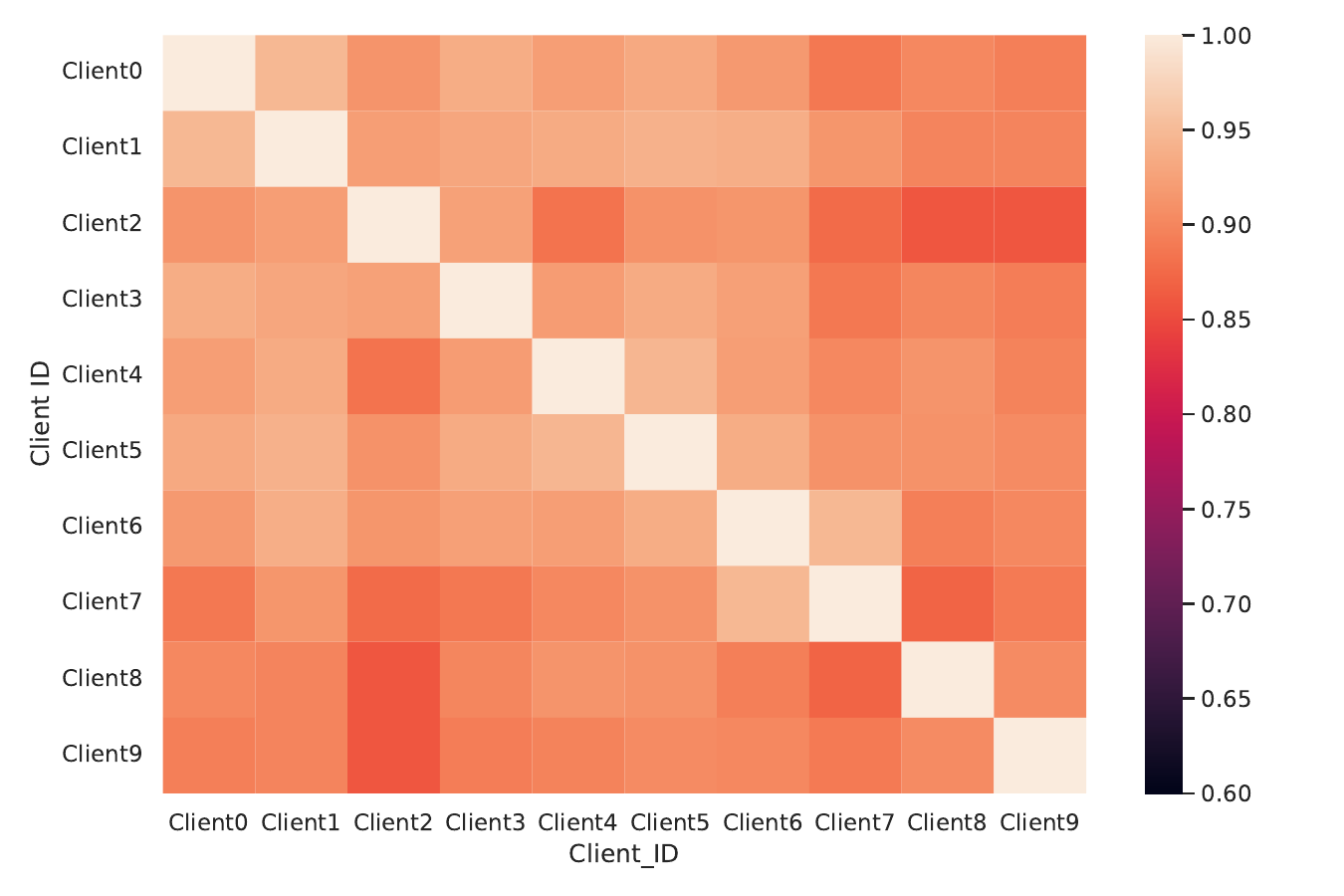}
\label{fig:apd_noniid_hetero_L4_vit}}
\hfil
\subfloat[The CKA similarity of Non-IID with hetero for layer 5.]
{\includegraphics[width=0.22\textwidth]{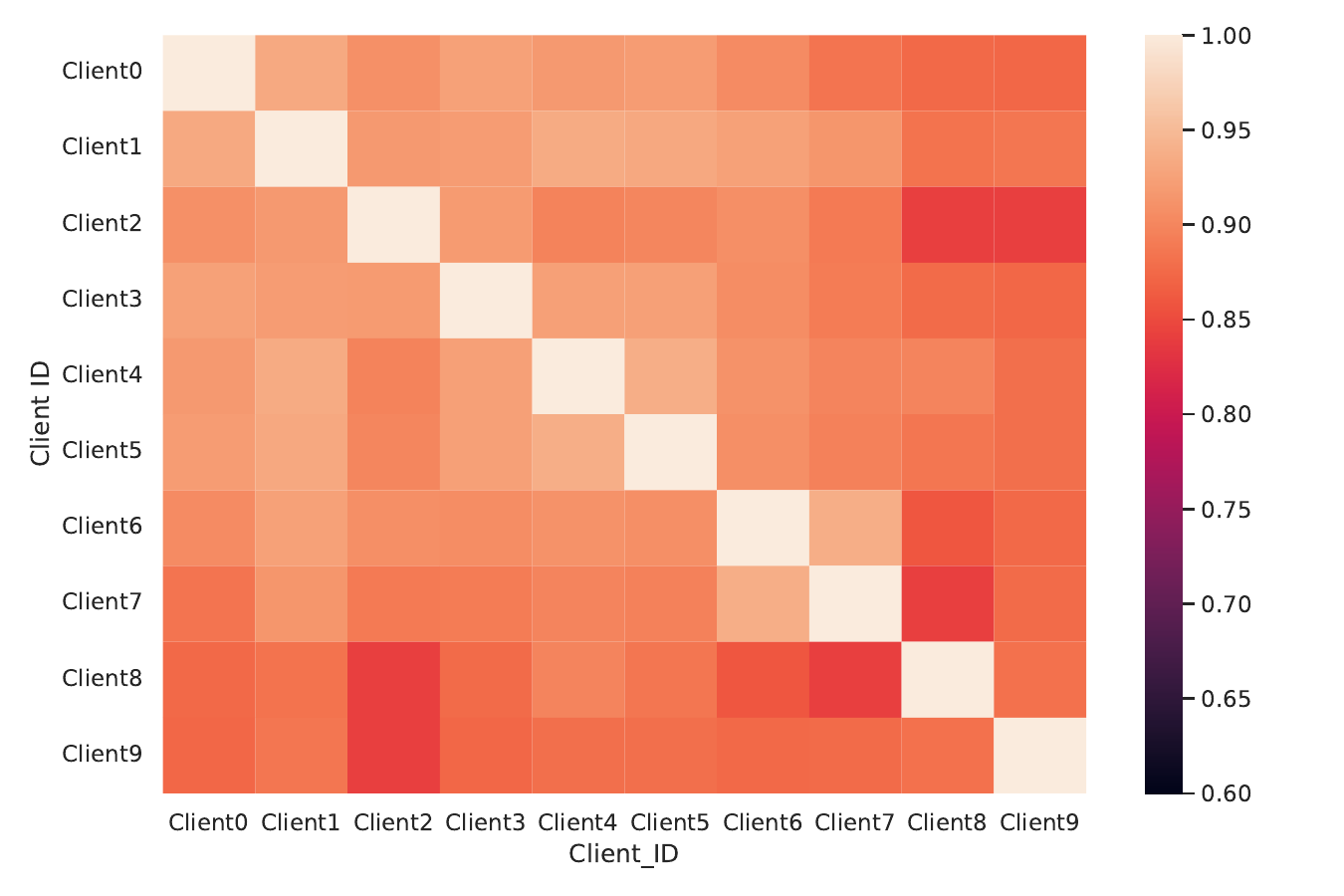}
\label{fig:apd_noniid_hetero_L5_vit}}
\hfil
\subfloat[The CKA similarity of Non-IID with hetero for layer 6.]
{\includegraphics[width=0.22\textwidth]{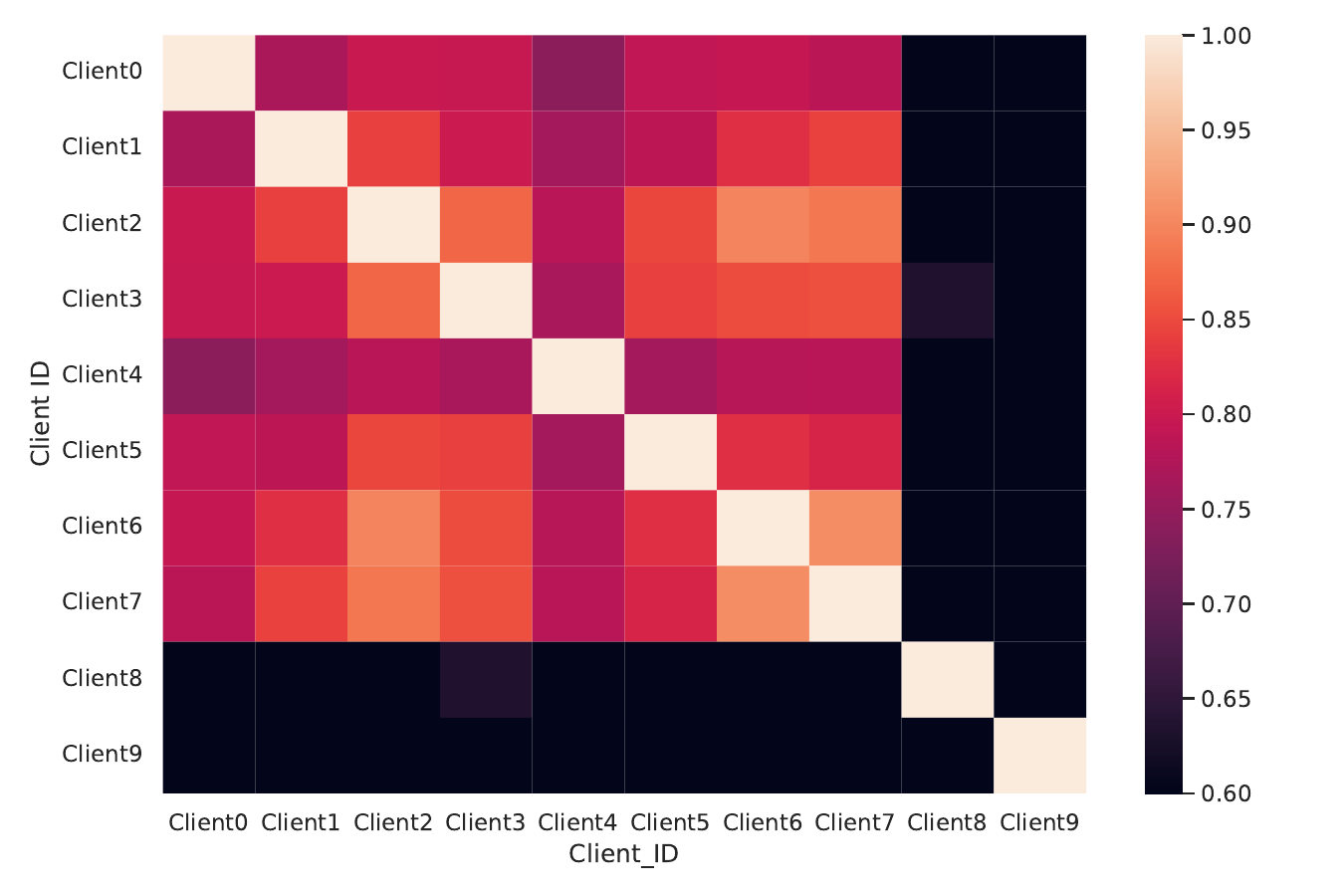}
\label{fig:apd_noniid_hetero_L6_vit}}
\hfil
\subfloat[The CKA similarity of Non-IID with hetero for layer 7.]
{\includegraphics[width=0.22\textwidth]{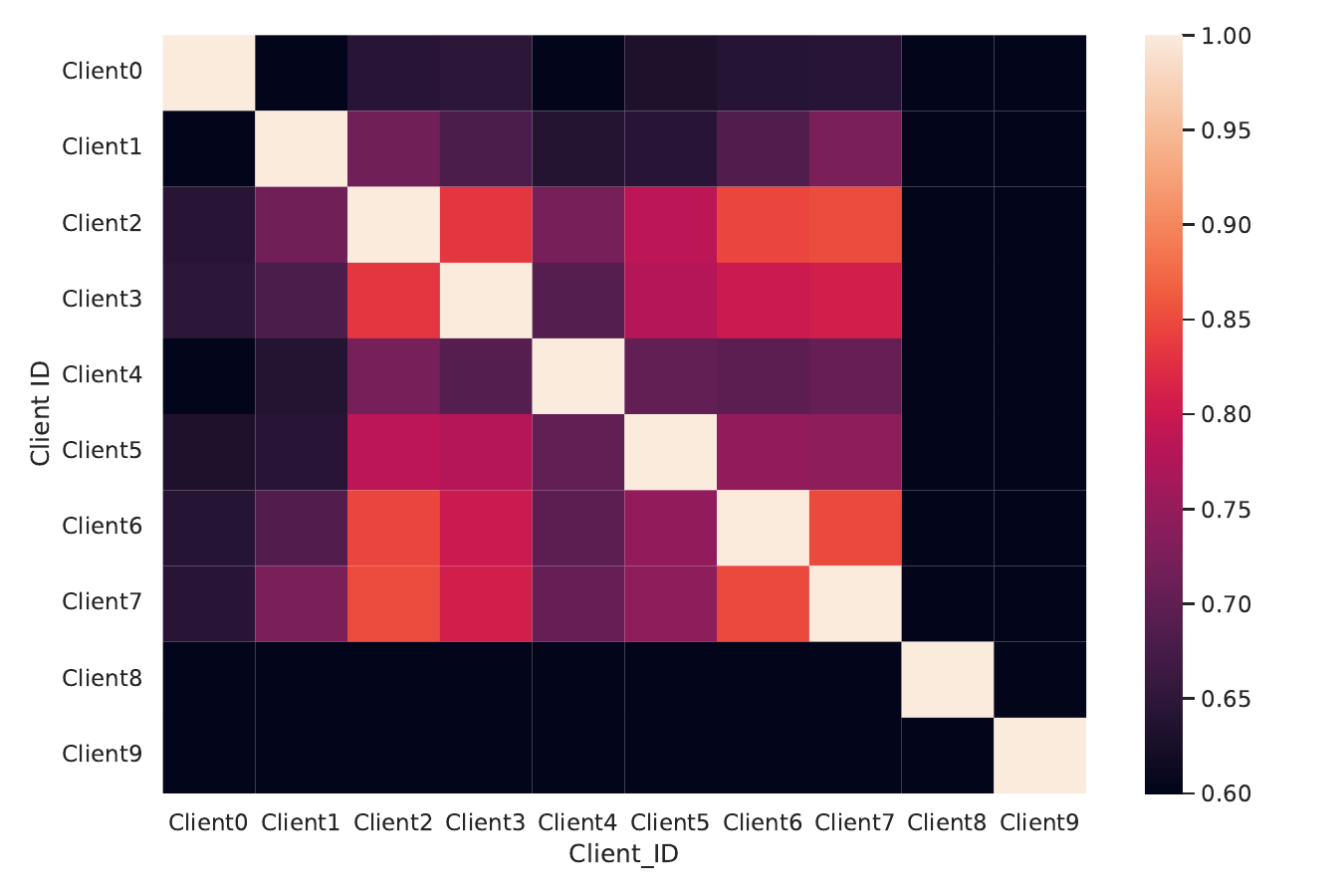}
\label{fig:apd_noniid_hetero_L7_vit}}

\caption{The CKA similarity of IID with homo, Non-IID with homo and Non-IID with hetero for ViTs.}
\label{fig:apd_heatmap_vits}
% \vskip -0.2in
\end{figure*}

In this part, we will show the heatmaps for all stages of ResNets and layer 4 to layer 7 of ViTs in Figure~\ref{fig:apd_heatmaps_resnets} and Figure~\ref{fig:apd_heatmap_vits}. These heatmaps are the concrete images for Figure~\ref{fig:Analysis_CKA}. We can see that the CKA similarity is lower with the deeper stages or layers no matter in ResNets and ViTs. However, it is notable that the different patterns for CKA similarity between ResNets and ViTs from the comparison between Figure~\ref{fig:apd_heatmaps_resnets} and Figure~\ref{fig:apd_heatmap_vits}. To get a clear analysis, we focus on the last stage of ResNets and layer 7 of ViTs, which are the most biased part of the entire model.
Like Figure~\ref{fig:apd_noniid_hetero_3} in ResNets, almost all clients are dissimilar, while only a part of clients has low similarity in ViTs from Figure~\ref{fig:apd_noniid_hetero_L7_vit}. Along with the experiment results from Table~\ref{tab:acc_100sr01}, the improvements in ViTs from FedInCo are modest. One possible reason is that we neglect more biased clients and regard all clients as having the same level of bias in ViTs, which is a possible improvement for FedInCo.

\section{More Details of the Experiments}
\label{sec:apd_exp}

\subsection{Procedure for InCo Aggregation}
The pseudo-codes for InCo Aggregation are shown in Algorithm~\ref{Algorithm:InCo_Aggregation}. InCo Aggregation is operated in a server model, indicating that the methods focused on the client can be aligned with InCo Aggregation, as shown in our experiments.

\begin{algorithm}[htbp]
    \caption{InCo Aggregation (InCoAvg as the example)}\label{Algorithm:InCo_Aggregation}
    \begin{algorithmic}[1]
    \vspace{-\baselineskip}
    \setlength\columnsep{30pt}
    \begin{multicols}{2}
    \Require Dataset $D_k, k\in\{1,...,K\}$, $K$ clients, and their weights $w_1, ..., w_K$.
    \Ensure Weights for all clients $w_1, ..., w_K$.
    \State \textbf{Server process:}
    \While{\textit{not converge}}
        \State Receives $g_{w_i}^t$ from the sampled client. 
        \State Parameter aggregation for $g_{w_i}^t$. 
        \For {\textit{each layer $l_k$ in the server model}} 
            \If {\textit{$l_k$ needs cross-layer gradients}} 
            %\tikzmark{a}
                \State ${g_{l_k}^t}', {g_{l_0}^t}'\gets$Normalizes $g_{l_k}^t$ and $g_{l_0}^t$.
                \State $\theta^t, \alpha, \beta$ from Theorem~\ref{thm:vec_solution}.
                \State $g_{l_k}^{t+1}=\frac{{(g_{l_k}^t}'-\theta^t {g_{l_0}^t}') \times({||g_{l_k}^t||+||g_{l_0}^t||})}{2}$.
                % \tikzmark{b}
            \Else
                \State $g_{l_k}^{t+1}=g_{l_k}^{t}$
            \EndIf
            \State $w_{l_k}^{t+1}=w_{l_k}^t+g_{l_k}^{t+1}$
        \EndFor
        \State Sends the updated $w_i^{t+1}$ to sampled clients.
    \EndWhile
    \State \textbf{Client processes:} 
    \While {\textit{random clients $i, i\in {1,...,K}$}}
        \State Receives model weights $w_i^{t-1}$.% from the server. 
        \State Updates client models $w_i^{t-1}$ to $w_i^t$. %according to $D_i$.
        \State Sends $g_{w_i}^t=w_i^t-w_i^{t-1}$ to the server.
    \EndWhile
    \end{multicols}
    \vspace{-\baselineskip}
    \end{algorithmic}

\end{algorithm}

\subsection{Datasets}
We conduct experiments on Fashion-MNIST, SVHN, CIFAR-10, and CINIC-10. CINIC-10 is a dataset of the mix of CIFAR-10 and ImageNet within ten classes. We use 3$\times$224$\times$224 with the ViT models and 3$\times$32$\times$32 with the ResNet models for all datasets.

\subsection{Hyper-parameters}
We deploy stage splitting for ResNets and obtain five sub-models, which can be recognized as ResNet10, ResNet14, ResNet18, ResNet22, and ResNet26. For the pre-trained ViT models, we employ layer splitting and obtain five sub-models, which are ViT-S/8, ViT-S/9, ViT-S/10, ViT-S/11, and ViT-S/12 from the Py\textbf{T}orch \textbf{Im}age \textbf{M}odels (timm)\footnote{https://github.com/rwightman/pytorch-image-models}.
Our implementations of FedAvg, FedProx, FedNova, Scaffold and MOON are referred to \citep{li2022federated}.
We use Adam optimizer with a learning rate of 0.001, $\beta_1=0.9$ and $\beta_2=0.999$, default parameter settings for all methods of ResNets.
The local training epochs are fixed to 5. The batch size is 64 for all experiments. Furthermore, the global communication rounds are 500 for ResNets, and 200 for ViTs for all datasets. Global communication rounds for MOON and InCoMOON are 100 to prevent the extreme overfitting in Fashion-MNIST. The hyper-parameter $\frac{\mu}{2}$ for \textbf{FedProx} and \textbf{InCoProx} is 0.05 for ViTs and ResNets. We conduct our experiments with 4 NVIDIA GeForce RTX 3090s. All baselines and their InCo extensions are conducted in the same hyper-parameters. The settings of hetero splitting for ScaleFL followed the source codes.\footnote{https://github.com/git-disl/scale-fl}

\begin{table*}[!t]
% \vskip -0.2in
\def\arraystretch{1.6}
% \scalebox{}
\tiny
    \caption{Test accuracy of 100 clients and sample ratio 0.1. We shade in gray the methods that are combined with our proposed method, InCo Aggregation. We show the error bars for InCo Aggregation in this table.}
    \label{tab:apd_acc_100sr01_error_bars}
    \centering
    \begin{tabular}{cccccccccc}
    \hline
    % \toprule
         \multirow{2}*{Base} & \multirow{2}*{Methods} & \multicolumn{2}{c}{Fashion-MNIST} & \multicolumn{2}{c}{SVHN} & \multicolumn{2}{c}{CIFAR10} &  \multicolumn{2}{c}{CINIC10} \\
        \cline{3-10}
         & & $\alpha=0.5$ & $\alpha=1.0$ & $\alpha=0.5$ & $\alpha=1.0$ &  $\alpha=0.5$ & $\alpha=1.0$ & $\alpha=0.5$ & $\alpha=1.0$ \\
      \hline
      % \midrule
      \multirow{10}{*}{\rotatebox{90}{ResNet}}

        & FedAvg & 86.7$\pm$1.0 & 87.7$\pm$0.6 & 74.8$\pm$3.2 & 81.6$\pm$2.5 & 52.3$\pm$3.4 & 61.3$\pm$3.2 & 43.1$\pm$2.7 & 49.2$\pm$3.1 \\
        & FedProx & 75.1$\pm$1.8 & 76.6$\pm$1.5 & 32.0$\pm$2.8 & 43.7$\pm$2.9 & 19.2$\pm$2.2 & 23.4$\pm$2.4 & 17.4$\pm$1.7 & 19.8$\pm$1.4\\
        & Scaffold & 87.9$\pm$0.5 & 88.0$\pm$0.3 & 76.3$\pm$3.4 & 82.4$\pm$3.1 & 54.3$\pm$3.6 & 61.8$\pm$3.0 & 43.5$\pm$2.4 & 49.4$\pm$3.1 \\
        & FedNova & 12.7$\pm$0.2 & 15.6$\pm$0.2 & 13.4$\pm$ 0.4 & 15.3$\pm$0.3 & 10.4$\pm$0.3 & 14.3$\pm$0.2 & 12.0$\pm$0.3 & 14.0$\pm$0.2 \\
        & MOON & 87.7$\pm$0.4 & 87.5$\pm$0.3 & 72.8$\pm$4.3 & 81.2$\pm$3.2 & 47.2$\pm$2.7 & 58.8$\pm$2.6 & 40.8$\pm$2.1 & 49.2$\pm$ 1.9\\
        %& InclusiveFL & 87.0 & 88.1 & 86.3 & 86.6 & 64.9 & 69.5 & 48.8 & 56.0\\
        \cline{2-10}
        
        & \mycc InCoAvg &  \mycc \textbf{90.2$\pm$1.2} & \mycc 88.4$\pm$1.8 & \mycc 87.6$\pm$2.8 & \mycc 89.0$\pm$2.6 & \mycc 67.8$\pm$3.2 & \mycc 70.7$\pm$3.4 & \mycc 53.0$\pm$3.2 & \mycc 57.5$\pm$3.3 \\
        
        & \mycc InCoProx & \mycc 88.8$\pm$2.3 & \mycc 86.4$\pm$3.2 & \mycc \textbf{89.0$\pm$1.3} & \mycc \textbf{90.8$\pm$1.2} & \mycc \textbf{74.5$\pm$2.3} & \mycc \textbf{76.8$\pm$1.8} & \mycc \textbf{59.1$\pm$3.2} & \mycc \textbf{62.5$\pm$2.4} \\

        & \mycc InCoScaffold & \mycc 88.3$\pm$1.4 & \mycc \textbf{90.1$\pm$1.2} & \mycc 85.4$\pm$2.4 & \mycc 87.8$\pm$3.5 & \mycc 67.3$\pm$3.6 & \mycc 73.8$\pm$2.9 & \mycc 53.5$\pm$3.3 & \mycc 61.7$\pm$3.0 \\

        & \mycc InCoNova & \mycc 86.6$\pm$1.4 & \mycc 87.4$\pm$1.3 & \mycc 86.4$\pm$2.5 & \mycc 88.4$\pm$1.8 & \mycc 62.8$\pm$3.9 & \mycc 69.7$\pm$4.2 & \mycc 48.0$\pm$2.7 & \mycc 54.1$\pm$1.7\\

       % \hdashline[1pt/5pt]

       & \mycc InCoMOON & \mycc 89.1$\pm$1.3 & \mycc 89.5$\pm$1.2 & \mycc 85.6$\pm$3.8 & \mycc 89.3$\pm$2.0 & \mycc 68.2$\pm$3.1 & \mycc 71.8$\pm$2.3 & \mycc 54.3$\pm$3.0 & \mycc 57.6$\pm$2.7\\
        \hline
        
        \multirow{10}{*}{\rotatebox{90}{ViT}}  
        & FedAvg & 92.0$\pm$0.7 & 91.9$\pm$0.5 & 92.4$\pm$0.9 & 93.9$\pm$0.8 & 93.7$\pm$1.0 & 94.2$\pm$0.8 & 83.8$\pm$1.4 & 85.1$\pm$0.9 \\
        & FedProx & 89.8$\pm$0.5 & 89.7$\pm$0.5 & 71.4$\pm$3.8 & 81.1$\pm$2.9 & 82.6$\pm$3.3 & 84.7$\pm$2.3 & 67.8$\pm$2.8 & 71.3$\pm$3.0 \\
        & Scaffold & 92.0$\pm$0.4 & 92.0$\pm$0.5 & 92.2$\pm$0.8 & 93.8$\pm$0.6 & 93.5$\pm$0.7 & 94.5$\pm$0.5 & 83.3$\pm$1.6 & 85.5$\pm$1.2\\
        & FedNova & 70.3$\pm$0.5 & 76.7$\pm$0.4 & 27.4$\pm$0.4 & 49.8$\pm$0.5 & 30.7$\pm$0.3 & 54.4$\pm$0.5 & 31.6$\pm$1.5 & 50.7$\pm$1.3 \\
        & MOON & 92.1$\pm$0.3 & 92.1$\pm$0.2 & 92.5$\pm$1.2 & 93.9$\pm$0.9 & 93.6$\pm$0.8 & 94.6$\pm$0.3 & 84.3$\pm$1.6 & 85.3$\pm$1.2 \\
        %& InclusiveFL & \\
        \cline{2-10}
       & \mycc InCoAvg & \mycc 93.0$\pm$0.6 & \mycc 93.1$\pm$0.5  & \mycc 94.2$\pm$0.6 & \mycc 95.0$\pm$0.4 & \mycc 94.6$\pm$0.7 & \mycc 95.0$\pm$0.6 & \mycc 85.9$\pm$1.9 & \mycc 86.8$\pm$1.3 \\
       
        & \mycc InCoProx & \mycc 92.6$\pm$0.3 & \mycc 92.5$\pm$0.3 & \mycc 93.9$\pm$0.7 & \mycc 94.4$\pm$0.6 & \mycc 94.0$\pm$1.0 & \mycc 94.8$\pm$0.7 & \mycc 85.1$\pm$1.4 & \mycc 86.0$\pm$0.8\\

        & \mycc InCoScaffold & \mycc 92.9$\pm$0.3 & \mycc 93.0$\pm$0.2 & \mycc 94.0$\pm$1.1 & \mycc 94.8$\pm$0.6 & \mycc 94.6$\pm$0.5 & \mycc 95.0$\pm$0.2 & \mycc 85.7$\pm$1.3 & \mycc 86.5$\pm$1.1\\

        & \mycc InCoNova & \mycc \textbf{93.1}$\pm$0.3 & \mycc \textbf{93.6}$\pm$0.3 & \mycc 94.7$\pm$0.9 & \mycc \textbf{95.6$\pm$0.5} & \mycc \textbf{94.8}$\pm$0.4 & \mycc \textbf{95.7$\pm$0.3} & \mycc \textbf{86.2$\pm$1.8} & \mycc \textbf{88.2$\pm$1.0} \\

       % \hdashline[1pt/5pt]

       & \mycc InCoMOON & \mycc 92.8$\pm$0.5 & \mycc 93.0$\pm$0.3 & \mycc \textbf{94.7$\pm$0.8} & \mycc 95.1$\pm$0.5 & \mycc 94.2$\pm$0.8 & \mycc 95.1$\pm$0.5 & \mycc 86.0$\pm$0.9 & \mycc 86.8$\pm$1.3\\

    % \hline
    \hline
    % \bottomrule
    \end{tabular}
% \vskip -0.2in
\end{table*}

\begin{table*}[!t]
\vskip -0.1in
\def\arraystretch{1.6}
% \scalebox{}
\scriptsize
    \caption{Test accuracy of model-heterogeneity methods with 100 clients and sample ratio 0.1. We shade in gray the methods that are combined with our proposed method, InCo Aggregation. We show the error bars for InCo Aggregation in this table.}
    \label{tab:apd_acc_100sr01_hetero_error_bars}
    \centering
    \begin{tabular}{ccccccccc}
    % \hline
    \toprule
         \multirow{2}*{Base} & \multirow{2}*{Splitting} & \multirow{2}*{Methods} & \multicolumn{2}{c}{Fashion-MNIST} & \multicolumn{2}{c}{SVHN} & \multicolumn{2}{c}{CIFAR10} \\
        \cline{4-9}
         & & & $\alpha=0.5$ & $\alpha=1.0$ & $\alpha=0.5$ & $\alpha=1.0$ &  $\alpha=0.5$ & $\alpha=1.0$ \\
      % \hline
      \midrule
      \multirow{10}{*}{\rotatebox{90}{ResNet}}  

        & \multirow{2}*{Hetero} & HeteroFL & 88.9$\pm$1.0 & 89.7$\pm$0.7 & 90.5$\pm$1.6 & 92.2$\pm$1.3 & 65.2$\pm$3.2 & 68.4$\pm$3.6  \\

        & & \mycc +InCo &  \mycc 90.0$\pm$1.2 & \mycc 90.4$\pm$1.1 & \mycc 92.1$\pm$1.0 & \mycc 93.5$\pm$1.5 & \mycc 68.2$\pm$3.8 & \mycc 71.2$\pm$3.4  \\

        % \cline{2-9}
        
        & \multirow{2}*{Stage} & InclusiveFL & 89.1$\pm$1.1 & 89.8$\pm$1.0 & 88.6$\pm$2.0 & 90.0$\pm$2.2 & 65.7$\pm$3.5 & 68.4$\pm$3.3 \\

        &  & \mycc +InCo & \mycc 90.1$\pm$1.5 & \mycc 90.5$\pm$1.3 & \mycc 90.6$\pm$1.7 & \mycc 90.9$\pm$1.9 & \mycc 69.1$\pm$2.8 & \mycc 72.3$\pm$3.1 \\

        % \cline{2-9}
        
        & \multirow{2}*{Hetero} & FedRolex & 88.2$\pm$1.0 & 90.2$\pm$0.8 & 90.9$\pm$1.3 & 91.6$\pm$1.7 & 64.7$\pm$4.1 & 72.3$\pm$3.0 \\

        & & \mycc +InCo & \mycc 90.4$\pm$1.4 & \mycc 91.3$\pm$1.1 & \mycc 92.8$\pm$1.5 & \mycc 93.4$\pm$1.6 & \mycc 67.9$\pm$2.9 & \mycc 75.6$\pm$2.6 \\

        % \cline{2-9}
        
        & \multirow{2}*{Hetero} & ScaleFL & 90.9$\pm$0.5 & 91.0$\pm$0.4 & 92.6$\pm$1.0 & 92.9$\pm$0.9 & 71.1$\pm$2.9 & 74.7$\pm$3.1 \\

        & & \mycc +InCo & \mycc 91.5$\pm$1.0 & \mycc 91.7$\pm$1.1 & \mycc 93.4$\pm$0.9 & \mycc 93.6$\pm$0.9 & \mycc 73.8$\pm$3.2 & \mycc 76.1$\pm$2.6 \\

        \cline{2-9}

        & N/A & AllSmall & 83.5$\pm$1.7 & 84.0$\pm$1.7 & 72.1$\pm$3.5 & 81.0$\pm$2.9 & 39.2$\pm$2.0 & 44.9$\pm$2.3 \\
        
        & N/A & AllLarge & 91.8$\pm$0.5 & 92.5$\pm$0.8 & 93.4$\pm$0.8 & 93.8$\pm$0.5 & 79.6$\pm$2.9 & 82.5$\pm$1.0 \\
       
    % \hline
    % \hline
    \bottomrule
    \end{tabular}
\vskip -0.1in
\end{table*}

% \textcolor{red}{explained FedProx and FedNova for Table~\ref{tab:acc_100sr01}.}

\subsection{Model Sizes}
\label{apd:model_sizes}
We demonstrate the model sizes for each client model in Table~\ref{tab:apd_model_sizes_resnet_stage}, Table~\ref{tab:apd_model_sizes_vit_layer}, and Table~\ref{tab:apd_model_sizes_resnet_hetero}.

\begin{table*}[ht]
% \vskip -0.2in
\def\arraystretch{1.6}
% \scalebox{}
\scriptsize
    \caption{Model parameters for different architectures of ResNets (Stage splitting).}
    \label{tab:apd_model_sizes_resnet_stage}
    \centering
    \begin{tabular}{cccccc}
    \hline
    % \toprule
         \multirow{2}*{Sizes} & \multicolumn{5}{c}{ResNets (Stage splitting)} \\
        \cline{2-6}
          & ResNet10 & ResNet14 & ResNet18 & ResNet22 & ResNet26  \\
        \hline
          Params & 4.91M ($\times 0.281$) & 10.81M ($\times 0.619$) & 11.18M ($\times 0.641$) & 17.08M ($\times 0.979$) & 17.45M ($\times 1$) \\
    \hline
    % \bottomrule
    \end{tabular}
% \vskip -0.2in
\end{table*}

\begin{table*}[ht]
% \vskip -0.2in
\def\arraystretch{1.6}
% \scalebox{}
\scriptsize
    \caption{Model parameters for different architectures of ViTs (Layer splitting).}
    \label{tab:apd_model_sizes_vit_layer}
    \centering
    \begin{tabular}{cccccc}
    \hline
    % \toprule
         \multirow{2}*{Sizes} & \multicolumn{5}{c}{ViTs (Layer splitting)} \\
        \cline{2-6}
          & ViT-S/8 & ViT-S/9 & ViT-S/10 & ViT-S/11 & ViT-S/12  \\
        \hline
          Params & 14.57M ($\times 0.672$) & 16.34M ($\times 0.754$) & 18.12M ($\times 0.836$) & 19.90M ($\times 0.912$) & 21.67M ($\times 1$) \\
    \hline
    % \bottomrule
    \end{tabular}
% \vskip -0.2in
\end{table*}

\begin{table*}[ht]
% \vskip -0.2in
\def\arraystretch{1.6}
% \scalebox{}
\scriptsize
    \caption{Model parameters for different architectures of ResNets (Heterogeneous splitting).}
    \label{tab:apd_model_sizes_resnet_hetero}
    \centering
    \begin{tabular}{cccccc}
    \hline
    % \toprule
         \multirow{2}*{Sizes} & \multicolumn{5}{c}{ResNets (Hetero splitting)} \\
        \cline{2-6}
          & \sfrac{1}{16} & \sfrac{1}{8} & \sfrac{1}{4} & \sfrac{1}{2} & ResNet26  \\
        \hline
          Params & 0.07M ($\times 0.004$) & 0.28M ($\times 0.016$) & 1.10M ($\times 0.06$) & 4.37 ($\times 0.25$) & 17.45M ($\times 1$) \\
    \hline
    % \bottomrule
    \end{tabular}
% \vskip -0.2in
\end{table*}

\subsection{Error Bars of InCo Aggregation}
\label{error_bars}
We illustrate the error bars of InCo Aggregation and the results from model-homogeneous baselines (not use stage or layer splitting in the model heterogeneous environment) in Table~\ref{tab:apd_acc_100sr01_error_bars}. For the model-heterogeneity methods, we demonstrate the error bars of InCo Aggregation in Table~\ref{tab:apd_acc_100sr01_hetero_error_bars}. These results show the stability of InCo Aggregation. In all cases of ResNets and many cases of ViTs, the worst results of InCo Aggregation are better than the Averaging Aggregation, demonstrating the efficacy of InCo Aggregation.

\begin{figure*}[!t]
% \vskip -0.3in
\centering
% \hspace{-0.5in}
% \begin{minipage}[b]{.76\textwidth}
    \subfloat[CIFAR10 with $\alpha=0.5$.]
    {\includegraphics[width=0.25\textwidth]{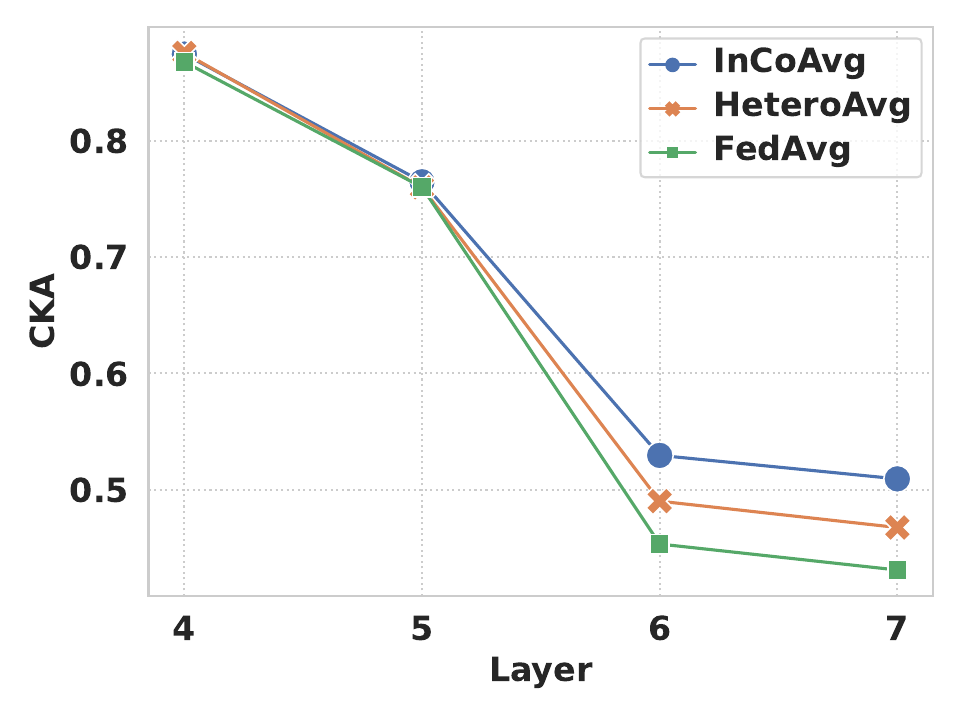}
    \label{fig:exp_vit_05cifar}}
    \hfil
    \subfloat[InCoAvg.] 
    { 
    \includegraphics[width=0.163\textwidth]{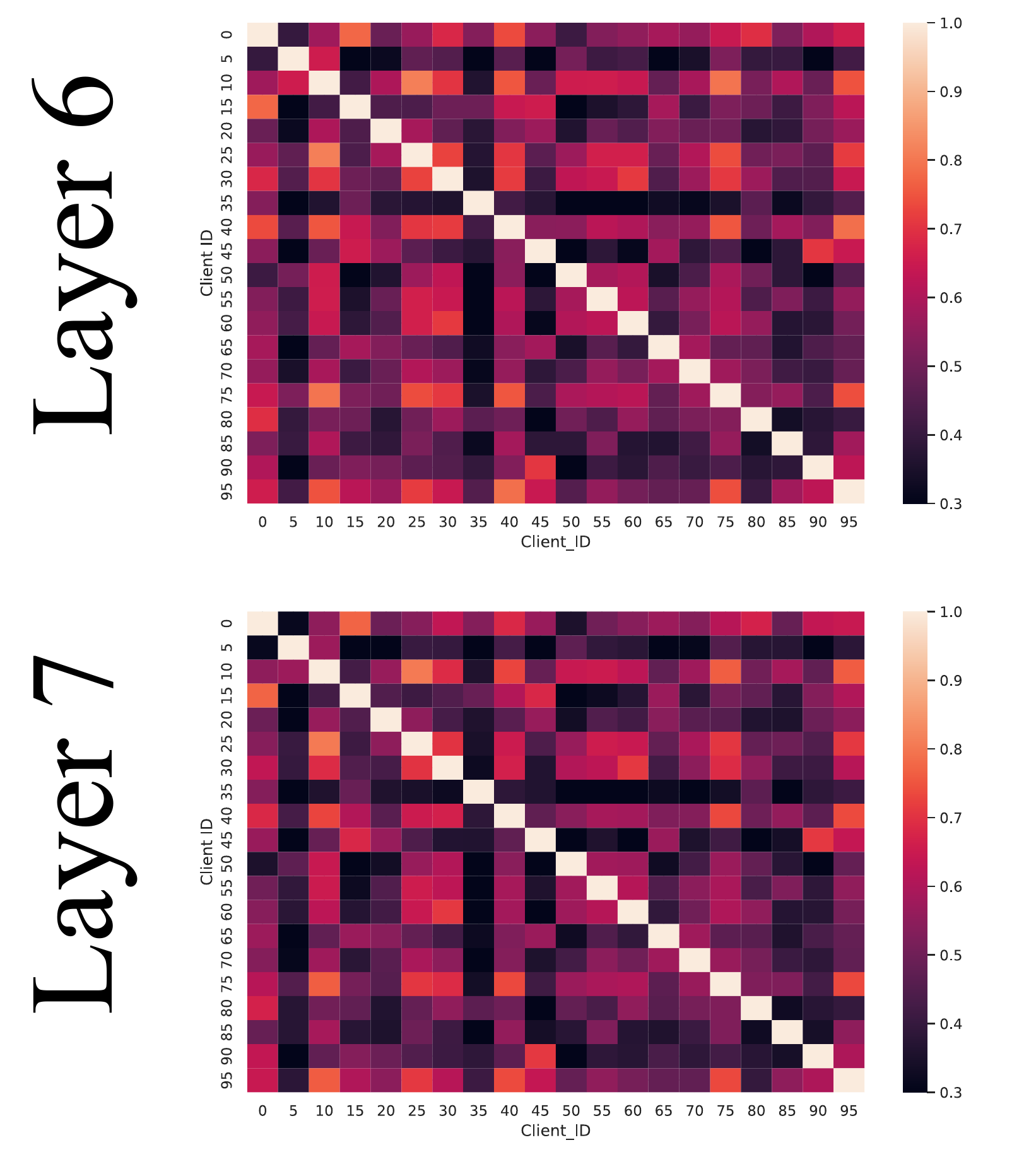}
    \label{fig:exp_vit_FedInCo_layer7_05cifar}}
    % \hspace{5pt}
    \subfloat[HeteroAvg.]
    {\includegraphics[width=0.133\textwidth]{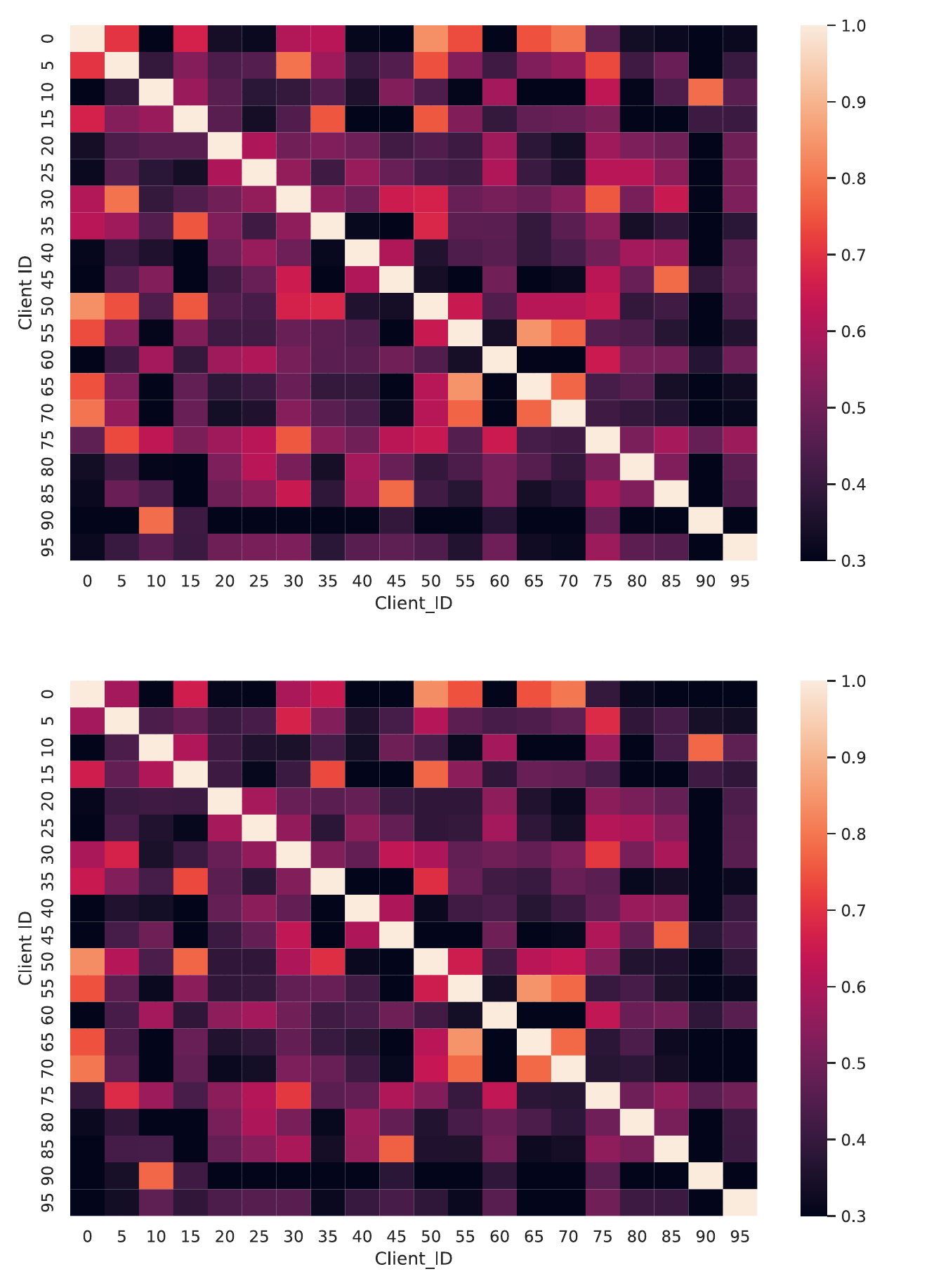}
    \label{fig:exp_vit_heteroAvg_layer7_05cifar}}
    % \hspace{5pt}
    \subfloat[FedAvg.]
    {\includegraphics[width=0.133\textwidth]{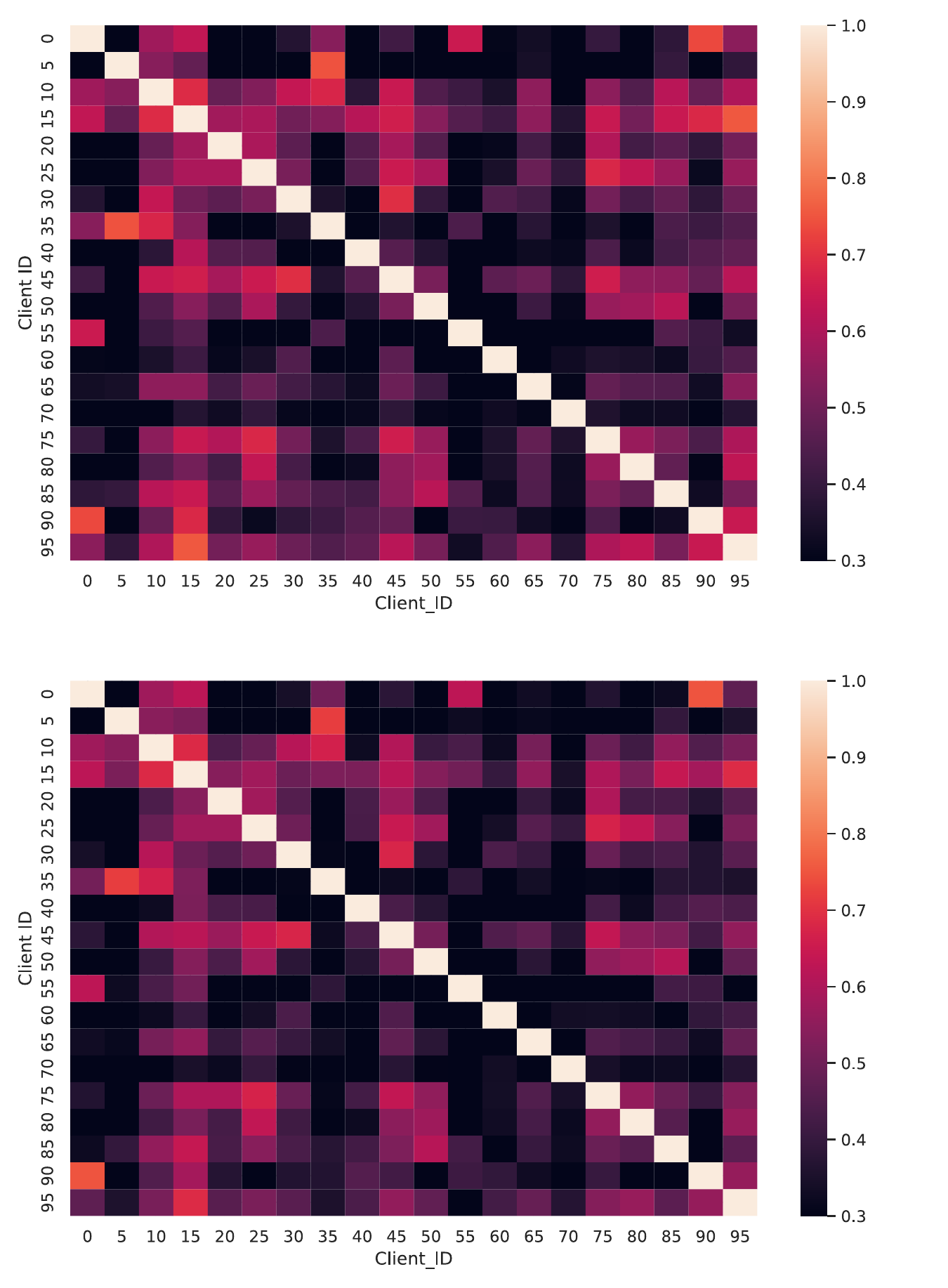}
    \label{fig:exp_vit_fedAvg_layer7_05cifar}
    }
    
% \vskip -0.07in
\caption{CKA layer similarity and Heatmaps of ViTs. (a): The layer similarity of different methods. (b) to (d): Heatmaps for different methods in layer 6 and layer 7.}
\label{fig:exp_cka_vit}
% \vspace{-0.285in}
\end{figure*}

\begin{figure*}[ht]
% \vskip -0.5in
\centering
% \hspace{-0.5in}
% \begin{minipage}[b]{.76\textwidth}
    \subfloat[Convergence speeds.]
    {\includegraphics[width=0.24\columnwidth]{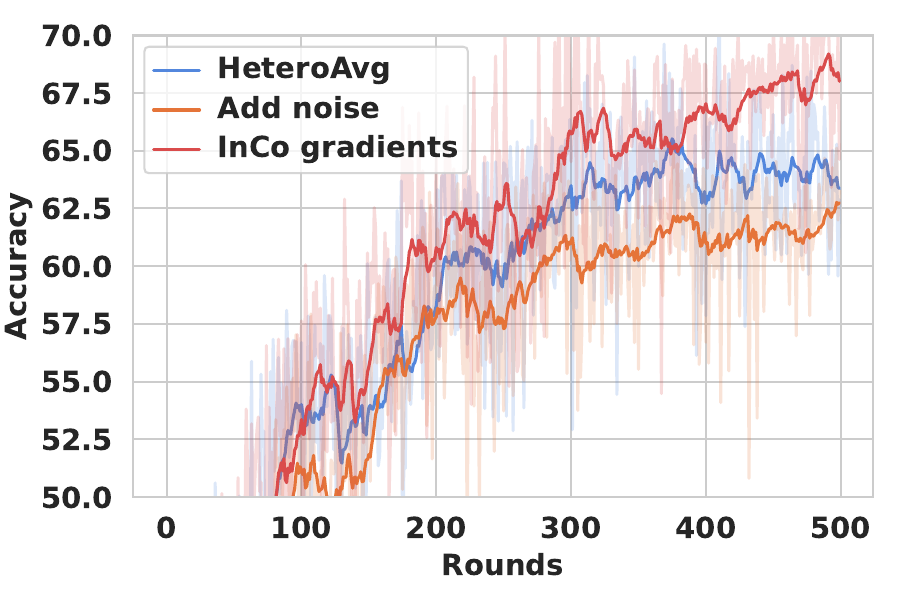}
    \label{fig:cifar10_noise_inco}}
    \hfil
    \subfloat[Frobenius norm.]
    {\includegraphics[width=0.24\columnwidth]{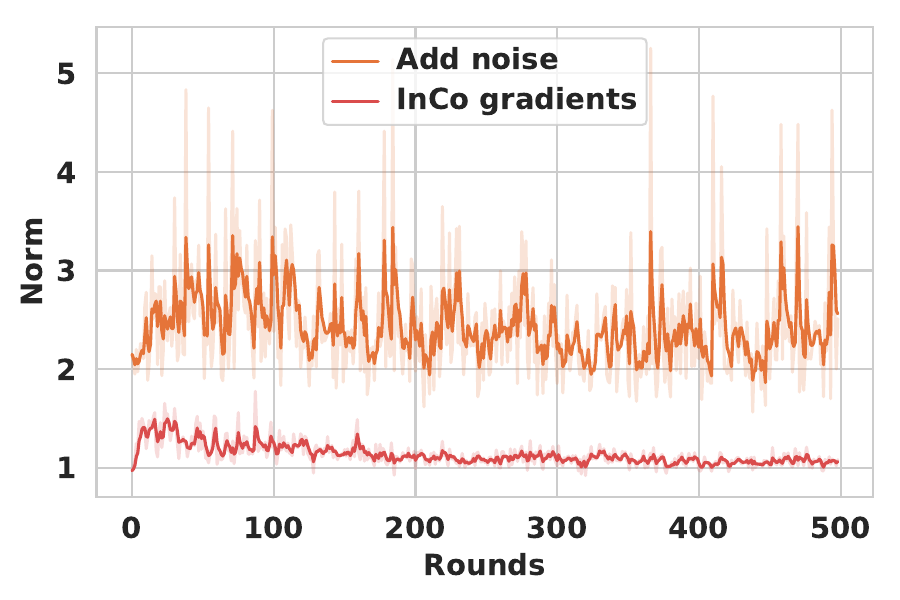}
    \label{fig:cifar10_noise_inco_norm}}
    % \vspace{-0.2cm}
    % \caption{Cross-layer gradients for the server model in InCo.}
    \hfil
    \subfloat[Accuracy for different datasets.]{
     \raisebox{\height}{
        \def\arraystretch{1.3}
        \tiny
        \begin{tabular}{ccccccc}
        \toprule
           \multirow{2}*{Methods}   & \multicolumn{2}{c}{FashionMNIST} & \multicolumn{2}{c}{SVHN} & \multicolumn{2}{c}{CIFAR10} \\
           \cline{2-7}
           & 0.5 & 1.0 & 0.5 & 1.0 & 0.5 & 1.0 \\
          \midrule
           HeteroAvg & 87.8 & 86.0 & 85.1 & 86.9 & 64.8 & 66.7  \\
           Add noises & 87.0 & 86.7 & 83.3 & 86.2 & 62.5 & 64.9  \\
           \mycc InCo gradients & \mycc \textbf{90.2} & \mycc \textbf{88.4} & \mycc \textbf{87.6} & \mycc \textbf{89.0} & \mycc \textbf{67.8} & \mycc \textbf{70.7}  \\
        \bottomrule
        \end{tabular}
        \label{tab:noise_inco_acc}
        }}
\vskip -0.1in
\caption{Convergence speeds, Frobenius norm of adding noise and InCo gradients (InCoAvg), and accuracy results for different datasets. (a): Convergence speeds of HeteroAvg, adding noise and InCo gradients. (b): Frobenius norm of noise and InCo gradients. (c): Accuracy for different datasets.}
\label{fig:noise_inco}
% \vskip -0.15in
\end{figure*}

\subsection{Differences between adding noises and InCo gradients.}
\label{apd:differences_noise_inco}
The convergence speed of InCo gradients surpasses that of the other two methods, as illustrated in Figure~\ref{fig:cifar10_noise_inco}. Table~\ref{tab:noise_inco_acc} demonstrates that InCo gradients outperform other methods across different datasets.
The primary distinction between InCo gradients and adding noises lies in their ability to determine the precise gradient modification for each node in the model. Gaussian noise lacks the capability to specify the exact modification required for each node, leading to less controlled and targeted adjustments. This is evident in Figure~\ref{fig:cifar10_noise_inco_norm}, where the Frobenius norm of noises is larger and exhibits greater instability compared to InCo gradients.

\subsection{Layer Similarity and Heatmaps of ViTs}
Figure~\ref{fig:exp_cka_vit} illustrates the layer similarity of the last four layers, along with the corresponding heatmaps for Layer 6 and Layer 7. Furthermore, Figure~\ref{fig:exp_vit_05cifar} demonstrates that InCo Aggregation significantly enhances the layer similarity, validating the initial motivation behind our proposed method. Additionally, since the disparity in layer similarity between Layer 6 and Layer 7 is minimal, the heatmaps for these layers do not exhibit significant differences, as depicted in Figure~\ref{fig:exp_vit_FedInCo_layer7_05cifar} through Figure~\ref{fig:exp_vit_fedAvg_layer7_05cifar}.

\begin{figure}[!t]
% \vskip 0.2in
\centering
\subfloat[Fashion-MNIST.]
{\includegraphics[width=0.23\columnwidth]{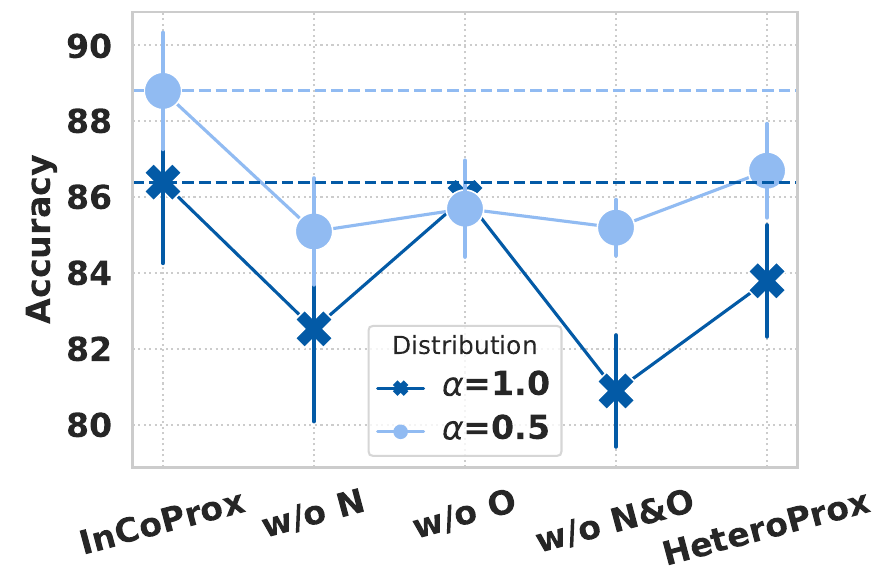}
\label{fig:ablation_resnet_fashion_mnist_prox}}
\hfil
\subfloat[SVHN.]
{\includegraphics[width=0.23\columnwidth]{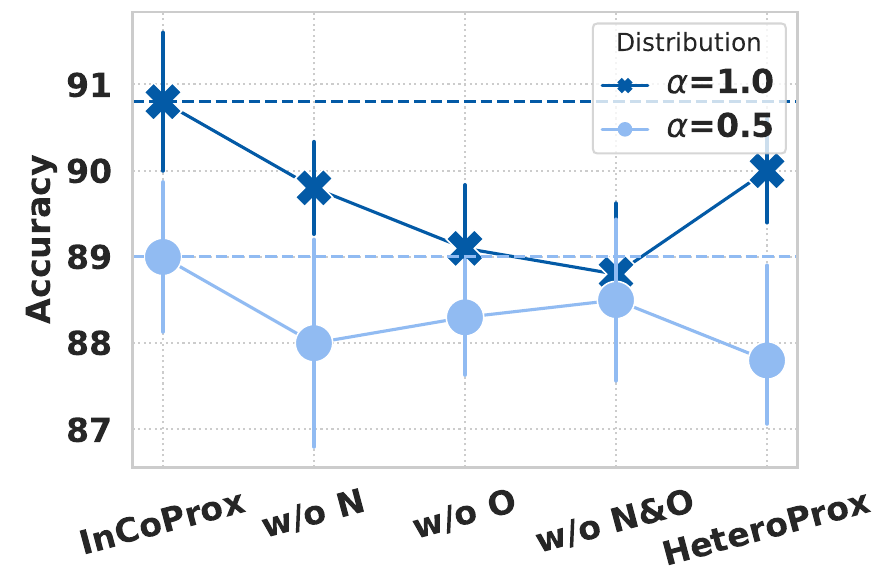}
\label{fig:ablation_resnet_svhn_prox}}
\hfil
\subfloat[CIFAR-10.]
{\includegraphics[width=0.23\columnwidth]{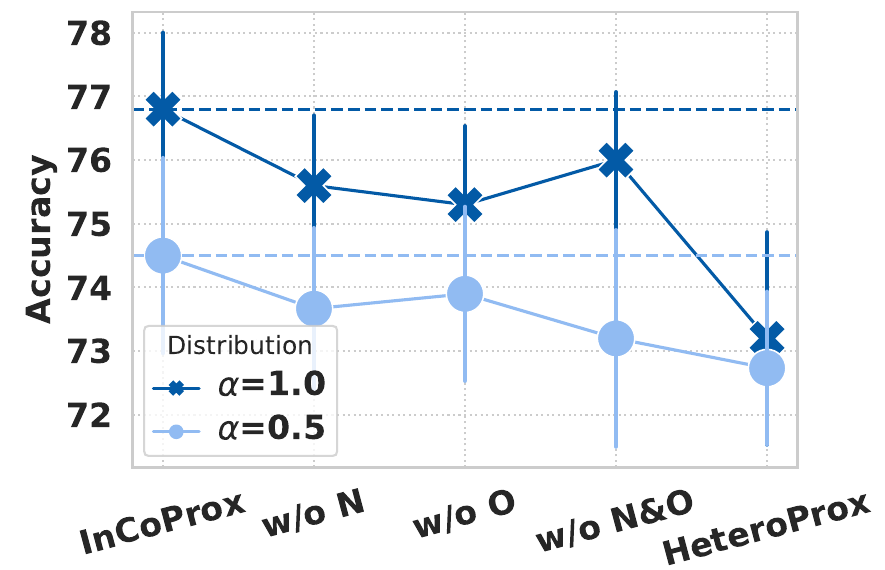}
\label{fig:ablation_resnet_cifar_prox}}
\hfil
\subfloat[CINIC-10.]
{\includegraphics[width=0.23\columnwidth]{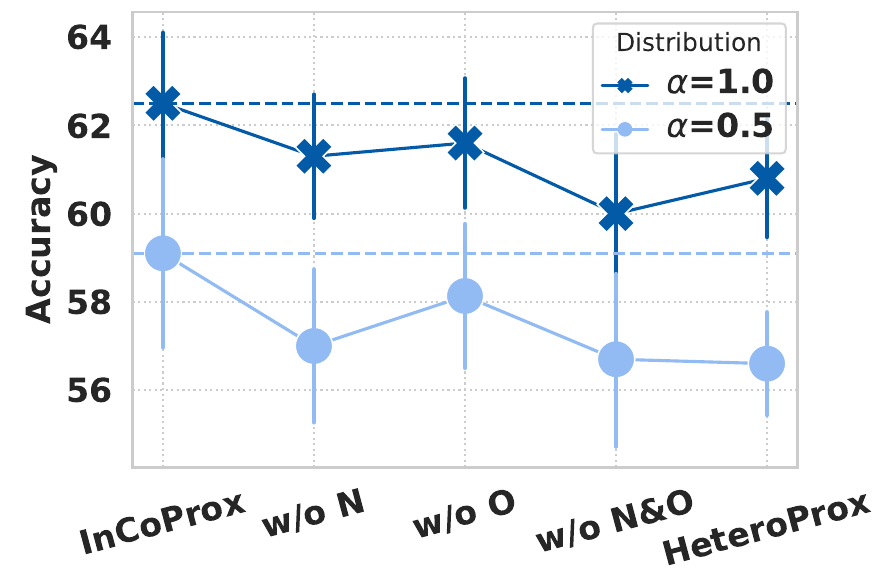}
\label{fig:ablation_resnet_cinic_prox}}
\vskip -0.07in
\caption{Ablation studies for InCo Aggregation for FedProx. The federated settings are the same as Table~\ref{tab:acc_100sr01}.}
\label{fig:ablation_resnet_prox}
\end{figure}

\begin{figure}[!t]
% \vskip 0.2in
\centering
\subfloat[Fashion-MNIST.]
{\includegraphics[width=0.23\columnwidth]{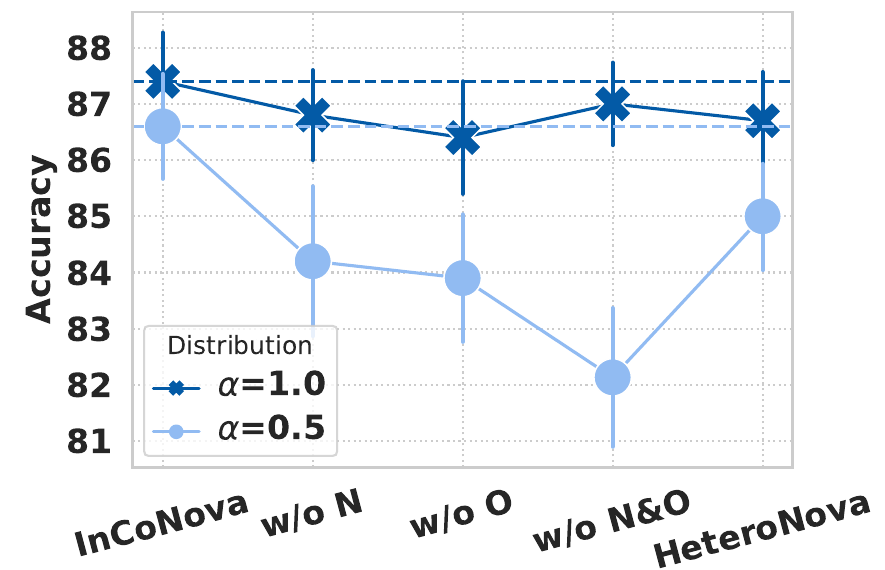}
\label{fig:ablation_resnet_fashion_mnist_nova}}
\hfil
\subfloat[SVHN.]
{\includegraphics[width=0.23\columnwidth]{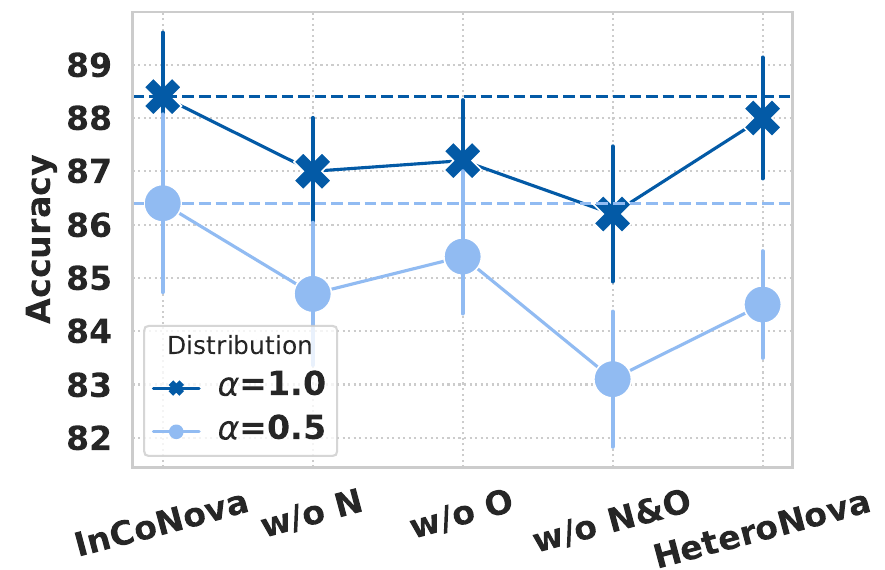}
\label{fig:ablation_resnet_svhn_nova}}
\hfil
\subfloat[CIFAR-10.]
{\includegraphics[width=0.23\columnwidth]{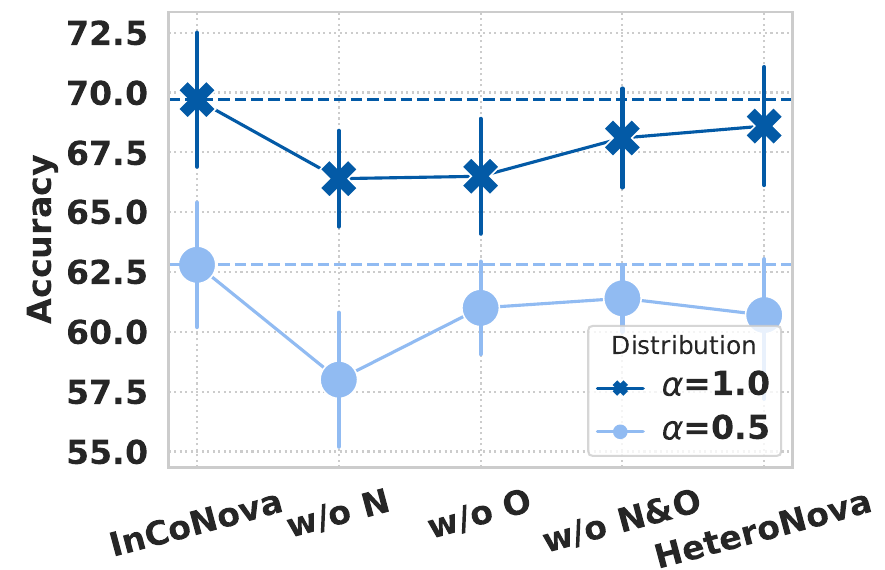}
\label{fig:ablation_resnet_cifar_nova}}
\hfil
\subfloat[CINIC-10.]
{\includegraphics[width=0.23\columnwidth]{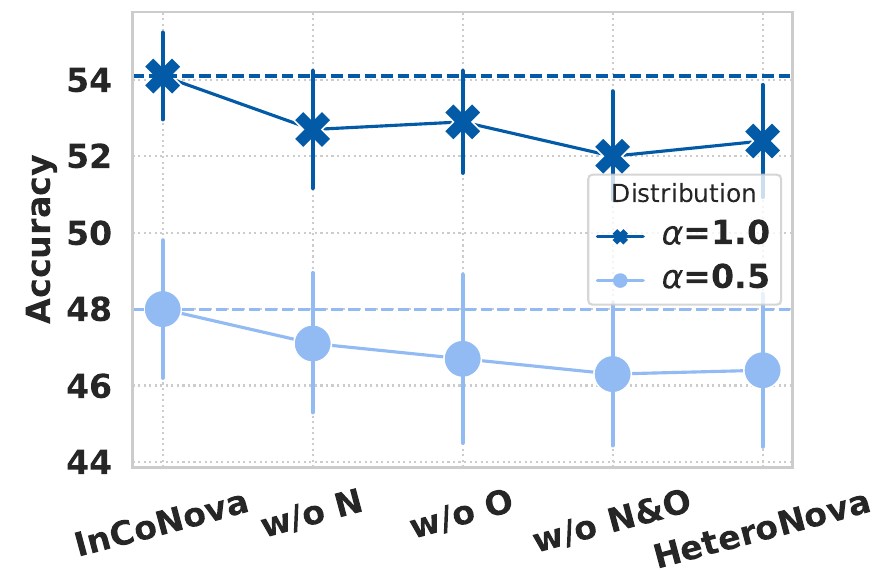}
\label{fig:ablation_resnet_cinic_nova}}
\vskip -0.07in
\caption{Ablation studies for InCo Aggregation for FedNova. The federated settings are the same as Table~\ref{tab:acc_100sr01}.}
\label{fig:ablation_resnet_nova}
\end{figure}

\subsection{More Ablation Studies and Robustness Analysis.}
\begin{figure}[!t]
% \vskip 0.2in
\centering
\subfloat[Fashion-MNIST.]
{\includegraphics[width=0.23\columnwidth]{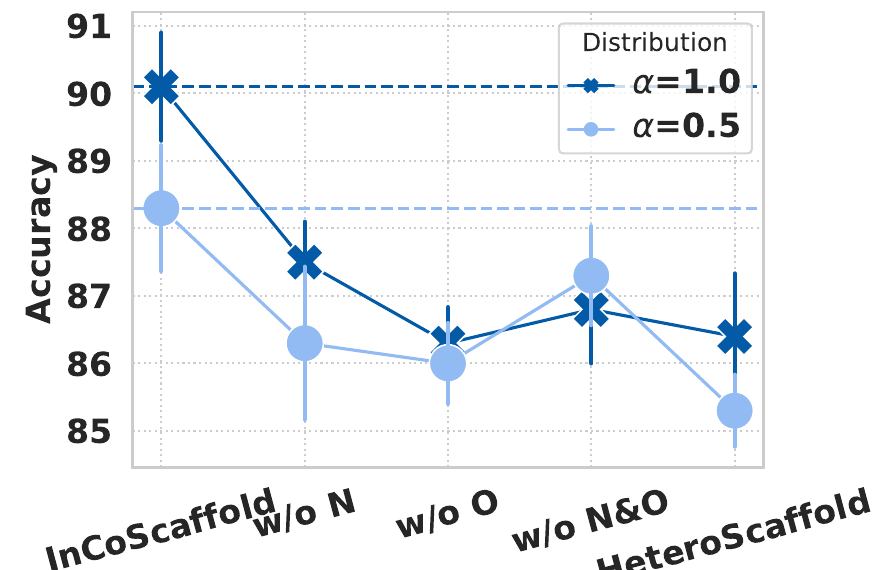}
\label{fig:ablation_resnet_fashion_mnist_scaffold}}
\hfil
\subfloat[SVHN.]
{\includegraphics[width=0.23\columnwidth]{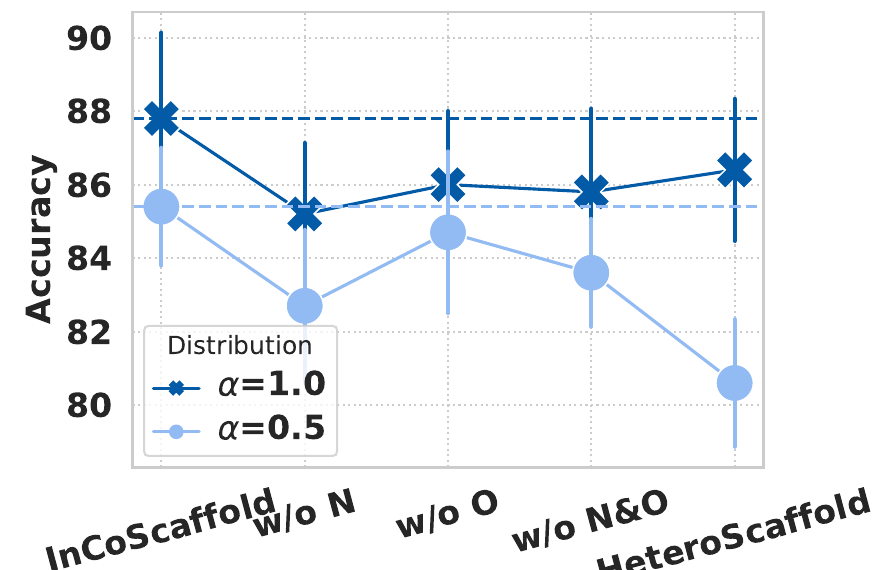}
\label{fig:ablation_resnet_svhn_scaffold}}
\hfil
\subfloat[CIFAR-10.]
{\includegraphics[width=0.23\columnwidth]{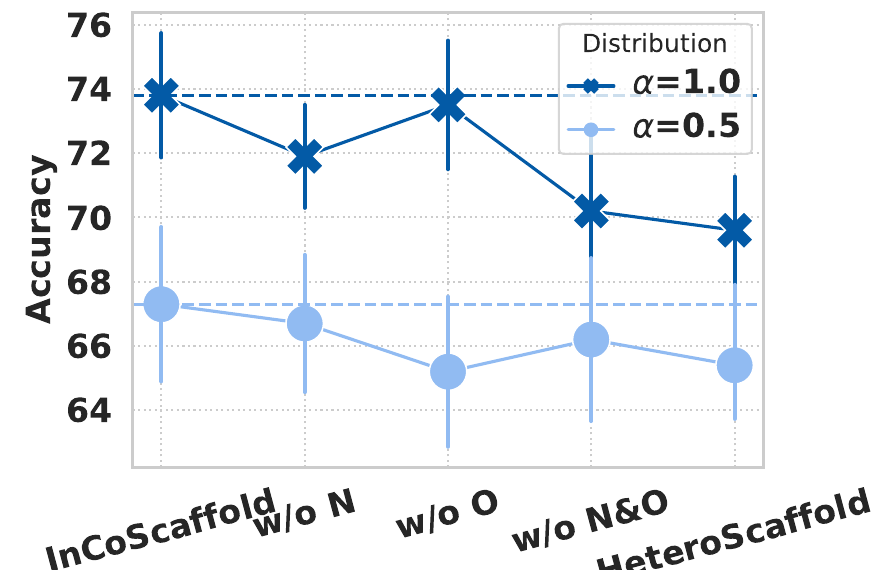}
\label{fig:ablation_resnet_cifar_scaffold}}
\hfil
\subfloat[CINIC-10.]
{\includegraphics[width=0.23\columnwidth]{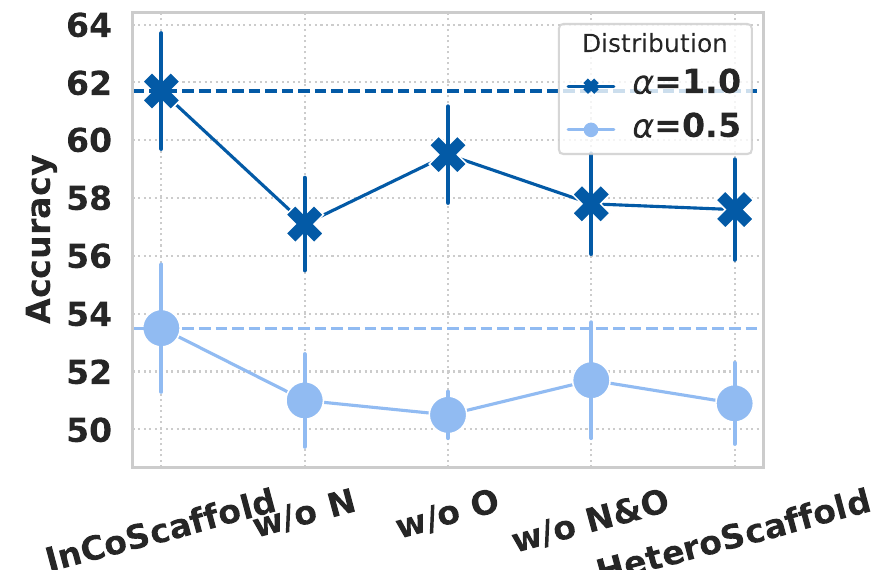}
\label{fig:ablation_resnet_cinic_scaffold}}
\vskip -0.07in
\caption{Ablation studies for InCo Aggregation for Scaffold. The federated settings are the same as Table~\ref{tab:acc_100sr01}.}
\label{fig:ablation_resnet_scaffold}
\end{figure}

\begin{figure}[!t]
% \vskip 0.2in
\centering
\subfloat[Different batch sizes.]
{\includegraphics[width=0.46\columnwidth]{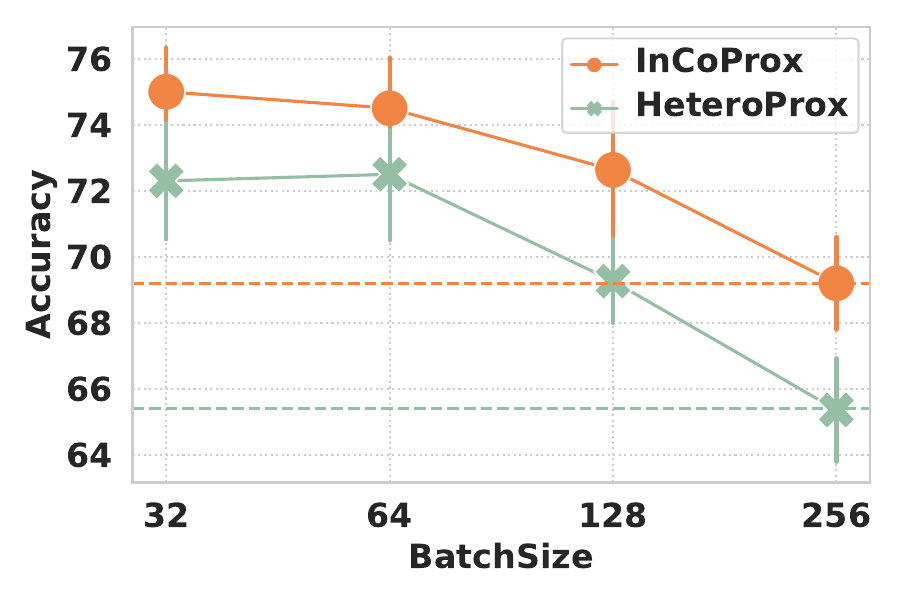}
\label{fig:batchsize_resnet_cifar_prox}}
\hfil
\subfloat[Different noise perturbations.]
{\includegraphics[width=0.46\columnwidth]{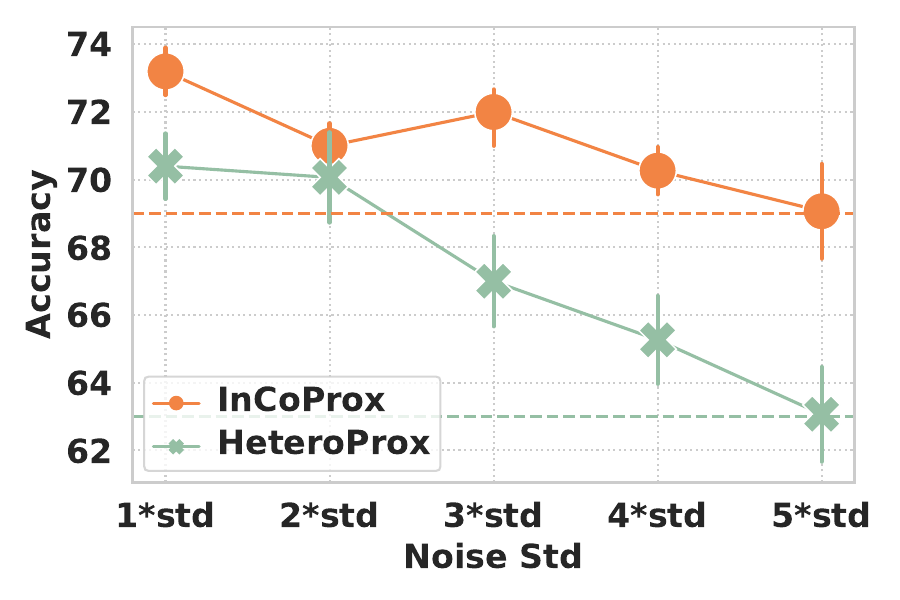}
\label{fig:noise_resnet_cifar_prox}}
\vskip -0.07in
\caption{Robustness analysis for InCo Aggregation for FedProx in CIFAR-10.}
\label{fig:robustness_resnet_prox}
\end{figure}

\begin{figure}[!t]
% \vskip 0.2in
\centering
\subfloat[Different batch sizes.]
{\includegraphics[width=0.46\columnwidth]{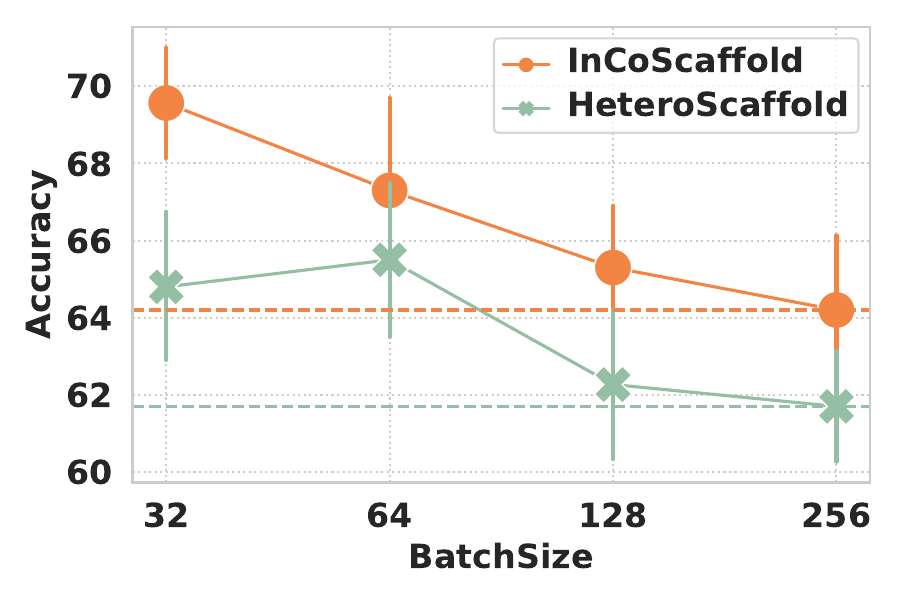}
\label{fig:batchsize_resnet_cifar_scaffold}}
\hfil
\subfloat[Different noise perturbations.]
{\includegraphics[width=0.46\columnwidth]{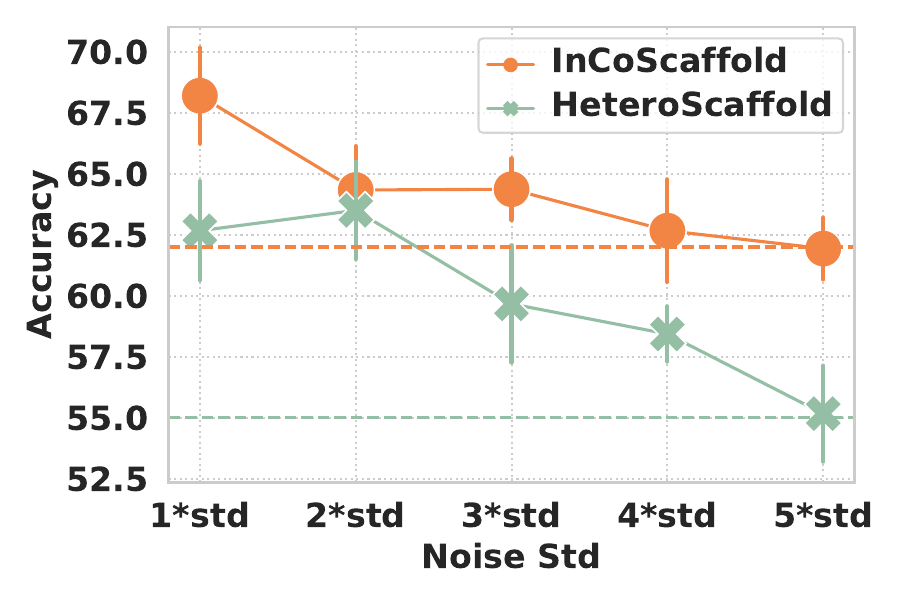}
\label{fig:noise_resnet_cifar_scaffold}}
\vskip -0.07in
\caption{Robustness analysis for InCo Aggregation for Scaffold in CIFAR-10.}
\label{fig:robustness_resnet_scaffold}
\end{figure}

We conduct additional experiments on different baselines to demonstrate the effectiveness of InCo Aggregation. Figure~\ref{fig:ablation_resnet_prox} to Figure~\ref{fig:ablation_resnet_scaffold} present the results of the ablation study for FedProx, FedNova, and Scaffold, incorporating InCo Aggregation. These results highlight the efficacy of InCo Aggregation across different baselines. Additionally, Figure~\ref{fig:robustness_resnet_prox} and Figure~\ref{fig:robustness_resnet_scaffold} illustrate the robustness analysis for FedProx and Scaffold. In Figure~\ref{fig:robustness_resnet_prox} and Figure~\ref{fig:robustness_resnet_scaffold}, InCoProx and InCoScaffold consistently obtains the best performances across all settings. These experiments provide further evidence of the efficiency of InCo Aggregation.

\section{Limitations and Future Directions}

The objective of this study is to expand the capabilities of model-homogeneous methods to effectively handle model-heterogeneous FL environments.
However, the analysis of layer similarity reveals that the smallest models do not derive substantial benefits from InCo Aggregation, implying the limited extensions for these smallest models. Exploring methods to enhance the performance of the smallest models warrants further investigation.
Furthermore, our research mainly focuses on image classification tasks, specifically CNN models (ResNets) and Transformers (ViTs). However, it is imperative to validate our conclusions in the context of language tasks, and other model architectures such as LSTM \cite{hochreiter1997long}. Additionally, it is important to consider that the participating clients in the training process may have different model architectures. For example, some clients may employ CNN models, while others may use Vision Transformers (ViTs). We believe that it is worth extending this work to encompass a wider range of tasks and diverse model architectures that hold great value and potential for future research.

%%%%%%%%%%%%%%%%%%%%%%%%%%%%%%%%%%%%%%%%%%%%%%%%%%%%%%%%%%%%

\end{document}